\documentclass[twocolumn,english,10pt,journal,cspaper,compsoc]{IEEEtran}
\usepackage[T1]{fontenc}
\usepackage{color}
\usepackage{array}
\usepackage{float}
\usepackage{multirow}
\usepackage{amsmath}
\usepackage{amssymb}
\usepackage{graphicx}
\usepackage{fancybox}
\usepackage{tikz}
\usepackage{verbatim}
\usetikzlibrary{arrows,shapes,chains,matrix,positioning}
\usetikzlibrary{scopes,decorations,shadows,backgrounds,fit}
\usetikzlibrary{decorations.pathreplacing}

\providecommand{\tabularnewline}{\\}
\floatstyle{ruled}
\newfloat{algorithm}{tbp}{loa}
\providecommand{\algorithmname}{Algorithm}
\floatname{algorithm}{\protect\algorithmname}

\usepackage{multicol}
\usepackage{times}
\usepackage{epsfig}
\usepackage[center]{subfigure}
\usepackage{color,soul}
\usepackage[font=small, labelsep=period]{caption}
\usepackage{algorithmic}
\usepackage{hyperref}

\IEEEcompsoctitleabstractindextext{%
\begin{abstract}
Correlation filters (CFs) are a class of classifiers that are attractive for object localization and tracking applications. Traditionally, CFs have been designed in the frequency domain using the discrete Fourier transform (DFT), where correlation is efficiently implemented. However, existing CF designs do not account for the fact that the multiplication of two DFTs in the frequency domain corresponds to a circular correlation in the time/spatial domain. Because this was previously unaccounted for, prior CF designs are not truly optimal, as their optimization criteria do not accurately quantify their optimization intention. In this paper, we introduce new zero-aliasing constraints that completely eliminate this aliasing problem by ensuring that the optimization criterion for a given CF corresponds to a linear correlation rather than a circular correlation. This means that previous CF designs can be significantly improved by this reformulation. We demonstrate the benefits of this new CF design approach with several important CFs. We present experimental results on diverse data sets and present solutions to the computational challenges associated with computing these CFs. Code for the CFs described in this paper and their respective zero-aliasing versions is available at \color{red}{\href{http://vishnu.boddeti.net/projects/correlation-filters.html}{http://vishnu.boddeti.net/projects/correlation-filters.html}}
\end{abstract}
}

\begin{document}

\title{Zero-Aliasing Correlation Filters for Object Recognition}

\author{Joseph A. Fernandez, \emph{Student Member, IEEE}, Vishnu Naresh Boddeti, \emph{Member, IEEE}, Andres Rodriguez, \emph{Member, IEEE}, B. V. K. Vijaya Kumar, \emph{Fellow, IEEE}%
\thanks{This work was supported in part by the Air Force Office
of Scientific Research (AFOSR), the National Defense Science \& Engineering
Graduate Fellowship (NDSEG) Program, and KACST, Saudi Arabia.

J. A. Fernandez and B. V. K. Vijaya Kumar are with the Department
of Electrical and Computer Engineering, and V. N. Boddeti is with
the Robotics Institute, Carnegie Mellon University (e-mail: jafernan@andrew.cmu.edu,
kumar@ece.cmu.edu, naresh@cmu.edu).

A. Rodriguez is with Air Force Research Laboratory, Dayton, Ohio (e-mail:
andres.rodriguez.8@us.af.mil).%
}}

\markboth{IEEE Transactions on Pattern Analysis and Machine Intelligence}{}

\maketitle
\pagenumbering{gobble}
\renewcommand{\figurename}{Fig.}

\section{Introduction}

\PARstart{C}{orrelation} filters (CFs) are a useful tool for a variety of tasks in signal processing and pattern recognition, such as biometric recognition \cite{Kumar_CF4_face_recognition,Thornton_PAMI}, object alignment \cite{Boddeti_VCF}, action recognition \cite{MDRodriguez:2008}, object detection \cite{MMCF, henriques2013beyond, boddeti2014maximum}, object tracking \cite{henriques2012exploiting, MOSSE} and event retrieval from videos \cite{revaud2013event}. CFs possess many attractive properties that make them useful for these tasks. Most importantly, CFs are capable of performing both classification and localization simultaneously. In other words, CFs do not assume prior segmentation of objects in test scenes, and are capable of detecting and classifying multiple objects in a single scene simultaneously. CFs are shift-invariant, which means that the objects to be recognized do not have to be centered in the test scene as CFs produce correlation peaks at locations corresponding to the object location in the test scene. These properties make CFs attractive for different applications.

There are two stages in the use of CFs. First, a CF template (a template refers to a 2D array in spatial domain) is designed from a set of training images (we will refer to images and their 2-D Fourier transforms in this paper, but the concepts and methods are equally valid for 1-D temporal signals and higher-dimensional signals such as image sequences). This template is designed such that cross-correlating it with centered positive training images leads to correlation outputs with sharp peaks at the origin and cross-correlating it with negative training images leads to, either no such discernible peaks or negative peaks. Once designed, the CF is then applied to the test scene, which may be larger than the template itself. The output correlation plane is then searched for positive peaks to determine if any objects from the positive class are detected within the test image. The location of the correlation peak denotes the location of the target; the sharpness of the peak, i.e., how large the peak value is compared to surrounding values, may serve as a measure of confidence regarding the existence of the target.

\begin{figure}[t]
\begin{center}
\subfigure[Input Signal]{ \includegraphics[width=0.32\columnwidth]{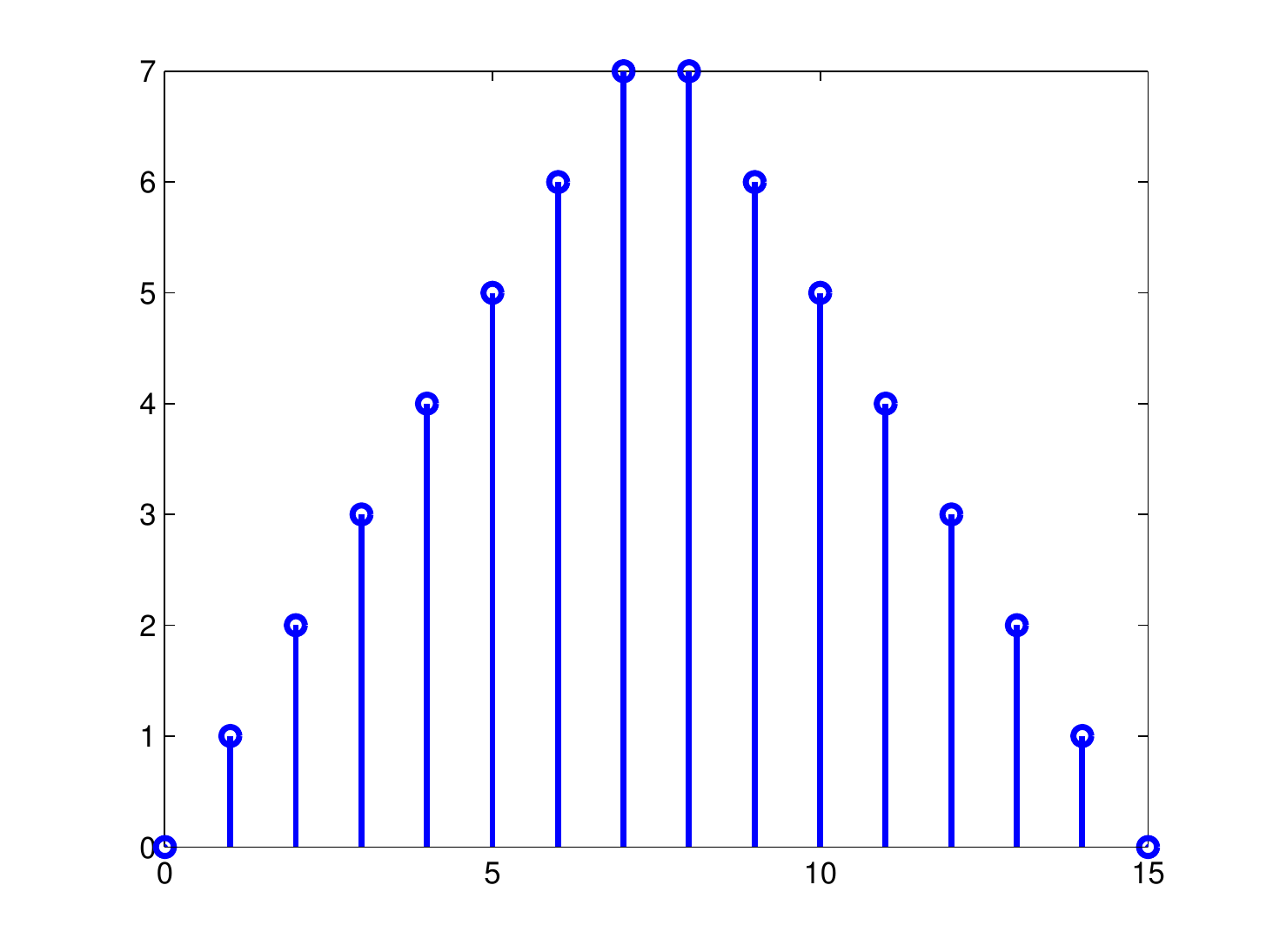}}
\subfigure[Linear Correlation]{\includegraphics[width=0.32\columnwidth]{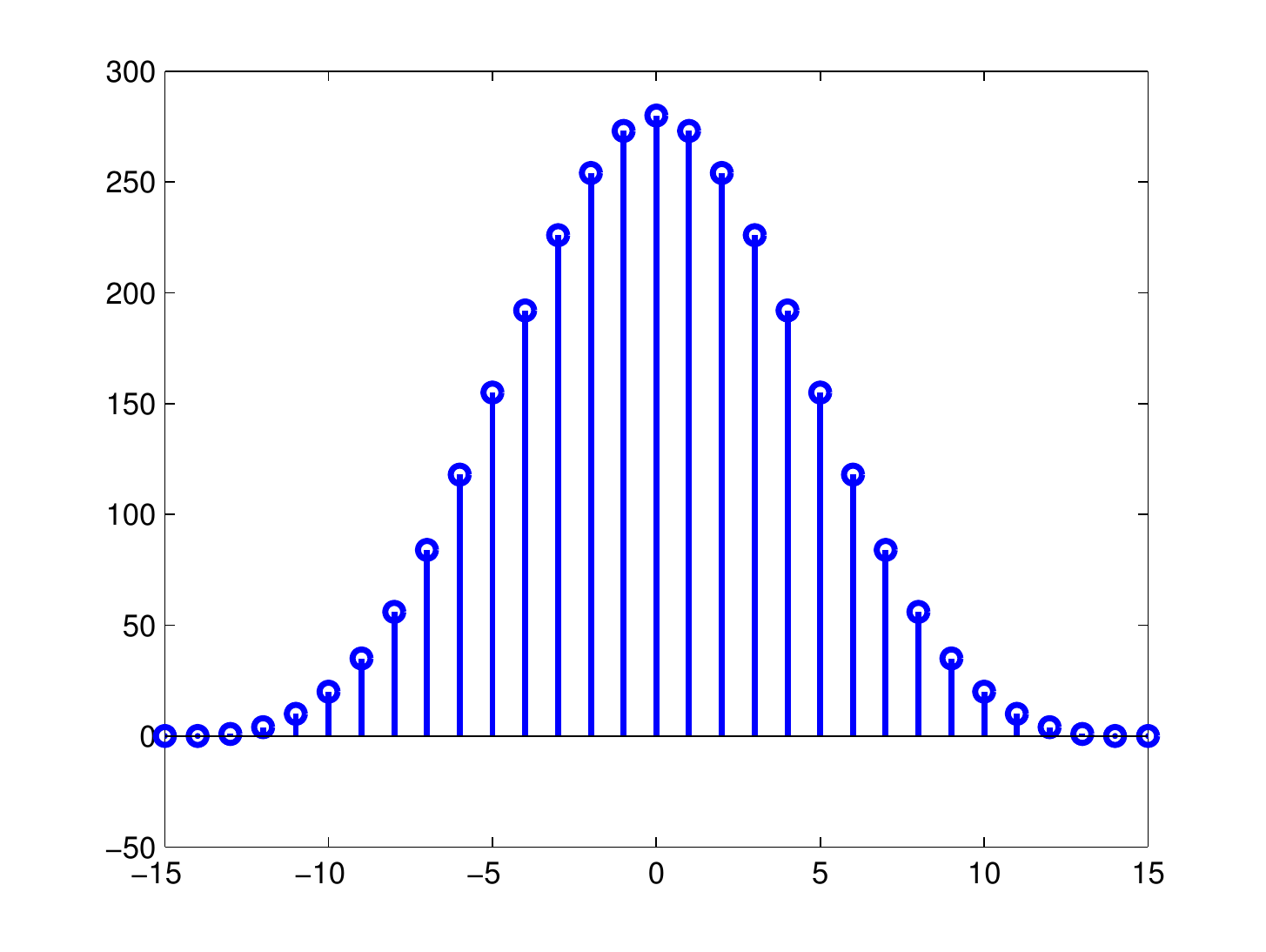}}
\subfigure[Circular Correlation]{\includegraphics[width=0.32\columnwidth]{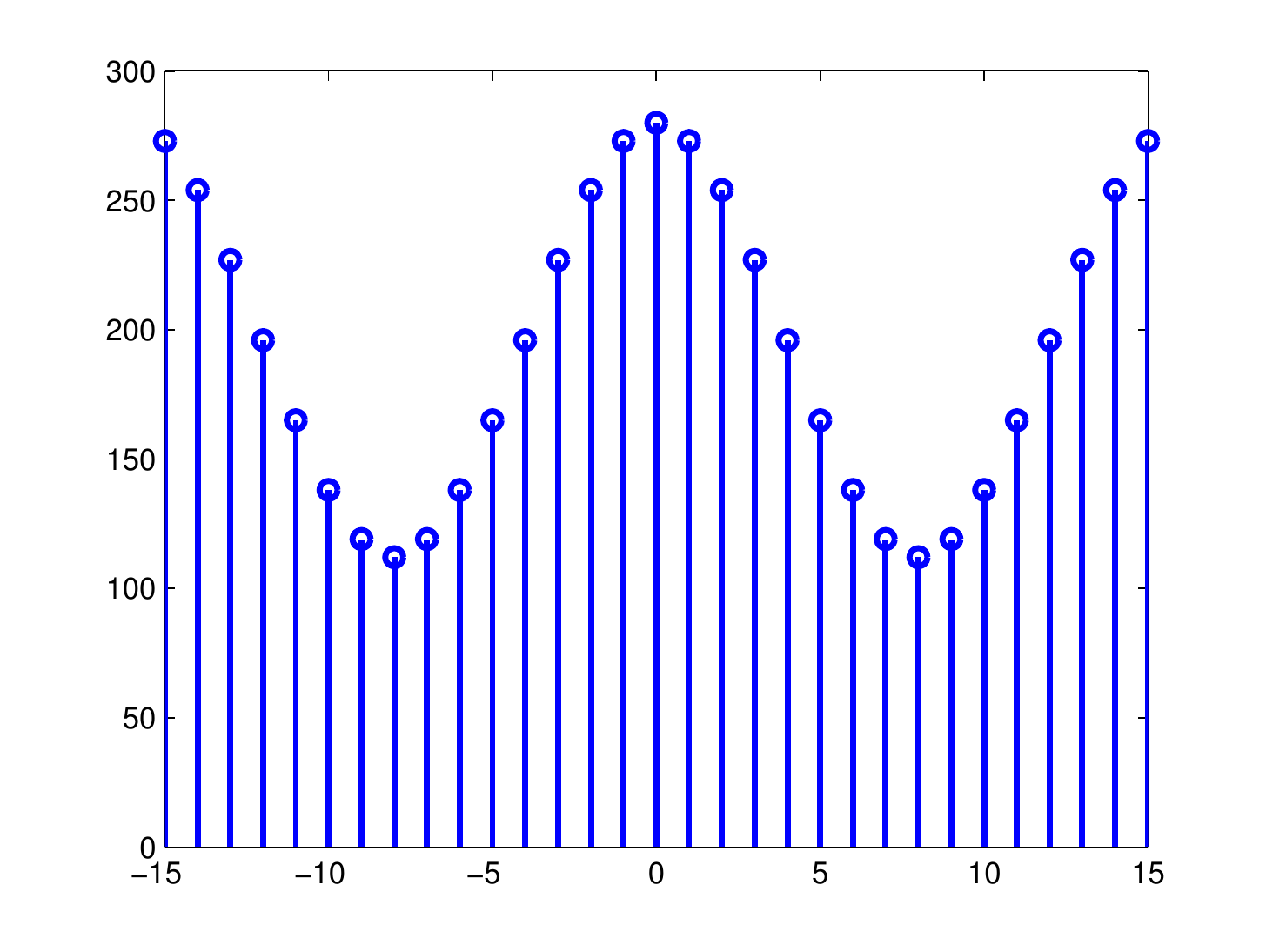}}
\protect\caption{\label{fig:circ_corr_example}An example illustrating the differences between linear and circular correlation. Here, the input signal in
(a) is correlated with itself by appropriately padding the input signal with zeros. The result is the linear correlation output (shown in (b)). This is accomplished with a DFT of size $N=2\times16-1=31$. The circular correlation output (in part (c)) is generated with a DFT of size $16$. Because the signals are not zero-padded, the output in part (c) is a circular correlation. In this example, the effects of aliasing due to circular correlation are very obvious; in other examples, the effects may be less noticeable, but still can affect performance.}
\end{center}
\end{figure}

CFs have traditionally been designed in the spatial frequency domain for computational efficiency. Typically, various metrics are optimized during filter design. One example is minimizing the mean squared error (MSE) between a desired correlation output and the actual correlation output. In formulating these optimizations, cross-correlation in spatial domain is expressed in the frequency domain as the element-wise multiplication of the CF and the conjugate of the Discrete Fourier Transform (DFT) of the training image. Cross-correlating two images of size $N\times M$, using fast Fourier transform (FFT) algorithms (for obtaining 2-D DFTs) requires on the order of $NM\log(NM)$ multiplications whereas a direct spatial-domain correlation requires on the order of $N^{2}M^{2}$ multiplications. However, this DFT-based correlation operation produces a circular correlation rather than a linear correlation \cite{DSP_book}. The difference between a linear correlation and a circular correlation can be significant (see Fig. \ref{fig:circ_corr_example}). Circular correlation results in aliasing, i.e., parts of linear correlation being added to other parts of itself. This aliasing directly influences the localization performance of the CF. When the CF is \emph{applied} to a test image, aliasing from circular correlation in that operation can be avoided by appropriately padding the template and test image with zeros prior to computing the DFTs. However, simply zero-padding the images does not resolve the aliasing problem in the CF \emph{design} stage. In the past, this fact was mostly ignored, and the optimization needed for CF design was carried out with the incorrect assumption (as we will show in this work) that the circular correlation in CF designs was roughly the same as the linear correlation.

This inconsistency between the CF design and CF use exists not only in the well-known Minimum Average Correlation Energy (MACE) filter
\cite{MACE}, but also in many other CF designs, including (but not limited to) the designs presented in \cite{MMCF,MOSSE,ASEF,UOTSDF,GMACE,Boddeti_VCF, henriques2013beyond,boddeti2014maximum}, and \cite{OTSDF1-Refregier:90}. As we will show in this paper, significantly improved object classification and localization results can be obtained in all CF approaches by removing the aliasing effects due to circular correlation.
The problem of circular correlation in CF designs has been mostly ignored for many years; in this paper, we present a solution to this problem. This solution is a \textit{fundamental} and \textit{significant} advancement to the design of CFs. Specifically, we introduce what we term \textit{zero-aliasing (ZA) constraints}, which force the tail of the CF template to zero, which means that the corresponding optimization metric will correspond to a linear correlation rather than a circular correlation. These ZA constraints yield zero-aliasing correlation filters (ZACFs) that are consistent with the original design criteria of existing filter designs. This means that the correlation outputs resemble their intended design. As a result, the performance of ZACFs is significantly improved over conventional CFs. While CFs may not be the best solution for all object recognition problems (e.g., in some problems, the test objects may be centered, thus not benefiting from the CF's localization capabilities), our ZACF approach does offer improvements over existing CF designs that span over multiple decades.

A high level overview of our approach is shown in Fig. \ref{fig:Overview}. We introduced ZA constraints for only the MACE filter in \cite{FernandezISPA13}; in this paper, we extend our work on this subject. The contributions of this paper are as follows:
\begin{itemize}
\item Extension of ZA constraints to major classes of CF designs (for both scalar and vector features representations),
\item Reduced-aliasing correlation filters (RACFs), which offer a computationally more tractable closed-form solution to the problem, while still allowing some aliasing,
\item Fast, efficient and scalable iterative proximal gradient descent based approach for numerically solving ZACF designs,
\item A computational and performance comparison of the different methods,
\item Numerical results on a variety of datasets demonstrating the superior performance of ZACFs.
\end{itemize}
Although the main focus of our paper is CFs, we note that many other problems (e.g., convolutional sparse coding \cite{Lucey_fast_conv_sparse_coding}, convolutional neural networks \cite{mathieu2013fast}) which are dependent on convolutions or correlations and are solved in the Fourier domain can potentially benefit from our general observations and the addition of ZA constraints to their respective optimization formulations.    
\tikzstyle{format} = [draw, thin, fill=blue!20]
\tikzstyle{medium} = [ellipse, , draw, thin, fill=green!20, minimum
height=2.0em]
\tikzstyle{decision} = [diamond, draw, fill=blue!20,
text width=4.5em, text badly centered, node distance=2cm, inner sep=0pt]
\tikzstyle{block} = [rectangle, draw, fill=blue!20,
text width=4.0em, text centered, rounded corners, minimum height=2em]
\tikzstyle{line} = [draw, -latex']
\tikzstyle{cloud} = [draw, ellipse,fill=red!20, node
distance=2cm,minimum height=1.5em]
\tikzstyle{circ} = [draw, circle, fill=red!20, node
distance=2cm]
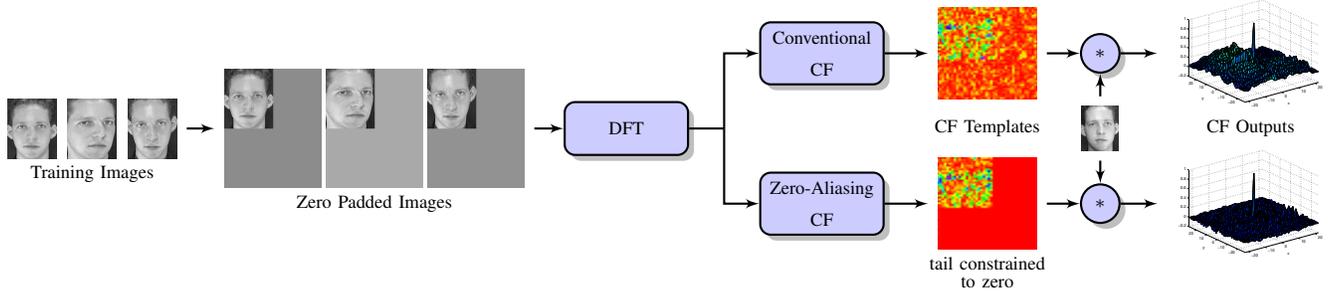
\begin{figure*}[t]
\centering
\begin{tikzpicture}[node distance=3cm, auto,>=latex', thick]
\node[] (x1) {\includegraphics[scale=0.2]{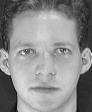}};
\node[right of=x1,node distance=0.8cm] (x2) {\includegraphics[scale=0.2]{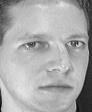}};
\node[right of=x2,node distance=0.8cm] (x3) {\includegraphics[scale=0.2]{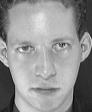}};

\node[right of=x3,node distance=1.6cm] (y1) {\includegraphics[scale=0.2]{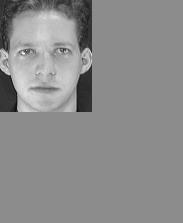}};
\node[right of=y1,node distance=1.35cm] (y2) {\includegraphics[scale=0.2]{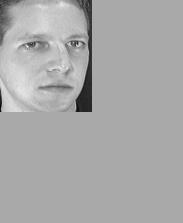}};
\node[right of=y2,node distance=1.35cm] (y3) {\includegraphics[scale=0.2]{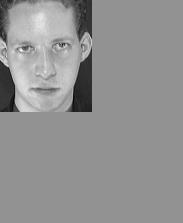}};

\node[block, right of=y3, node distance=2cm, drop shadow] (dft) {{\scriptsize DFT}};
\coordinate [right of=dft, node distance=1.3cm] (d1) {};
\coordinate [right of=d1, node distance=1.3cm] (d2) {};
\node[block, above of=d2, node distance=1cm, drop shadow] (cf1) {{\scriptsize Conventional CF}};
\node[block, below of=d2, node distance=1cm, drop shadow] (cf2) {{\scriptsize Zero-Aliasing CF}};
\node[right of=cf1, node distance=2.2cm] (cf11) {\includegraphics[scale=0.15,trim=2cm 2cm 4cm 1cm,clip]{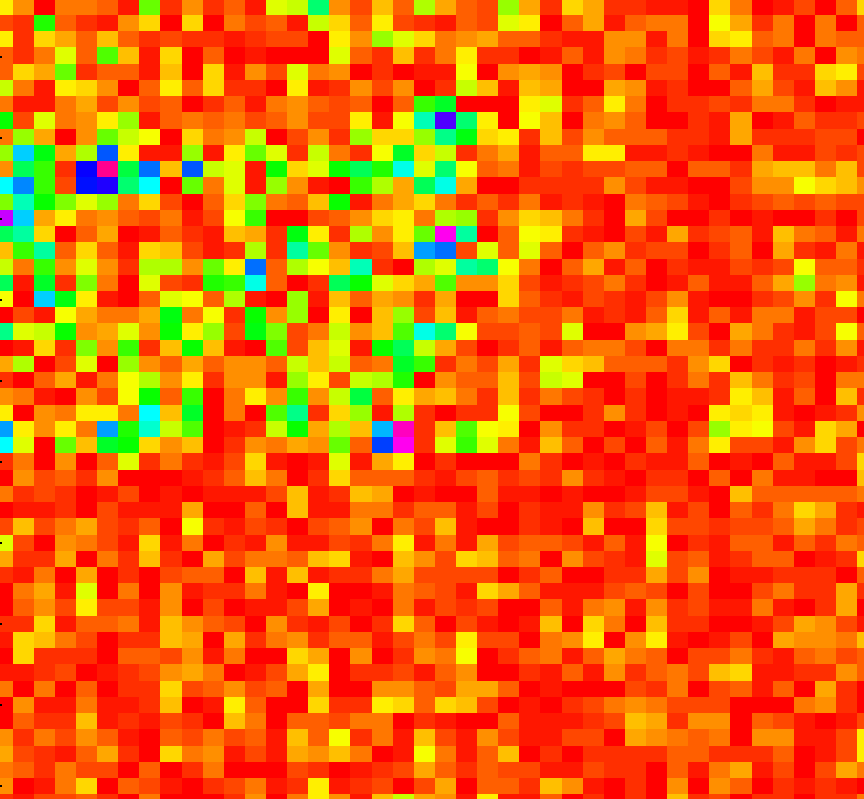}};
\node[right of=cf2, node distance=2.2cm] (cf22) {\includegraphics[scale=0.15,trim=2cm 2cm 4cm 1cm,clip]{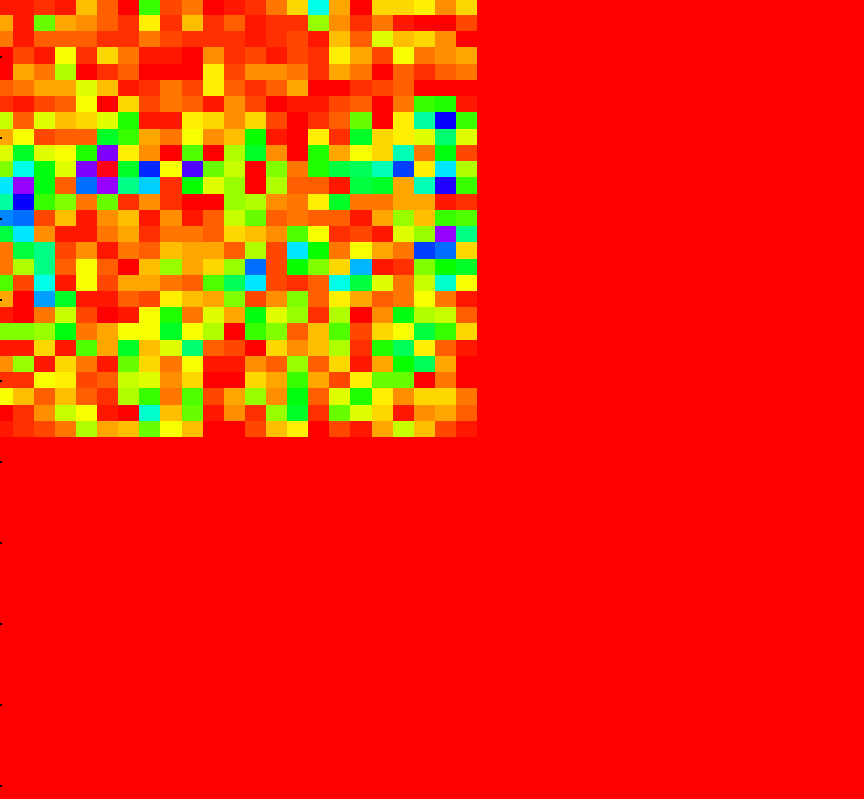}};
\node[circ, right of=cf11, node distance=1.5cm, fill=blue!20, drop shadow] (c1) {{\scriptsize $\ast$}};
\node[circ, right of=cf22, node distance=1.5cm, fill=blue!20, drop shadow] (c2) {{\scriptsize $\ast$}};
\node[right of=c1, node distance=2.0cm] (p1) {\includegraphics[scale=0.15]{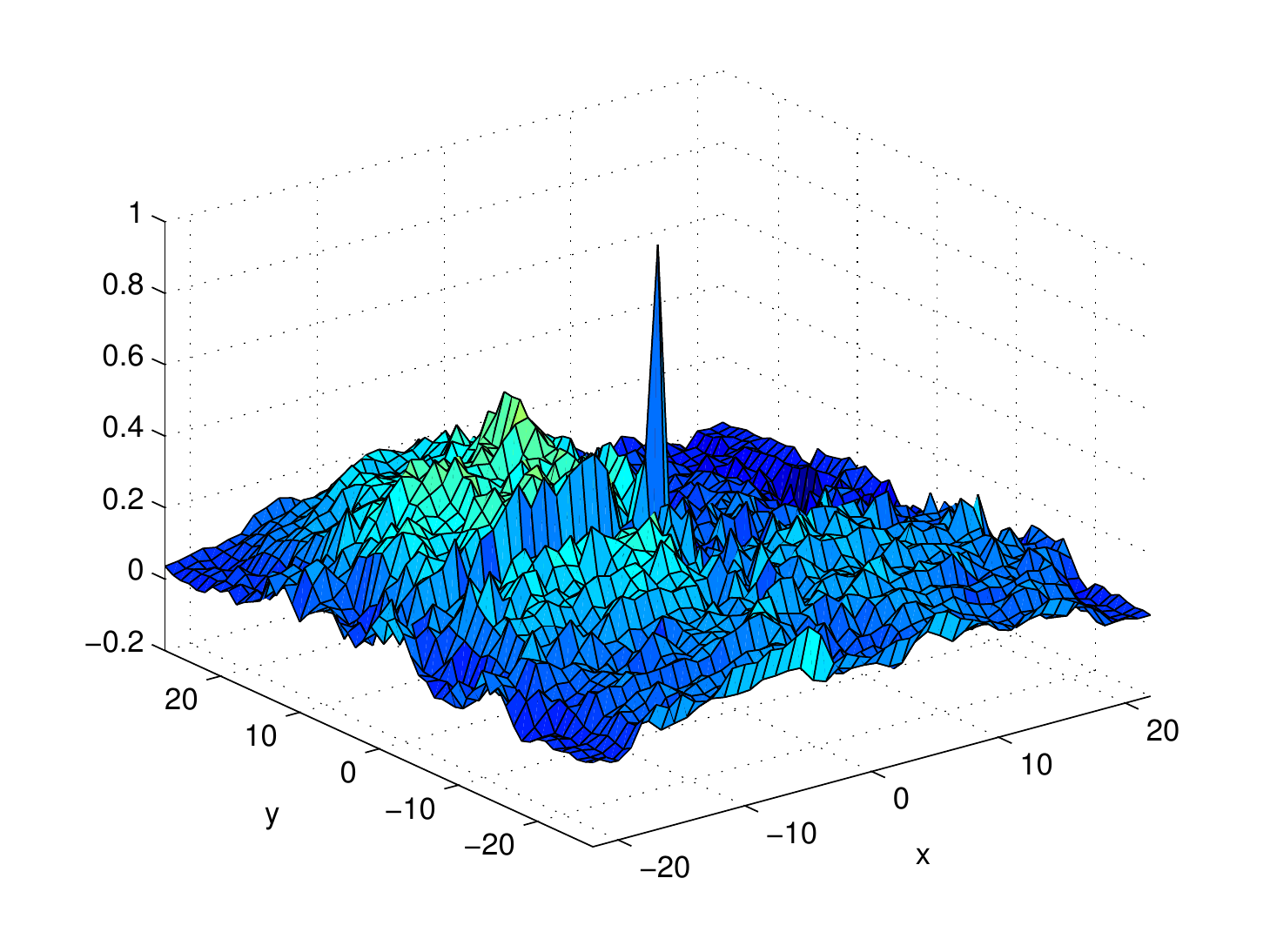}};
\node[right of=c2, node distance=2.0cm] (p2) {\includegraphics[scale=0.15]{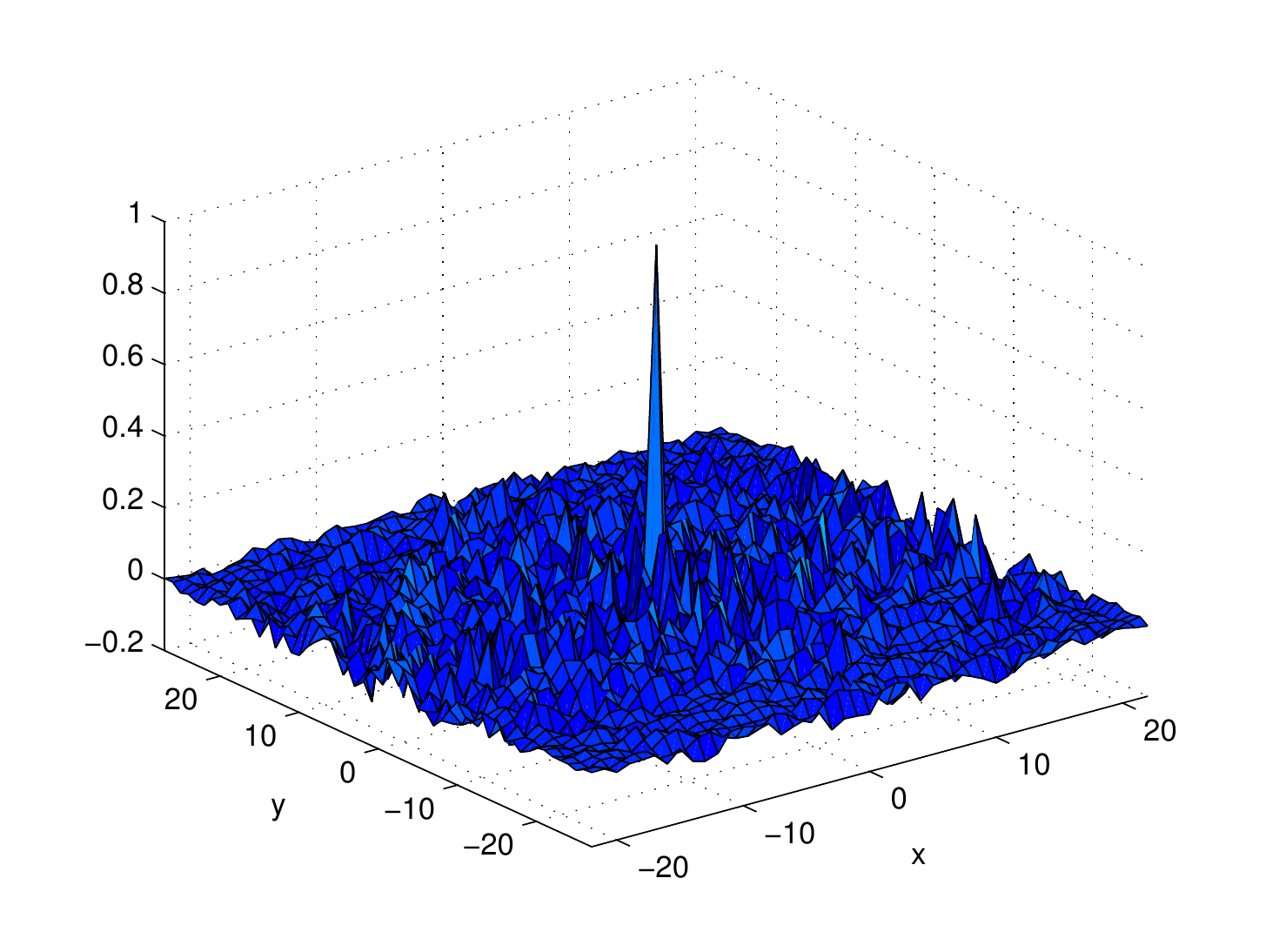}};
\node[below of=c1, node distance=1.0cm] (r1) {\includegraphics[scale=0.15]{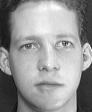}};

\node[below of=x2, node distance=0.6cm] (t1) {{\scriptsize Training Images}};
\node[below of=y2, node distance=1cm] (t1) {{\scriptsize Zero Padded Images}};
\node[below of=p1, node distance=1cm] (r2) {{\scriptsize CF Outputs}};
\node[below of=cf11, node distance=1cm] (r3) {{\scriptsize CF Templates}};
\node[below of=cf22, node distance=0.8cm] (r3) {{\scriptsize tail constrained}};
\node[below of=cf22, node distance=1.05cm] (r3) {{\scriptsize to zero}};

\path[line] (x3) -- (y1);
\path[line] (y3) -- (dft);
\draw [-] (dft) -- (d1);
\path[line] (d1) |- (cf1);
\path[line] (d1) |- (cf2);
\path[line] (cf1) -- (cf11);
\path[line] (cf2) -- (cf22);
\path[line] (cf11) -- (c1);
\path[line] (cf22) -- (c2);
\path[line] (c1) -- (p1);
\path[line] (c2) -- (p2);
\path[line] (r1) -- (c1);
\path[line] (r1) -- (c2);

\end{tikzpicture}
\caption{\label{fig:Overview}Overview of our proposed zero-aliasing correlation filters (ZACFs) approach. Conventional CF designs result in templates that are non-zero for all values. This means that, during the optimization, the correlation between the template and the training images is in fact a circular correlation. In our design, constraining the tail of the template using ZA constraints guarantees that the optimization step corresponds to a linear correlation. This results in correlation planes that resemble the original design criteria--in this case, a sharp peak with low side-lobes.}
\end{figure*}

This paper is organized as follows. In Section \ref{sec:Related-Work}, we summarize related literature. We illustrate the details of circular correlation issue using the well-known MACE filter as an example in Section \ref{sec:The-Circular-Correlation-Problem}. In Section \ref{sec:Zero-Aliasing-Correlation-Filter-Formulations}, we show how ZA constraints may be added to several other popular CF designs, namely the Optimal Tradeoff Synthetic Discriminant Function (OTSDF) filter \cite{OTSDF1-Refregier:90}; the Minimum Output Sum of Squared Errors (MOSSE) filter \cite{MOSSE,Boddeti_VCF}, and the Maximum-Margin Correlation Filter (MMCF) \cite{MMCF,boddeti2014maximum}. We discuss only a representative sample of CFs, but this approach is applicable to many more CF designs. We discuss some methods to efficiently solve for the ZACF designs in Section \ref{sec:Computational-Considerations}. In Section \ref{sec:Experimental-Results}, we present experimental results for a variety of applications, including face recognition, Automatic Target Recognition (ATR), eye localization and object detection. We conclude in Section \ref{sec:Conclusions}.

\section{\label{sec:Related-Work}Related Work}
There are a wide variety of applications for CFs in biometrics, object detection, landmark detection, and action recognition. For example, CFs are used for face recognition in \cite{Kumar_CF4_face_recognition}. A deformable pattern matching method is coupled with CFs in \cite{Thornton_PAMI} to recognize human iris images. More recently, ocular recognition has become of greater interest to the biometric community, and CFs have been applied to this problem in \cite{boddeti2011comparative}. CFs have also been suggested for biometric key-binding in \cite{Bodetti_PAMI_key_binding}. Phase-only correlation methods have been used to correlate the phase information in two images. Biometric applications of these techniques include iris recognition \cite{Miyazawa_PhaseOnlyCorr_IrisPAMI}, as well as face, fingerprint, palmprint, and finger-knuckle recognition \cite{nakajima2004fingerprint,itophase}.

CFs have been used for ATR \cite{MMCF} and to track moving objects in video \cite{MFCF,MOSSE,henriques2012exploiting}. Pedestrian detection has been demonstrated in \cite{Bolme_ped_detector,henriques2013beyond}. CFs have been also used as local part detectors. For example, \cite{MMCF,ASEF} uses CFs for eye localization in images of human faces, and \cite{Boddeti_VCF} uses CFs as the appearance model for landmark localization in object alignment. CFs have also proven useful for detecting and locating text strings for document processing \cite{Li_CFtext_detectionPAMI}. In \cite{revaud2013event} CFs have been used to accurately localize video clips in time thereby dramatically improving (computationally as well as performance wise) video search and retrieval from a large video corpus. Recently \cite{movshovitz3d} proposed a method to discriminatively learn bases of exemplar correlation filters for accurate 3D pose estimation.

CFs have also been extended to videos and have been applied to detect and classify human actions \cite{MDRodriguez:2008,Ali_and_Lucey:2010,Mahalanobis:11,ulHaq2011_DCCF_AR}. In \cite{Mahalanobis:11}, a motion model is combined with two-dimensional CFs to classify human activities. The use of CFs is not limited to pixels; they have been applied to work with feature vectors such as optical flow \cite{MDRodriguez:2008}, histogram of oriented gradient (HOG), or scale invariant feature transform (SIFT) features \cite{Boddeti_VCF,MCCF}.

Despite all the work in CFs, the circular correlation problem has been mostly ignored. This problem was initially observed in \cite{SMACE}, which proposed reformulating the popular MACE filter \cite{MACE} in the spatial domain to avoid circular correlation. However, computation of the MACE filter in the spatial domain is of much higher complexity than computation via the frequency domain. Furthermore, there was never a proposed frequency domain solution to handle the circular correlation problem. More recently, \cite{Rodriguez_CircCorr13} explored several methods to reduce circular correlation effects. These methods use zero padding and/or windowed training images to reduce the effects of circular correlation. One of these methods, originally mentioned in \cite{ASEF}, multiplies the training images by a cosine window to reduce edge effects (this approach is explored in greater detail in \cite{Rodriguez_CircCorr13}). Windowing is undesirable, however, because it fundamentally changes the content of the training images. All of the methods discussed in \cite{Rodriguez_CircCorr13} fall short, as they do not eliminate circular correlation effects.

In practice, CFs are typically designed by zero-padding training images (of dimension $N\times M$) to a size of at least $(2N-1)\times(2M-1)$ prior to taking the DFT since zero-padding the input images will result in linear correlation. Although this is true in the testing stage, it is not true in the training stage, and as we will show in Section \ref{sec:The-Circular-Correlation-Problem}, it does not solve the aliasing problem and results in a suboptimal CF design. In \cite{FernandezISPA13}, we introduced a new way to completely eliminate aliasing--namely that the CF template must be explicitly constrained during the CF design stage such that the template tail is set to zero. By coupling these ZA constraints with zero-padding the training images, the optimization metric corresponds to a linear correlation, and the resulting CF template is the same as the template we would get if we were to design the CF in the spatial domain. In the next sections, we describe the problem and our solution in detail.

\section{\label{sec:The-Circular-Correlation-Problem}The Circular Correlation Problem}

In this section, we first introduce the main idea that is common to most CF designs and then discuss the circular correlation problem in these designs using the MACE filter \cite{MACE} as an illustrative example. For notational ease all the expressions through the rest of this paper are given for 1-D signals with $K$ channels. Most prior CF designs were aimed at images (i.e., gray scale values of pixels), but our formulations can accommodate images as well as vector features (e.g., HOG, SIFT features) extracted from the images.

Vectors are denoted by lower-case bold ($\mathbf{x}$) and matrices in upper-case bold ($\mathbf{X}$). $\mathbf{\hat{x}} \leftarrow \mathcal{F}_K(\mathbf{x})$ and $\mathbf{x} \leftarrow \mathcal{F}^{-1}_K(\mathbf{\hat{x}})$ denotes the Fourier transform of $\mathbf{x}$ and the inverse Fourier transform of $\mathbf{\hat{x}}$, respectively, where $\mbox{ }\hat{}\mbox{ }$ denotes variables in the frequency domain, $\mathcal{F}_K()$ is the Fourier transform operator and $\mathcal{F}^{-1}_K()$ is the inverse Fourier transform operator with the operators acting on each of the $K$ channels independently. Superscript $^{\dagger}$ denotes the complex conjugate transpose operation.

\subsection{\label{sub:Correlation-Filter-Design}Correlation Filter Design}
Let us assume that we train a filter using $L$ training signals of size $N$. Typically, we train the filter $\mathbf{h}$ in the frequency domain using DFTs (of size $N_{\mathcal{F}}=N$) of the training signals, $\mathbf{\hat{x}}_{l}$, for $l=1,\ldots,L$. Many CF designs can be interpreted as a regression problem optimizing the \emph{localization loss} defined as the MSE between an ideal desired correlation output $\mathbf{g}_l$ for an input signal and the correlation output of the training signals with the template\footnote{We refer to the CF as a ``template'' in the time or spatial domain, and as a ``filter'' in the frequency domain.} (see \cite{Boddeti_VCF} for details and see \cite{henriques2013beyond} for a circulant decomposition based interpretation and motivation for CF design formulations),
\begin{eqnarray}
\label{eq:mse_localization_loss}
J(\mathbf{h}) &=& \frac{1}{L}\sum_{l=1}^L\left\|\sum_{k=1}^K\mathbf{x}^{k}_{l}\ast\mathbf{h}^{k}-\mathbf{g}_{l}\right\|_2^2 \\
&=& \frac{1}{N_{\mathcal{F}}L}\sum_{l=1}^L\left\|\sum_{k=1}^K\mathbf{\hat{X}}^{k\dagger}_l\mathbf{\hat{h}}^k - \mathbf{\hat{g}}_l\right\|^2_2 \nonumber \\
&=& \mathbf{\hat{h}^{\dagger}\hat{D}\hat{h}} - 2\mathbf{\hat{h}^{\dagger}\hat{p}} + E_{g} \nonumber	
\end{eqnarray}
\noindent where $\ast$ denotes the correlation operation, we use the Parseval's theorem to express the MSE in the Fourier domain. $E_{g}$ is a constant which depends on the desired correlation response, $\mathbf{\hat{X}}_{l}^{k\dagger}\mathbf{\hat{h}}^{k}$ is the DFT of the correlation of the $k$-th channel of the $l$-th training signal with the corresponding $k$-th channel of the CF template where the diagonal matrix $\mathbf{\hat{X}}^{k}_{l}$ contains the vector $\mathbf{\hat{x}}^{k}_{l}$ along its diagonal and,
\begin{equation}
\label{eq:define_d_p}
\mathbf{\hat{D}}=\frac{1}{N_{\mathcal{F}}L}\left[\begin{array}{ccc}
\sum_{l=1}^L\mathbf{\hat{X}}_{l}^{1}\mathbf{\hat{X}}_{l}^{1\dagger} & \cdots & \sum_{l=1}^L\mathbf{\hat{X}}_{l}^{1}\mathbf{\hat{X}}_{l}^{K\dagger}\\
\vdots & \ddots & \vdots\\
\sum_{l=1}^L\mathbf{\hat{X}}_{l}^{K}\mathbf{\hat{X}}_{l}^{1\dagger} & \cdots & \sum_{l=1}^L\mathbf{\hat{X}}_{l}^{K}\mathbf{\hat{X}}_{l}^{K\dagger}
\end{array}\right]
\end{equation}
\begin{equation}
\mathbf{\hat{p}}=\frac{1}{N_{\mathcal{F}}L}\left[\begin{array}{c}
\sum_{l=1}^L\mathbf{\hat{X}}^1_l\mathbf{\hat{g}}_l\\
\vdots\\
\sum_{l=1}^L\mathbf{\hat{X}}^K_l\mathbf{\hat{g}}_l\\
\end{array}\right],\:
\mathbf{\hat{h}}=\left[\begin{array}{c}
\mathbf{\hat{h}}^1\\
\vdots\\
\mathbf{\hat{h}}^{K}
\end{array}\right],\:
\mathbf{\hat{x}}_l=\left[\begin{array}{c}
\mathbf{\hat{x}}^1_l\\
\vdots\\
\mathbf{\hat{x}}^{K}_l
\end{array}\right] \nonumber
\end{equation}

In addition to minimizing the \emph{localization loss} in Eq. \ref{eq:mse_localization_loss}, some CFs are constrained to produce pre-specified values at the origin (known as correlation peak constraints) in response to centered training signals (this is equivalent to constraining the dot product of the training signals and the template).
\begin{eqnarray}
\label{eq:fdmace_constraints}
\min_{\mathbf{\hat{h}}} && \mathbf{\hat{h}^{\dagger}\hat{D}\hat{h}} - 2\mathbf{\hat{h}^{\dagger}\hat{p}} \\ 
s.t. && \mathbf{\hat{X}}^{\dagger}\mathbf{\hat{h}}=\mathbf{u} \nonumber
\end{eqnarray}
In this case, $\mathbf{\hat{X}}$ (with no subscripts) is an $N_{\mathcal{F}}K\times L$ matrix with columns $\mathbf{\hat{x}}_{l}$, and $\mathbf{u}$ is the correlation peak constraint vector, whose elements are usually set to $1$ for positive class training signals and to $0$ for negative class training signals (if used). This optimization problem results in a closed form solution for the filter.
\begin{eqnarray}
\label{eq:new_mace}
\mathbf{\hat{h}} &=& \mathbf{\hat{D}}^{-1}\mathbf{\hat{X}}\left(\mathbf{\hat{X}}^{\dagger}\mathbf{\hat{D}}^{-1}\mathbf{\hat{X}}\right)^{-1}\mathbf{u} + \\
&& \left(\mathbf{I}-\mathbf{\hat{D}}^{-1}\mathbf{\hat{X}}(\mathbf{\hat{X}}^{\dagger}\mathbf{\hat{D}}^{-1}\mathbf{\hat{X}})^{-1}\mathbf{\hat{X}}^{\dagger}\right)\mathbf{\hat{D}}^{-1}\mathbf{\hat{p}} \nonumber
\end{eqnarray}
The conventional MACE filter \cite{MACE} is designed to minimize the average correlation energy (ACE, defined below)
\begin{equation}
\label{eq:FDMACE-ACE-final}
\mbox{ACE}=\frac{1}{N_{\mathcal{F}}L}\sum_{l=1}^{L}\sum_{k=1}^K\mathbf{\hat{h}}^{k\dagger}\mathbf{\hat{X}}^{k}_{l}\mathbf{\hat{X}}_{l}^{k\dagger}\mathbf{\hat{h}}^{k}
\end{equation}
\noindent of the entire correlation output (this is equivalent to setting the desired correlation output $\mathbf{g}_l$ to an all-zero plane and equivalently setting $\mathbf{\hat{p}}$ to an all-zero vector) resulting in the following conventional closed form solution for the MACE filter. 
\begin{equation}
\label{eq:FDMACE}
\mathbf{\hat{h}}=\mathbf{\hat{D}}^{-1}\mathbf{\hat{X}}\left(\mathbf{\hat{X}}^{\dagger}\mathbf{\hat{D}}^{-1}\mathbf{\hat{X}}\right)^{-1}\mathbf{u}
\end{equation}
This formulation results in correlation outputs exhibiting sharp correlation peaks with low side-lobes, leading to good localization.

However, there is a fundamental problem with this formulation. The term $\mathbf{\hat{X}}_{l}^{k\dagger}\mathbf{\hat{h}}^{k}$ in the ACE expression corresponds to an element-wise multiplication of two DFTs. This term is precisely the problem with the original MACE formulation. When two DFTs are multiplied together in the frequency domain, the result corresponds to a circular correlation in the spatial domain. As shown in Fig.\ref{fig:circ_corr_example}, circular correlation is an aliased version of the desired linear correlation. To attempt to compensate for the effects of aliasing, the MACE filter is sometimes trained \cite{Rodriguez_CircCorr13} using zero-padded training signals, such that the DFT size is $N_{\mathcal{F}}=(2N-1)$. Such zero-padding of training signals does not solve the circular correlation problem. This is because, even when training signals are zero-padded, the MACE template obtained using Eq. \ref{eq:FDMACE} is usually nonzero for the full template of size $N_{\mathcal{F}}$. Therefore, the multiplication of $\mathbf{\hat{h}}$ and the vectorized DFT of each training signal always results in a circular correlation, regardless of the number of zeros the training signals are padded with. For the term $\mathbf{\hat{X}}_{l}^{k\dagger}\mathbf{\hat{h}}^{k}$ to represent a linear correlation $\mathbf{\hat{h}}^{k}$ must be constrained such that the tail of the spatial domain template $h^k(n)$ is forced to zero. We recently introduced the zero-aliasing MACE (ZAMACE) filter design in \cite{FernandezISPA13} to eliminate this aliasing problem. The mathematical details are presented in Section \ref{sub:ZAMACE} and extensions to other filter types are presented in Sections \ref{sub:ZAMOSSE} and \ref{sub:ZAMMCF}.

\subsection{Comparison of MACE and TDMACE Filters}
We now explain the intuition and discuss several experiments to illustrate the benefits of our approach using the ZAMACE filter as an example. As a benchmark, we compute the time-domain MACE (TDMACE), which is a template computed in the time domain and is free of any aliasing (for a derivation of this in 1-D, see \cite{FernandezISPA13}; a generalized spatial-domain MACE template can be found in \cite{SMACE}; we use the terms ``time domain'' and ``spatial domain'' interchangeably). As mentioned previously, computing the MACE filter (or any other CF) in the spatial domain is computationally expensive.

We begin with a 1-D example that uses ECG signals from the MIT-BIH Arrhythmia Database \cite{MIT-BIH}. In this case, we have extracted $355$ heartbeat cycles. Each signal is $N=301$ samples long and represents a single heartbeat, which has been segmented by a cardiologist expert based on the location of the main peak.

We train MACE, TDMACE, and ZAMACE filters using a set of $L=10$ training signals. Here, we design CFs using training signals from only one class. We desire a correlation template of length $N$ (the same size as the training samples). As mentioned previously, training signals are typically zero-padded to size $N_{\mathcal{F}}\geq2N-1$ prior to taking the DFT. However, although the training signals are now zero padded, the conventional MACE filter design does nothing to ensure that the resulting template is zero in its tail. In contrast, our ZA approach (described more formally in Section \ref{sec:Zero-Aliasing-Correlation-Filter-Formulations}) constrains the optimization of the filter $\mathbf{\hat{h}}$ such that it is zero for the last $N_{\mathcal{F}}-N$ indices of the template $h(n)$. When $N_\mathcal{F}\geq 2N-1$, the circular correlation effects are completely eliminated.

To demonstrate this, we compare the MACE and the ZAMACE formulations while varying $q$ (the amount of zero padding). For the MACE formulation, we demonstrate two methods. First, we train a MACE template of length $N_{\mathcal{F}}=N+q$. Second, we train the same MACE template, but crop it to length $N$. For the ZAMACE template, we obtain a template of size $N_{\mathcal{F}}=N+q$. However, note that the last $q$ elements of this template are equal to zero because of the ZA constraints. We correlate each template (in the time domain) with the original training signals and compute the unaliased ACE for each of the formulations. We use the term ``unaliased ACE'' to make a distinction from ACE, defined in Eq. \ref{eq:FDMACE-ACE-final}. Unaliased ACE is the average correlation energy of the \textit{linear} correlations of the training samples and the template whereas the ACE term in Eq. \ref{eq:FDMACE-ACE-final} is the average correlation energy of the circular correlations. Note that a lower value of unaliased ACE implies sharper correlation peaks, which results in better pattern localization performance.

\begin{figure}[t]
\subfigure[Unaliased ACE versus amount of zero padding]{\includegraphics[width=0.48\columnwidth]{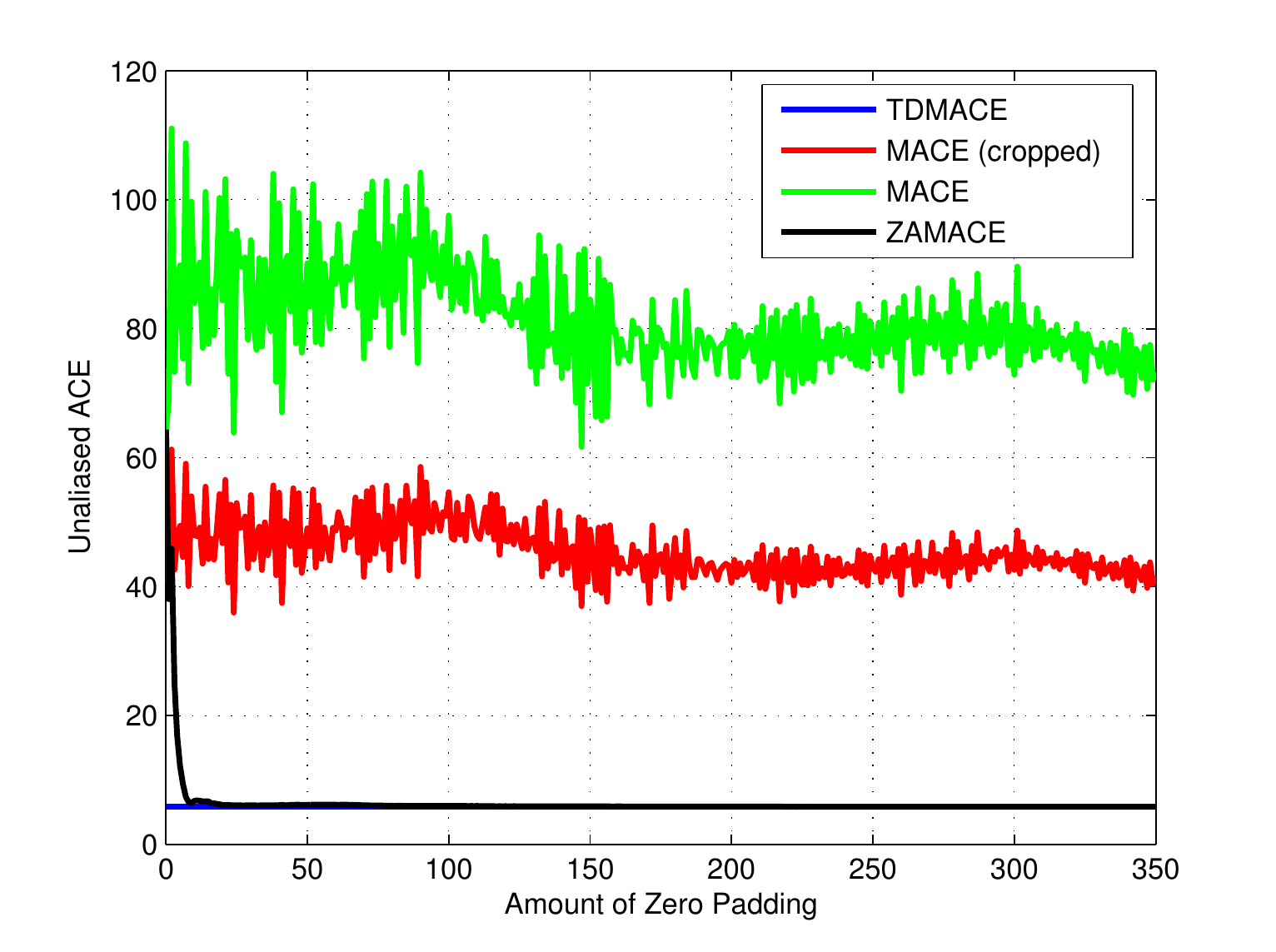}}
\subfigure[MSE to TDMACE]{\includegraphics[width=0.48\columnwidth]{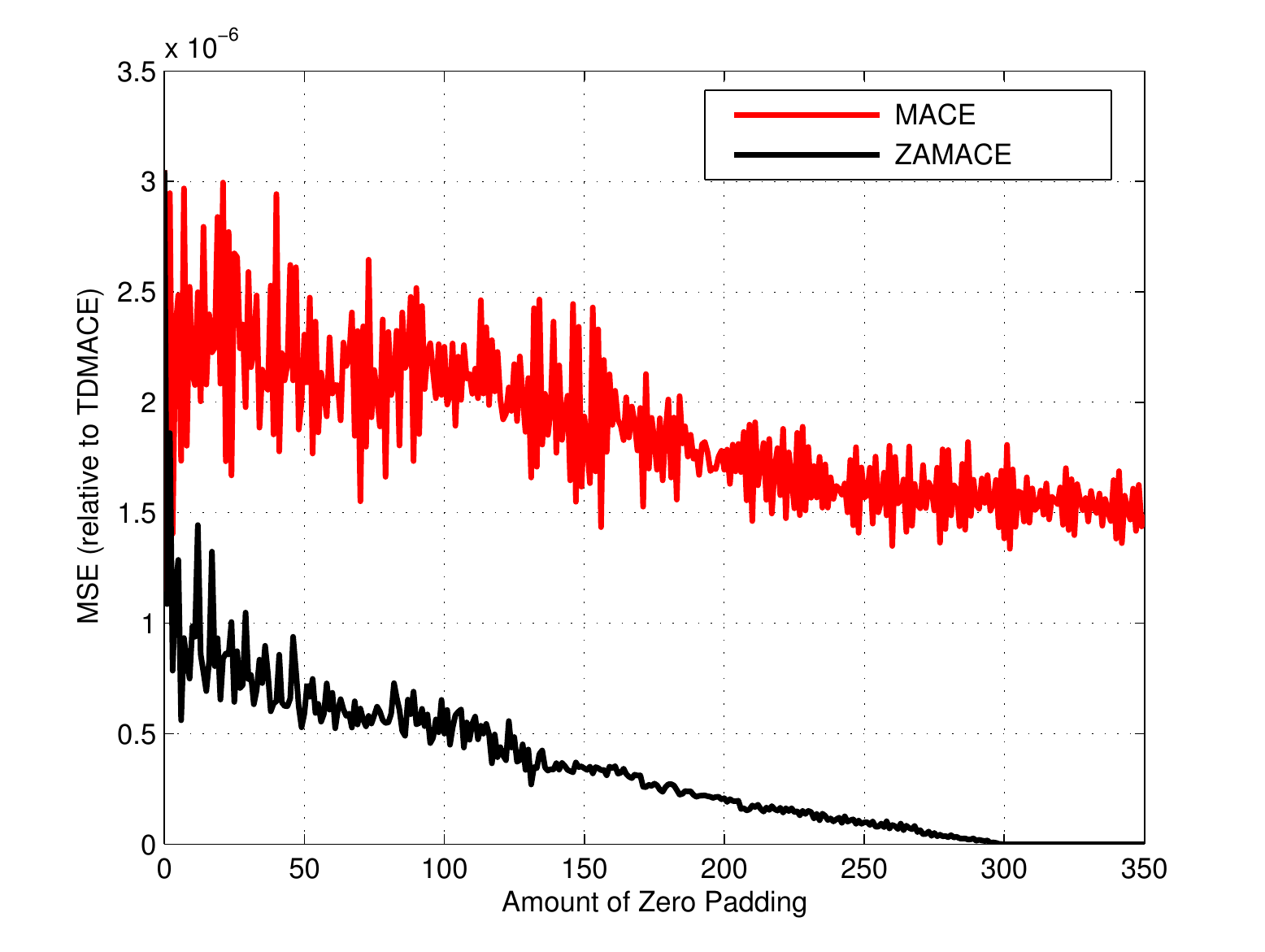}}
\protect\caption{\label{fig:Comparison-of-MACE-and-ZAMACE}Comparison of the conventional MACE, TDMACE and ZAMACE filters.}
\end{figure}

\begin{figure}[!ht]
\begin{center}
\subfigure[Conventional MACE]{\includegraphics[width=0.48\columnwidth]{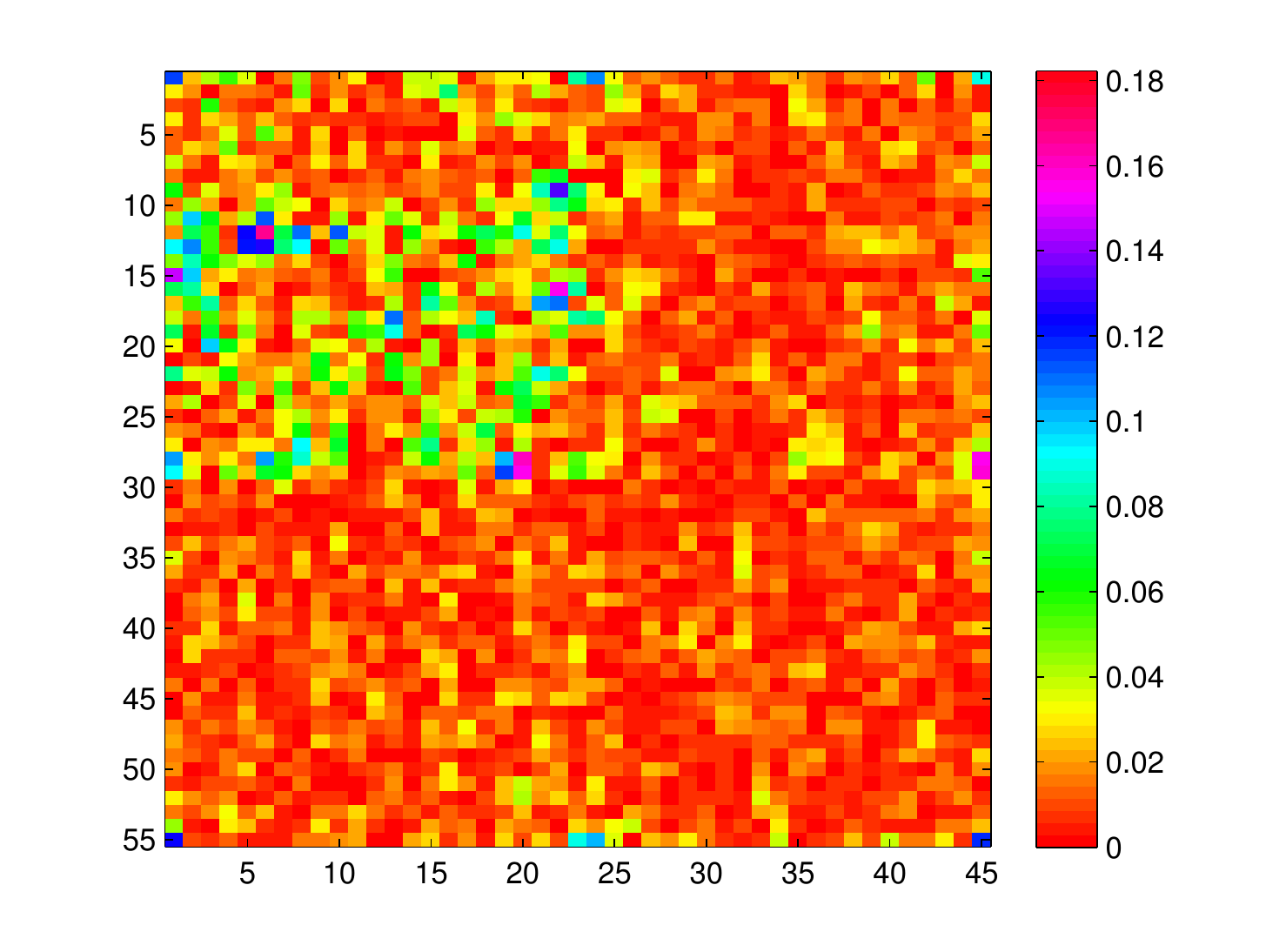}}
\subfigure[ZAMACE]{\includegraphics[width=0.48\columnwidth]{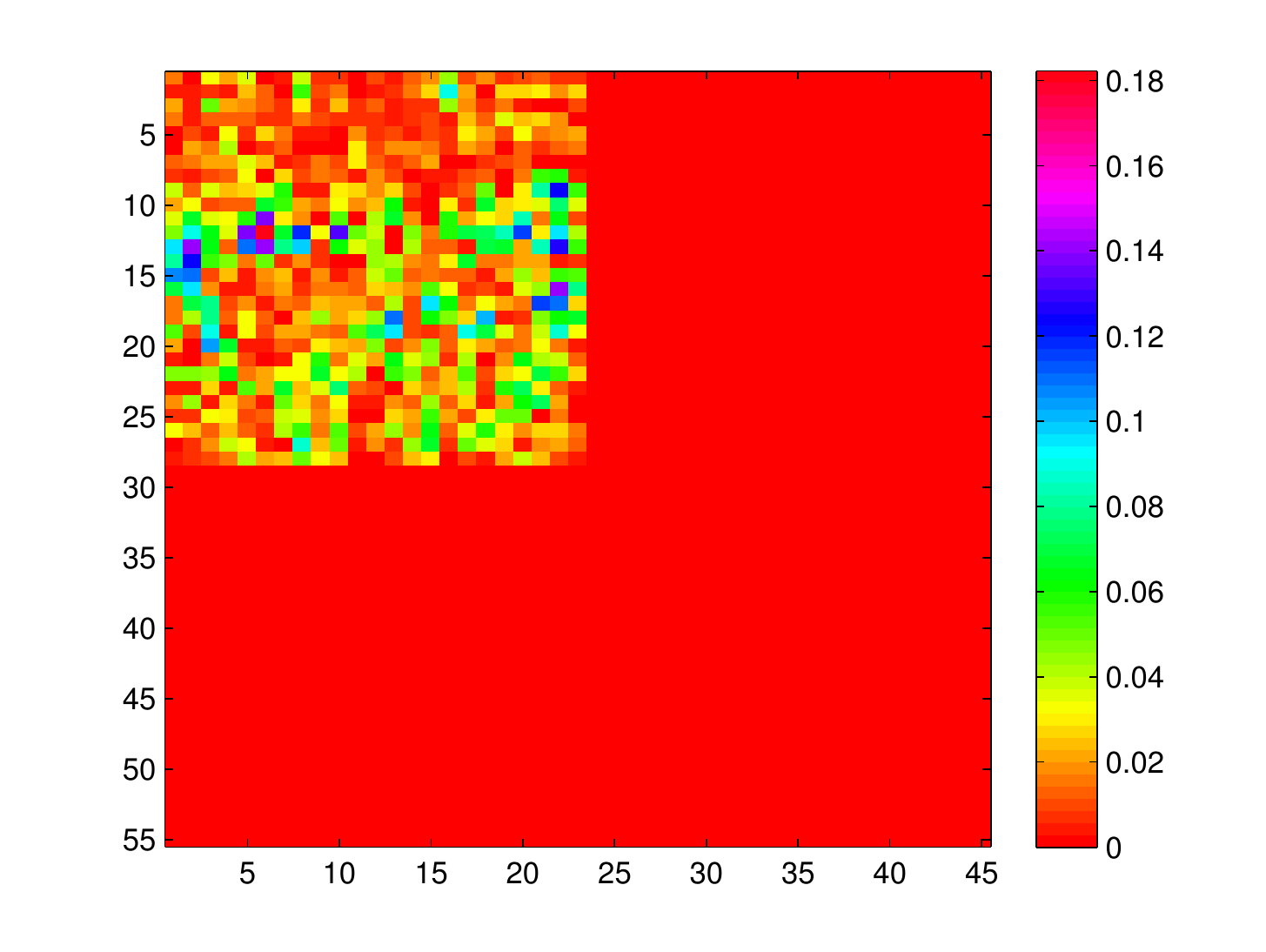}}
\protect\caption{\label{fig:Space-domain-FDMACE-plots}Spatial domain plots of the templates before cropping. Note that the conventional MACE template has large values outside of the ranges of the desired template size, where as the ZAMACE template does not, as the ZA constraints force these values to zero. In these plots, we display the absolute value of the templates.}
\end{center}
\end{figure}

\begin{figure}[!ht]
\begin{center}
\subfigure[Conventional MACE]{\includegraphics[width=0.48\columnwidth]{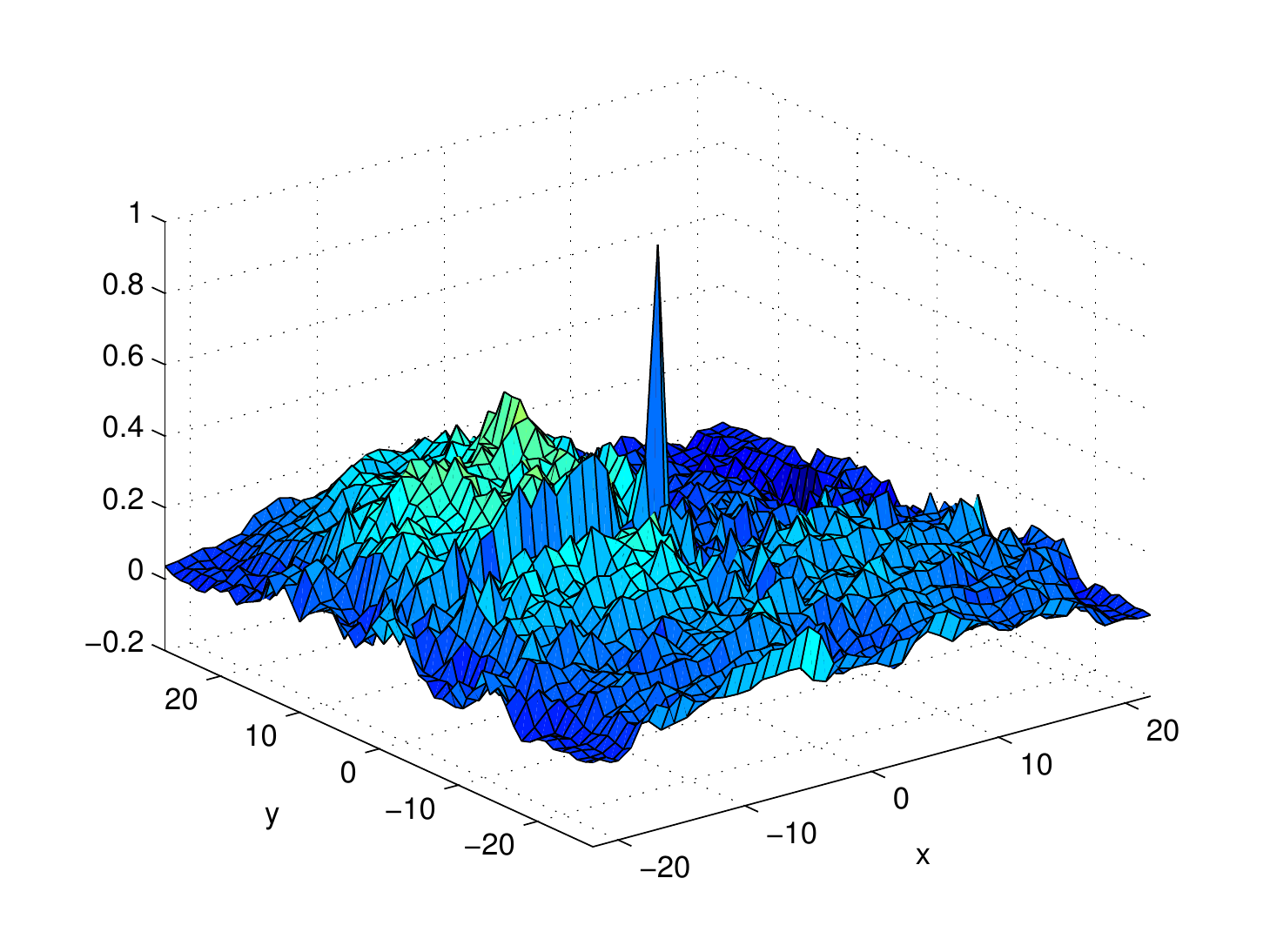}}
\subfigure[ZAMACE]{\includegraphics[width=0.48\columnwidth]{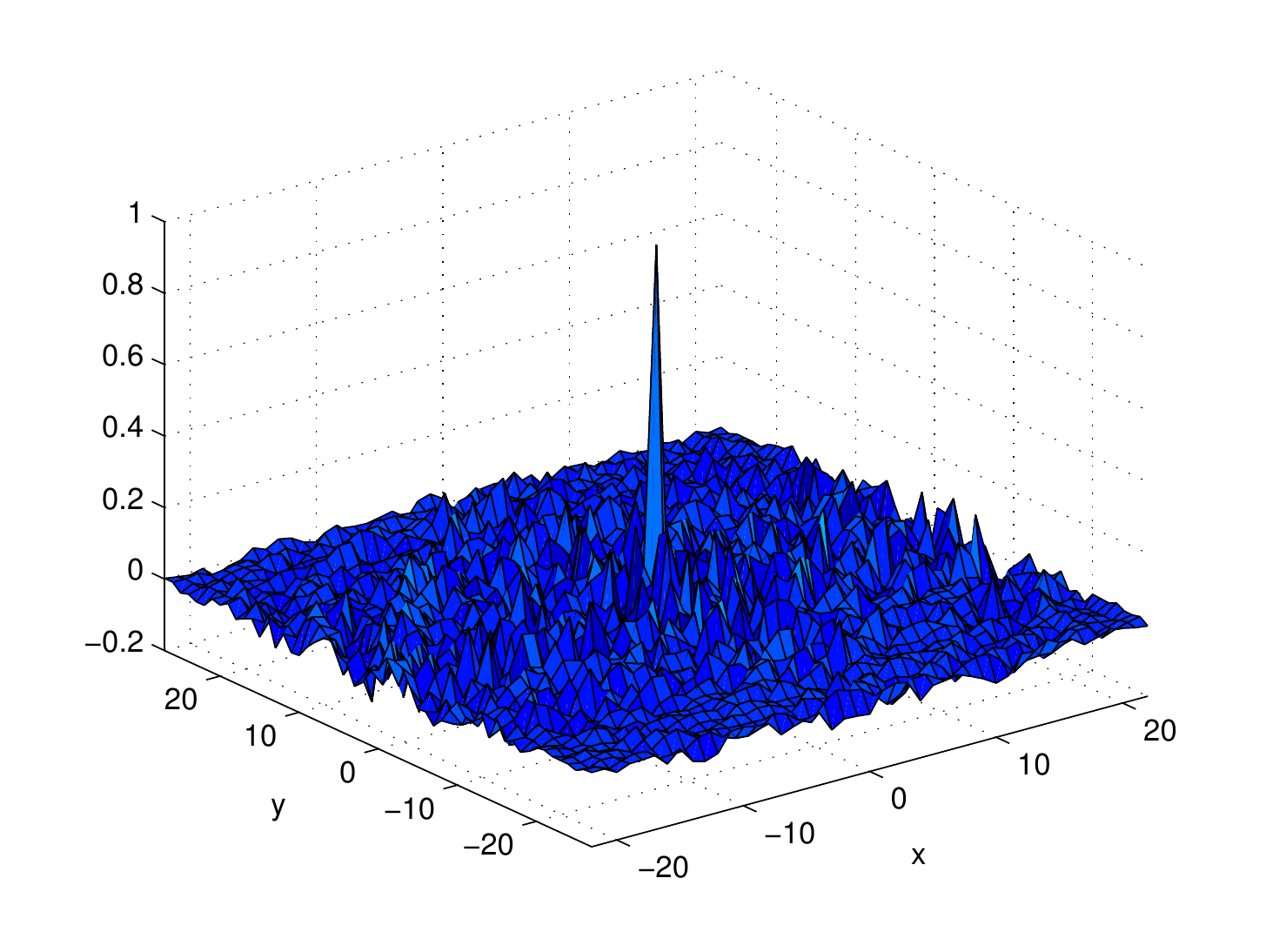}}
\protect\caption{\label{fig:FDMACE2-D_corrfields}Example correlation outputs for one of the training images. Note that the ZAMACE filter yields an output
that is much sharper than the original MACE filter design.}
\end{center}
\end{figure}

The results of this analysis are shown in Fig. \ref{fig:Comparison-of-MACE-and-ZAMACE}a. First, note that the unaliased ACE for the TDMACE is smaller than
both of the conventional frequency-domain MACE formulations for all $q$ values. This indicates that there is a problem with the conventional MACE filter design. Next, note that the unaliased ACE for both of the conventional MACE formulations does not decrease as $q$ increases, which indicates that simply zero padding the training signals does not provide an adequate solution to the aliasing problem. The ZA approach, however, features an unaliased ACE that decreases considerably as $q$ increases. The unaliased ACE of ZAMACE becomes the same as that of the TDMACE when $N_{\mathcal{F}} \geq 2N-1$, or equivalently when  $q\geq300$. We also show in Fig. \ref{fig:Comparison-of-MACE-and-ZAMACE}b the MSE between the (cropped) MACE/ZAMACE templates and the TDMACE template. This cropping is necessary to ensure that the MACE/ZAMACE templates are of the same length as the TDMACE template. Note that the ZAMACE template converges to the TDMACE template as the zero padding approaches $300$, whereas the conventional MACE filter differs significantly from TDMACE
filter even for $q>300$. Therefore, the ZAMACE formulation can be interpreted as a bridge between the original MACE formulation and the TDMACE formulation. With no zero padding, there are no ZA constraints and ZAMACE is the same as MACE; with zero-padding $q\geq N-1$, ZAMACE is the same as TDMACE, and there are no circular correlation effects.
We repeated this experiment in 2-D, training both MACE and ZAMACE filters with three face images of the same person in the AT\&T Database of Faces (formerly the ORL Database of Faces) \cite{ORL-ATT-dataset}. We downsampled the images to size $28\times23$ pixels for computational reasons. We observed the same trend as in the 1-D case--namely, the unaliased ACE decreases as zero padding increases in each dimension. We show the full ($N_{\mathcal{F}}\times M_{\mathcal{F}}=55\times45$) MACE and ZAMACE templates in Fig. \ref{fig:Space-domain-FDMACE-plots}. We show example correlation outputs obtained from these MACE and ZAMACE templates in Fig. \ref{fig:FDMACE2-D_corrfields}. Note that the correlation output has lower correlation energy for the ZA case than it does for the conventional case. The ZAMACE filter yields an output that is more consistent with the original MACE filter design criteria.

\section{\label{sec:Zero-Aliasing-Correlation-Filter-Formulations}Zero-Aliasing Correlation Filters}
In this section, we present the mathematical details of ZACFs and demonstrate how ZA constraints can be incorporated into different CF designs. From an optimization perspective we can categorize most CFs into three groups: equality constrained CFs, unconstrained CFs, and inequality constrained CFs. The MACE filter is an example of an equality constrained CF where the dot products between the training images and the template are constrained to equal some pre-specified values. The MOSSE filter \cite{MOSSE, Boddeti_VCF} is an example of an unconstrained filter that minimizes the \emph{localization loss} between a desired correlation output shape and the actual correlation output. Finally, MMCF \cite{MMCF, boddeti2014maximum} is an inequality constrained filter that leverages ideas from support vector machines (SVMs) to achieve better recognition. Although we discuss only a small subset of CFs, similar derivations can be obtained for other CFs.

\subsection{Zero-Aliasing Constraints} 
In the ZACF formulation, we minimize the \emph{localization loss} while enforcing constraints that force the template's tail to be equal to zero. We refer to these constraints as ZA constraints, because they eliminate the aliasing effects caused by circular correlation. For example, for a $K$-channel template $h(n)$, we can express these constraints as $h^k(n)=0$ for $n\geq N$ for every $k$-th channel. Recall that the time domain template $h^k(n)$ is related to the frequency domain filter $H^k(r)$ through the inverse-DFT as,
\begin{equation}
h^k(n)=\frac{1}{N_{\mathcal{F}}}\sum_{r=0}^{N_{\mathcal{F}}-1}H^k(r)e^{\frac{j2\pi rn}{N_{\mathcal{F}}}}
\end{equation}
To satisfy the ZA constraints, we need to satisfy the following set of linear equations,
\begin{equation}
\label{eq:zero-aliasing-constraints-FD}
h^k(n)=\frac{1}{N_{\mathcal{F}}}\sum_{r=0}^{N_{\mathcal{F}}-1}H^k(r)e^{\frac{j2\pi rn}{N_{\mathcal{F}}}}=0\mbox{  for }N\leq n < N_{\mathcal{F}}
\end{equation}
\noindent This can be written in matrix-vector notation as
\begin{gather}
\label{eq:ZA-constraints1}
\mathbf{\hat{Z}}^{\dagger}\mathbf{\hat{h}}^k=\mathbf{0}
\end{gather}
\noindent where
\begin{equation}
\mathbf{\hat{Z}}^{\dagger}=\left[\begin{array}{cccc}
1 & e^{j2\pi(1)(N)/N_{\mathcal{F}}} & \cdots & e^{j2\pi(N_{\mathcal{F}}-1)(N)/N_{\mathcal{F}}}\\
1 & e^{j2\pi(1)(N+1)/N_{\mathcal{F}}} & \cdots & e^{j2\pi(N_{\mathcal{F}}-1)(N+1)/N_{\mathcal{F}}}\\
\vdots & \vdots & \ddots & \vdots\\
1 & e^{j2\pi(1)(N_{\mathcal{F}}-1)/N_{\mathcal{F}}} & \cdots & e^{j2\pi(N_{\mathcal{F}}-1)(N_{\mathcal{F}}-1)/N_{\mathcal{F}}}
\end{array}\right]
\end{equation}
\noindent Note that the $\mathbf{\hat{Z}}^{\dagger}$ is a matrix of size $(N_{\mathcal{F}}-N)\times N_{\mathcal{F}}$,
and $\mathbf{0}$ is a zero vector of length $N_{\mathcal{F}}-N$. Aggregating the ZA constraints from all the $K$-channels of the signal we have,
\begin{gather}
\label{eq:ZA-constraints}
\mathbf{\hat{A}}^{\dagger}\mathbf{\hat{h}}=\mathbf{0}
\end{gather}
\noindent where $\mathbf{\hat{A}}^{\dagger} = \mathbf{I}_{K} \otimes \mathbf{\hat{Z}}^{\dagger}$, the kronecker product between $\mathbf{\hat{Z}}^{\dagger}$ and an identity matrix $\mathbf{I}_{K}$. These constraints in Eq. \ref{eq:ZA-constraints} are the ZA constraints that characterize ZACFs.

\subsection{ZAMACE\label{sub:ZAMACE}}
In Section \ref{sub:Correlation-Filter-Design}, we discussed the formulation of the MACE filter. In the ZAMACE formulation, we modify the MACE formulation to include the ZA constraints, i.e., we minimize the \emph{localization loss} while enforcing both the peak constraints in Eq. \ref{eq:fdmace_constraints} and ZA constraints in Eq. \ref{eq:ZA-constraints} that force the template's tail to be equal to zero.
\begin{eqnarray}
\min_{\mathbf{\hat{h}}} && \mathbf{\hat{h}^{\dagger}\hat{D}\hat{h}} - 2\mathbf{\hat{h}^{\dagger}\hat{p}} \\ 
s.t. && \mathbf{\hat{X}}^{\dagger}\mathbf{\hat{h}}=\mathbf{u} \nonumber \\
&& \mathbf{\hat{A}}^{\dagger}\mathbf{\hat{h}}=\mathbf{0} \nonumber
\label{eq:ZAMACE} 
\end{eqnarray}
\noindent This optimization problem results in a closed form solution for the new ZAMACE filter given by,
\begin{eqnarray}
\mathbf{\hat{h}}&=&\mathbf{\hat{D}}^{-1}\mathbf{\hat{B}}\left(\mathbf{\hat{B}}^{\dagger}\mathbf{\hat{D}}^{-1}\mathbf{\hat{B}}\right)^{-1}\mathbf{k} + \\
&& \left(\mathbf{I}-\mathbf{\hat{D}}^{-1}\mathbf{\hat{B}}(\mathbf{\hat{B}}^{\dagger}\mathbf{\hat{D}}^{-1}\mathbf{\hat{B}})^{-1}\mathbf{\hat{B}}^{\dagger}\right)\mathbf{\hat{D}}^{-1}\mathbf{\hat{p}} \nonumber
\end{eqnarray}
\noindent where
\begin{equation}
\mathbf{\hat{B}}=\left[\begin{array}{c}
\mathbf{\hat{X}}\\
\mathbf{\hat{A}}\\
\end{array}\right],\:
\mathbf{k}=\left[\begin{array}{c}
\mathbf{u}\\
\mathbf{0}\\
\end{array}\right] \nonumber
\end{equation}
\noindent The ZAMACE expression $\mathbf{\hat{h}}=\mathbf{\hat{D}}^{-1}\mathbf{\hat{B}}\left(\mathbf{\hat{B}}^{\dagger}\mathbf{\hat{D}}^{-1}\mathbf{\hat{B}}\right)^{-1}\mathbf{k}$ in \cite{FernandezISPA13} is obtained by setting the desired correlation outputs $\mathbf{g}_l$ to all-zero plane which leads to $\mathbf{\hat{p}} = \mathbf{0}$. The size of the filter, $\mathbf{\hat{h}}$, in Eq. \ref{eq:ZAMACE} is the same as in Eq. \ref{eq:FDMACE}. This means that the ZAMACE filter requires no additional storage, and for testing it requires no additional computation or memory than the MACE filter.

Note that the OTSDF filter \cite{OTSDF1-Refregier:90} is a CF that is very similar to the MACE filter. In the OTSDF formulation, the matrix $\mathbf{\hat{D}}$ is replaced by $\mathbf{\hat{T}}=\mathbf{\hat{D}}+\delta\mathbf{I}$, where $\mathbf{I}$ is an identity matrix and $\delta>0$. The inclusion of the identity matrix is to improve noise tolerance.

\subsection{ZAMOSSE\label{sub:ZAMOSSE}}
The MOSSE filter \cite{MOSSE, Boddeti_VCF, MCCF} is an unconstrained filter that minimizes the \emph{localization loss}, defined in Eq. \ref{eq:mse_localization_loss}, between the correlation of the CF template with the training signal(s) and the desired correlation output(s). For example, the desired correlation output in the spatial domain could take on a value of one at the location of a positive class signal and zeros elsewhere. Similar filters that specify a desired correlation output are found in \cite{ASEF,MSESDF,GMACE}, and \cite{Bodetti_PAMI_key_binding}. The MOSSE filter design is formulated as the following optimization problem,
\begin{eqnarray}
\min_{\mathbf{\hat{h}}} && \mathbf{\hat{h}^{\dagger}\hat{D}\hat{h}} - 2\mathbf{\hat{h}^{\dagger}\hat{p}}
\label{eq:MOSSE_optim_fxn}
\end{eqnarray}
\noindent which results in the following closed-form expression for the filter,
\begin{eqnarray}
\mathbf{\hat{h}} = \mathbf{\hat{D}}^{-1}\mathbf{\hat{p}}
\label{eq:initial_expression_nonblockform}
\end{eqnarray}
\noindent This expression is written in matrix-vector notation, but is equivalent to the expression in \cite{MOSSE} for a single channel signal. 

Again, this formulation does not account for the aliasing effects due to circular correlation. To remove the effects of circular correlation, we need to instead optimize Eq. \ref{eq:MOSSE_optim_fxn} subject to the ZA constraints given in Eq. \ref{eq:ZA-constraints}. The ZAMOSSE filter design results in the following closed form expression,
\begin{eqnarray}
\label{eq:ZAMOSSE_optim_fxn}
\mathbf{\hat{h}} = \boldsymbol{\Delta}_{\mathbf{\hat{D}}}\mathbf{\hat{D}}^{-1}\mathbf{\hat{p}}
\end{eqnarray}
\noindent where $\boldsymbol{\Delta}_{\mathbf{\hat{D}}}=\mathbf{I}-\mathbf{\hat{D}}^{-1}\mathbf{\hat{A}}(\mathbf{\hat{A}}^{\dagger}\mathbf{\hat{D}}^{-1}\mathbf{\hat{A}})^{-1}\mathbf{\hat{A}}^{\dagger}$. In our MOSSE and ZAMOSSE filter implementation, we replace $\mathbf{\hat{D}}$ with $\mathbf{\hat{T}}=\mathbf{\hat{D}}+\delta\mathbf{I}$ as described earlier.

\subsection{ZAMMCF\label{sub:ZAMMCF}}
The Maximum-Margin Correlation Filter was recently introduced in \cite{MMCF,boddeti2014maximum}. This filter combines the localization ability of the CFs and the generalization capability of large-margin based classifiers like Support Vector Machines (SVMs). Traditionally, constrained CF designs (e.g., MACE) are constrained such that the dot product of a training image and the CF template is set to a specific value. In the MMCF formulation, however, this hard equality constraint is relaxed and replaced with inequality constraints for maximizing the margin of separation between the positive and negative class training samples.
\begin{eqnarray}
\label{eq:mmcf}
\min_{\mathbf{\hat{h}}} && \mathbf{\hat{h}}^{\dagger}\mathbf{\hat{T}}\mathbf{\hat{h}}-2\hat{\mathbf{h}}^{\dagger}\mathbf{\hat{p}}+2C\mathbf{1}^{T}\boldsymbol{\xi} \\
s.t. && y_l\left(\mathbf{\hat{x}}^{\dagger}_l\mathbf{\hat{h}}+b\right)\geq y_lu_l-\xi_l \nonumber \\
&& \xi_l\geq0 \nonumber
\end{eqnarray}
where $\mathbf{\hat{T}}=\mathbf{\hat{D}}+\delta\mathbf{I}$ as described before, the vector $\boldsymbol{\xi}=[\xi_{1},\dots,\xi_{L}]^{T}$ is a real vector of slack variables that penalize training images that are on the wrong side of the margin (as in the SVM formulation), $C>0$ is a trade-off parameter, $y_l$ is the class label ($1$ for positive class and $-1$ for negative class), and $u_l$ is the minimum peak magnitude which is typically set to 1. As shown in \cite{AndresThesis}, the filter may be expressed as
\begin{equation}
\mathbf{\hat{h}}=\mathbf{\hat{T}}^{-1}\mathbf{\hat{p}}+\mathbf{\hat{T}}^{-1}\mathbf{\hat{X}}\mathbf{Y}\mathbf{a}
\end{equation}
\noindent where $\mathbf{Y}$ is a diagonal matrix with the class labels $y_l$ along the diagonal, and $\mathbf{a}$ is the solution of the dual problem,
\begin{equation}
\max_{\mathbf{0}\leq\mathbf{a}\leq C\mathbf{1}}\mathbf{a}^{T}\mathbf{M}\mathbf{a}+\mathbf{a}^{T}\mathbf{d}\label{eq:quadprog_MMCF}
\end{equation}
\noindent where
\begin{eqnarray}
\mathbf{M}&=&-\mathbf{Y}\mathbf{\hat{X}}^{\dagger}\mathbf{\hat{T}}^{-1}\mathbf{\hat{X}}\mathbf{Y} \nonumber \\
\mathbf{d}&=&2\mathbf{Y}(\mathbf{u}-\mathbf{\hat{X}}^{\dagger}\mathbf{\hat{T}}^{-1}\mathbf{\hat{p}}) \nonumber
\end{eqnarray}
To obtain the ZA formulation of MMCF (ZAMMCF), we add the ZA constraints, Eq. \ref{eq:ZA-constraints}, to the MMCF formulation, Eq. \ref{eq:mmcf}. The solution to this problem can be expressed as, 
\begin{eqnarray}
\mathbf{\hat{h}} & = & \mathbf{\hat{T}}^{-1}(\hat{\mathbf{p}}+\mathbf{\hat{X}}\mathbf{Y}\mathbf{a}+\mathbf{\hat{A}}\boldsymbol{\omega})\label{eq:h_ZAMMCF}
\end{eqnarray}
\noindent where $\boldsymbol{\omega}=-(\mathbf{\hat{A}}^{\dagger}\mathbf{\hat{T}}^{-1}\mathbf{\hat{A}})^{-1}\mathbf{\hat{A}}^{\dagger}(\mathbf{\hat{T}}^{-1}\hat{\mathbf{p}}+\mathbf{\hat{T}}^{-1}\mathbf{\hat{X}}\mathbf{Y}\mathbf{a})$ and $\mathbf{a}$ is the solution of the dual problem,
\begin{equation}
\max_{\mathbf{0}\leq\mathbf{a}\leq C\mathbf{1}}\mathbf{a}^{T}\mathbf{M}\mathbf{a}+\mathbf{a}^{T}\mathbf{d}\label{eq:quadprog_ZAMMCF}
\end{equation}
\noindent where
\begin{eqnarray}
\mathbf{M}&=&-\mathbf{Y}\mathbf{\hat{X}}^{\dagger}\boldsymbol{\Delta}_{\mathbf{\hat{T}}}\mathbf{\hat{T}}^{-1}\hat{\mathbf{X}}\mathbf{Y}\nonumber\\
\mathbf{d}&=&2\mathbf{Y}\left(\mathbf{u}-\mathbf{\hat{X}}^{\dagger}\boldsymbol{\Delta}_{\mathbf{\hat{T}}}\mathbf{\hat{T}}^{-1}\hat{\mathbf{p}}\right)\nonumber\\
\boldsymbol{\Delta}_{\mathbf{\hat{T}}}&=&\mathbf{I}-\mathbf{\hat{T}}^{-1}\mathbf{\hat{A}}(\mathbf{\hat{A}}^{\dagger}\mathbf{\hat{T}}^{-1}\mathbf{\hat{A}})^{-1}\mathbf{\hat{A}}^{\dagger}\nonumber
\end{eqnarray}

\subsection{Extension to Multi-Dimensional Signals}
\begin{figure}
\centering
\begin{tikzpicture}[outer sep=0.05cm,node distance=0.8cm,]
\tikzstyle{bigbox} = [minimum width=5cm, minimum height=3cm, fill=blue!50, rectangle, thick, drop shadow, draw]
\tikzstyle{box} = [minimum width=2.5cm, minimum height=1.5cm, rectangle, thick, fill=blue!10, draw]
\node[bigbox] (12) {};
\node[box] at (-1.25,0.75) (11) {};
\draw[thick,<->] (2,-1.5) -- (2,1.5);
\draw[thick,<->] (-2.5,-1) -- (2.5,-1);
\draw[thick,<->] (-0.3,0) -- (-0.3,1.5);
\draw[thick,<->] (-2.5,0.3) -- (0,0.3);
\node (1) at (-0.5,1) {$N$};
\node (2) at (-2.0,0.5) {$M$};
\node (3) at (1.7,1) {$N_{\mathcal{F}}$};
\node (4) at (-2.0,-0.7) {$M_{\mathcal{F}}$};
\end{tikzpicture}
\caption{\label{fig:Illustration-of-2-D-FDMACE}Illustration of the ZA constraints required for the 2-D formulation of the ZAMACE template. }
\end{figure}
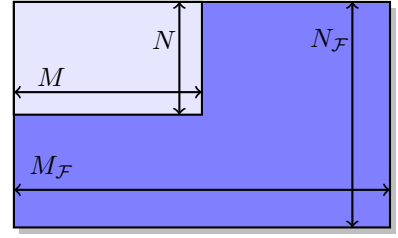

Extending the ZA formulation to multi-dimensional signals is a matter of determining the matrix $\mathbf{\hat{A}}$ such that $\mathbf{\hat{A}}^{\dagger}\mathbf{\hat{h}}=\mathbf{0}$ when $\mathbf{\hat{h}}$ is a vectorized $M$-D DFT. We illustrate the 2-D ZA constraints in Fig. \ref{fig:Illustration-of-2-D-FDMACE}. Here, the shaded region of the template must be set to zero via ZA constraints. We can express $\mathbf{h}$ (the vectorized template) as 
\begin{equation}
\mathbf{h}=\mathbf{W}\mathbf{\hat{h}}\label{eq:vectorized-Inverse-DFT}
\end{equation}
\noindent where the matrix $\mathbf{W}$ accomplishes a 2-D inverse-DFT. Note
that we only wish to constrain some of the entries of vector $\mathbf{h}$
to zero. The rows of matrix $\mathbf{\hat{A}}^{\dagger}$ are taken from the rows
of matrix $\mathbf{W}$ corresponding to the elements of $\mathbf{h}$
that we wish to constrain. Details may be found in \cite{FernandezISPA13}.
Note that matrix $\mathbf{\hat{A}}$ is considerably larger in 2-D thereby posing a computational challenge for numerically computing ZACFs. 

\section{\label{sec:Computational-Considerations}Computational Considerations}
In this section, we discuss the computational challenges that arise in the design of ZACFs. Although we have analytical solutions for ZACFs, designing the filter can become computationally intensive (memory wise), especially for large training images. \textit{Note that applying the resulting ZACF to test data requires no additional computation or memory compared to traditional CFs}. \textit{The computational considerations presented in this section are only for training the CF}. In this section, we present two different methods we have developed to overcome the computational challenge
and efficiently design ZACFs.

\subsection{Reduced-Aliasing Correlation Filters}
Analytical solutions for ZACFs typically result in greater computational complexity than their conventional counterparts. To illustrate this, we use the MACE filter as an example. In the original MACE formulation, the matrix $\mathbf{\hat{X}}^{\dagger}\mathbf{\hat{D}}^{-1}\hat{\mathbf{X}}$ must be inverted. This is usually of low computational complexity, as the size of this matrix ($L\times L$) is determined only by the number of training images, $L$, which is usually much smaller than the number of pixels in the training images. For the ZAMACE case, however, we must invert the matrix $\mathbf{\hat{B}}^{\dagger}\mathbf{\hat{D}}^{-1}\mathbf{\hat{B}}$, which is of size $(L+KN_{\mathcal{F}}-KN)\times(L+KN_{\mathcal{F}}-KN)$, which is usually much larger than the MACE case. Note that if we reduce the number of ZA constraints ($N_{\mathcal{F}}-N$), we can reduce the size of this matrix. We have explored several ways to do this. First, we explored constraining only a portion of the template tail based on the observation that more energy is usually contained in the main portion of the template. However, this technique is not nearly as successful as simply reducing the DFT size ($N<N_{\mathcal{F}}<2N-1$) and constraining the entire tail. Doing so means that there is still aliasing due to circular correlation; however, we have noticed that this aliasing is significantly reduced, leading to very good results while improving the computational requirements. We refer to this method as the reduced-aliasing correlation filter (RACF). RACFs were inspired by our earlier experiments (see Fig. \ref{fig:Comparison-of-MACE-and-ZAMACE}a), in which we noticed that the unaliased ACE actually decreases rapidly as zero-padding increases. We evaluate the RACF method in Section
\ref{sub:Computational-Experiments}.

\subsection{Proximal Gradient Descent Method}
Although we have derived closed-form solutions for some ZACFs, sometimes implementing these expressions may be impractical from a computational
and memory perspective. An alternative solution is to use an iterative algorithm to solve for the ZACF. One iterative method to do this is using the proximal gradient descent method \cite{prox_gradient}. We have developed this method for unconstrained filters (ZAMOSSE), equality constrained filters (ZAMACE/ZAOTSDF) and inequality constrained filters (ZAMMCF). The key idea to efficiently optimize the ZACF formulations is to impose the ZA constraints directly in the spatial domain instead of forming the matrix $\mathbf{\hat{A}}$ as in the closed form solution. Imposing the constraints in the spatial domain can be done by a simple projection of the template onto the space $\Omega$ whose tail is set to zero. We define $P_{\Omega}(\mathbf{x})$ as the operator which projects $\mathbf{x}$ onto the set $\Omega$ i.e., $P_{\Omega}(\mathbf{x})$ is the operator that sets the tail of $\mathbf{x}$ to zero. Further, $P_{Z}(\mathbf{x})$ and $P_{C}(\mathbf{x})$ are defined as complementary operators which zero-pad and crop the signal by an appropriate amount respectively.

As described earlier, most CF designs optimize the \emph{localization loss}, $f(\mathbf{\hat{h}})=\hat{\mathbf{h}}^{\dagger}\mathbf{\hat{T}}\mathbf{\hat{h}}-2\mathbf{\hat{h}}^{\dagger}\mathbf{\hat{p}}$, whose gradient is $\nabla f(\hat{\mathbf{h}})=2\mathbf{\hat{T}}\mathbf{\hat{h}}-2\mathbf{\hat{p}}$. Standard gradient descent \cite{Boyd_optim} finds an optimal solution $\mathbf{\hat{h}}_{opt}$ by choosing an initial solution, $\hat{\mathbf{h}}_{0}$ and iteratively reducing the cost function $f(\mathbf{\hat{h}})$. Each iteration is computed as a function of the previous solution,
\begin{equation}
\mathbf{\hat{h}}_{t+1}=\mathbf{\hat{h}}_{t}-\eta_{t}\nabla f\left(\hat{\mathbf{h}}_{t}\right)
\end{equation}
\noindent where $\eta_{t}$ is the step size at iteration $t$. However, standard gradient descent does not allow for the ZA constraints to be imposed on the filter. Therefore, we use proximal gradient descent \cite{prox_gradient} to find the optimal solution while satisfying the ZA constraints,
\begin{equation}
\mathbf{\hat{h}}_{t+1}=prox\left(\mathbf{\hat{h}}_{t}-\eta_{t}\nabla f\left(\hat{\mathbf{h}}_{t}\right)\right)
\end{equation}
\noindent Here, the $prox()$ operator imposes constraints on the filter. 

\begin{figure}[t]
\centering
\subfigure[Unconstrained and Inequality Constrained Case]{
\begin{tikzpicture}[outer sep=0.05cm,node distance=0.8cm,]
\tikzstyle{bigbox} = [minimum width=2cm, minimum height=1.5cm, draw, thick, fill=blue!50, rectangle, drop shadow]
\tikzstyle{box} = [minimum width=1cm, minimum height=0.75cm, rectangle, thick, fill=yellow!50, draw]
\node[bigbox] (1) {};
\node[bigbox, right of=1, node distance=3.2cm, fill=green!50] (2) {};
\node[bigbox, right of=2, node distance=3.2cm, fill=red!70] (3) {};
\node[box, right of=1, node distance=3.2cm] at (-0.5,0.375) (4) {};
\node[right of=1, node distance=3.2cm] at (0.2,-0.2) (5) {{\footnotesize 0}};
\node[right of=1, node distance=1cm] (11) {};
\node[left of=2, node distance=1.2cm] (12) {};
\node[right of=2, node distance=1.1cm] (21) {};
\node[left of=3, node distance=1.1cm] (22) {};
\draw[very thick, ->] (11)--(12) node[above,midway] {{\tiny $\mathcal{F}^{-1}$}};
\draw[very thick, ->] (21)--(22) node[above,midway] {{\tiny $\mathcal{F}$}};
\draw[very thick, ->] (5) -- (4.2,-0.2);
\draw[very thick, <-] (2.2,-0.2) -- (5);
\draw[very thick, ->] (5) -- (3.4,0.75);
\draw[very thick, <-] (3.4,-0.75) -- (5);
\draw [decorate,decoration={brace,amplitude=2pt,mirror},xshift=0pt,yshift=0pt] (2.2,-0.8)--(4.2,-0.8) node [black,midway,xshift=0pt,yshift=-10pt] {\tiny $M_{\mathcal{F}}$};
\draw [decorate,decoration={brace,amplitude=2pt,mirror},xshift=0pt,yshift=0pt] (4.3,-0.75)--(4.3,0.75) node [black,midway,xshift=10pt,yshift=-10pt] {\tiny $N_{\mathcal{F}}$};
\draw [decorate,decoration={brace,amplitude=2pt},xshift=0pt,yshift=0pt] (2.2,0.8)--(3.2,0.8) node [black,midway,xshift=0pt,yshift=10pt] {\tiny $M$};
\draw [decorate,decoration={brace,amplitude=2pt},xshift=0pt,yshift=0pt] (2.15,0)--(2.15,0.75) node [black,midway,xshift=-7pt,yshift=3pt] {\tiny $N$};
\end{tikzpicture}}\\
\subfigure[Equality Constrained Case]{
	\begin{tikzpicture}[outer sep=0.05cm,node distance=0.8cm]
\tikzstyle{bigbox} = [minimum width=2cm, minimum height=1.5cm, draw, thick, fill=blue!50, rectangle, drop shadow]
\tikzstyle{box} = [minimum width=1cm, minimum height=0.75cm, rectangle, thick, fill=yellow!50, draw]
\tikzstyle{mybar} = [minimum width=0.02cm, minimum height=0.2cm, rectangle, draw]

\node[bigbox] (a1) {};
\node[bigbox, right of=a1, node distance=3.2cm, fill=green!50] (a2) {};
\node[box, right of=a1, node distance=3.2cm] at (-0.5,0.375) (a4) {};
\node[bigbox, below of=a2, node distance=4.2cm, fill=green!50] (a5) {};
\node[bigbox, right of=a5, node distance=3.2cm, fill=violet!70] (a6) {};
\node[box, below of=a2, node distance=4.2cm, fill=red!70] at (2.7,0.375) (a7) {};
\node[box, below left of=a2, node distance=2.0cm, fill=yellow!50, drop shadow] at (3.0,-0.7) (a8) {};
\node[box, right of=a8, node distance=4.4cm, fill=red!70, drop shadow] (a9) {};

\node[right of=1, node distance=3.2cm] at (0.2,-0.2) (a5) {{\footnotesize 0}};
\node[right of=1, node distance=1cm] (a11) {};
\node[left of=2, node distance=1.2cm] (a12) {};
\node[left of=a6, node distance=1.1cm] (a62) {};
\node[left of=a6, node distance=2.2cm] (a61) {};
\node[below right of=a2] at (3.6,-2.9) (a55) {};
\node[below right of=a9] at (5.0,-1.9) (a99) {};
\node[left of=a9, node distance=1.45cm] (a91) {};
\node[left of=a9, node distance=0.45cm] (a92) {};
\node[right of=a8, node distance=0.4cm] (a81) {};
\node[right of=a8, node distance=1.3cm] (a82) {};
\node[right of=a8, node distance=1.8cm] (a71) {+};
\node[right of=a8, node distance=2.5cm] (a72) {=};
\node[mybar, right of=a8, node distance=1.5cm, fill=yellow!50] (a31) {};
\node[mybar, right of=a8, node distance=2.1cm, fill=brown!50] (a32) {};
\node[mybar, right of=a72, node distance=0.4cm, fill=red!70] (a33) {};

\node[above right of=a8] at (1.0,-2.3) (a142) {};
\node[above right of=a8] at (1.6,-1.3) (a141) {};

\node[mybar, below of=a31, node distance=0.3cm, fill=yellow!50] (b1) {};
\node[mybar, below of=a31, node distance=0.1cm, fill=yellow!50] (b2) {};
\node[mybar, above of=a31, node distance=0.1cm, fill=yellow!50] (b3) {};
\node[mybar, above of=a31, node distance=0.3cm, fill=yellow!50] (b4) {};

\node[mybar, below of=a32, node distance=0.3cm, fill=brown!50] (c1) {};
\node[mybar, below of=a32, node distance=0.1cm, fill=brown!50] (c2) {};
\node[mybar, above of=a32, node distance=0.1cm, fill=brown!50] (c3) {};
\node[mybar, above of=a32, node distance=0.3cm, fill=brown!50] (c4) {};

\node[mybar, below of=a33, node distance=0.3cm, fill=red!70] (d1) {};
\node[mybar, below of=a33, node distance=0.1cm, fill=red!70] (d2) {};
\node[mybar, above of=a33, node distance=0.1cm, fill=red!70] (d3) {};
\node[mybar, above of=a33, node distance=0.3cm, fill=red!70] (d4) {};

\node[below of=b1, node distance=0.3cm] (a41) {{\tiny $\mathbf{h}^{\#}$}};
\node[below of=c1, node distance=0.34cm] (a42) {{\tiny $\mathbf{h}_{\Delta}$}};

\draw[very thick, ->] (a141)--(a142);
\draw[very thick, ->] (a81)--(a82);
\draw[very thick, ->] (a91)--(a92);
\draw[very thick, ->] (a99)--(a55);
\draw[very thick, ->] (a11)--(a12) node[above,midway] {{\tiny $\mathcal{F}^{-1}$}};
\draw[very thick, ->] (a61)--(a62) node[above,midway] {{\tiny $\mathcal{F}$}};
\draw[very thick, ->] (a5) -- (4.2,-0.2);
\draw[very thick, <-] (2.2,-0.2) -- (a5);
\draw[very thick, ->] (a5) -- (3.4,0.75);
\draw[very thick, <-] (3.4,-0.75) -- (a5);
\draw [decorate,decoration={brace,amplitude=2pt},xshift=1pt,yshift=0pt] (1.0,-2.5)--(1.0,-1.7) node [black,midway,xshift=-8pt,yshift=0pt] {{\tiny $N$}};
\draw [decorate,decoration={brace,amplitude=2pt},xshift=2pt,yshift=0pt] (1.0,-1.7)--(2.0,-1.7) node [black,midway,xshift=-4pt,yshift=+7pt] {{\tiny $M$}};
\draw [decorate,decoration={brace,amplitude=2pt},xshift=-2pt,yshift=0pt] (3.0,-2.5)--(3.0,-1.7) node [black,midway,xshift=-8pt,yshift=-7pt] {{\tiny $NM$}};
\draw [decorate,decoration={brace,amplitude=2pt,mirror},xshift=0pt,yshift=0pt] (2.2,-0.8)--(4.2,-0.8) node [black,midway,xshift=0pt,yshift=-10pt] {\tiny $M_{\mathcal{F}}$};
\draw [decorate,decoration={brace,amplitude=2pt,mirror},xshift=0pt,yshift=0pt] (4.3,-0.75)--(4.3,0.75) node [black,midway,xshift=10pt,yshift=-10pt] {\tiny $N_{\mathcal{F}}$};
\draw [decorate,decoration={brace,amplitude=2pt},xshift=0pt,yshift=0pt] (2.2,0.8)--(3.2,0.8) node [black,midway,xshift=0pt,yshift=10pt] {\tiny $M$};
\draw [decorate,decoration={brace,amplitude=2pt},xshift=0pt,yshift=0pt] (2.15,0)--(2.15,0.75) node [black,midway,xshift=-7pt,yshift=3pt] {\tiny $N$};
\end{tikzpicture}}
\caption{\label{fig:Illustration-of-prox}Illustration of the proximal step, which is performed in the spatial domain. All DFTs and Inverse-DFTs are in 2-D.}
\end{figure}
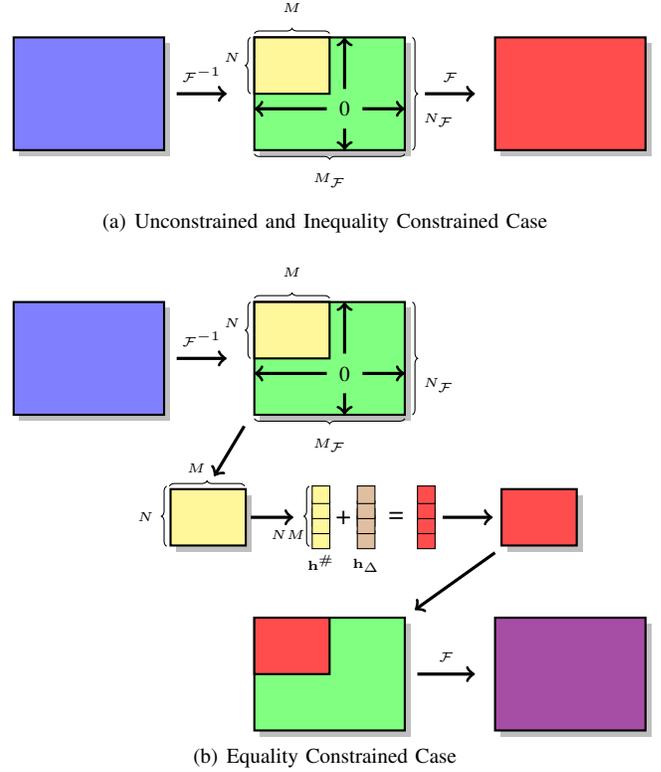

For unconstrained filters (e.g., ZAMOSSE), the $prox()$ operator (see Fig. \ref{fig:Illustration-of-prox}a and Algorithm \ref{alg:Prox_unconstrained}) transforms the filter update $\mathcal{F}^{-1}\left(\mathbf{\hat{h}}_{t}-\eta_{t}\nabla f\left(\hat{\mathbf{h}}_{t}\right)\right) \rightarrow \mathbf{h}_{t+\frac{1}{2}}$ into the spatial domain and sets the template's tail to zero i.e., $\mathbf{h}_{t+\frac{1}{2}} \leftarrow P_{\Omega}\left(\mathbf{h}_{t+\frac{1}{2}}\right)$. It then transforms the resulting template back into the frequency domain to obtain $\mathbf{\hat{h}}_{t+1} \leftarrow \mathcal{F}\left(\mathbf{h}_{t+\frac{1}{2}}\right)$. The full algorithm is described in Algorithm \ref{alg:Prox_unconstrained}.

For the equality constrained case (e.g. ZAMACE/ZAOTSDF), the $prox()$ operator contains two steps (see Fig. \ref{fig:Illustration-of-prox}b and Algorithm \ref{alg:Prox_constrained}). First, the filter update is transformed into the spatial domain $\mathcal{F}^{-1}\left(\mathbf{\hat{h}}_{t}-\eta_{t}\nabla f\left(\hat{\mathbf{h}}_{t}\right)\right) \rightarrow \mathbf{h}_{t+\frac{1}{2}}$ and the template's tail is set to zero i.e., $\mathbf{h}_{t+\frac{1}{2}} \leftarrow P_{\Omega}\left(\mathbf{h}_{t+\frac{1}{2}}\right)$. For the second step, we extract the main portion of the template and vectorize it i.e., $\mathbf{h}^{\#}_{t+\frac{1}{2}} \leftarrow P_C\left(\mathbf{h}_{t+\frac{1}{2}}\right)$. We now seek to update $\mathbf{h}^{\#}_{t+\frac{1}{2}}$ by $\mathbf{h}_{\Delta}$ such that the desired peak constraints are satisfied and that is closest to the current template $\mathbf{h}_{t}^{\#}$,
\begin{equation}
\mathbf{X}^{T}\left(\mathbf{h}_{t+\frac{1}{2}}^{\#}+\mathbf{h}_{\Delta}\right)=\mathbf{u}
\end{equation}
\noindent where $\mathbf{u}$ is a vector containing the desired peak constraints and $\mathbf{X}$ contains the training signals in the spatial domain along each column. We solve for $\mathbf{h}_{\Delta}$ by solving the following least squares problem,
\begin{eqnarray}
\min_{\mathbf{h}_{\Delta}} && \mathbf{h}_{\Delta}^{T}\mathbf{h}_{\Delta} \\
s.t. && \mathbf{X}^{T}\mathbf{h}_{\Delta}=\mathbf{u}-\mathbf{X}^{T}\mathbf{h}_{t+\frac{1}{2}}^{\#} \nonumber
\end{eqnarray}
\noindent which results in $\mathbf{h}_{\Delta}=\mathbf{X}(\mathbf{X}^{T}\mathbf{X})^{-1}\left(\mathbf{u}-\mathbf{X}^{T}\mathbf{h}_{t+\frac{1}{2}}^{\#}\right)$. Next, we form the vector $\mathbf{h}_{t+\frac{1}{2}}^{\#}+\mathbf{h}_{\Delta}$, zero-pad it by the appropriate amount to get $\mathbf{h}_{t+1} \leftarrow P_Z\left(\mathbf{h}_{t+\frac{1}{2}}^{\#}+\mathbf{h}_{\Delta}\right)$, and map it back to the Fourier domain $\mathbf{\hat{h}}_{t+1} \leftarrow \mathcal{F}\left(\mathbf{h}_{t+1}\right)$. The full algorithm is described in Algorithm \ref{alg:Prox_constrained}.

For the inequality constrained case (e.g., ZAMMCF), since the hinge loss is non-differentiable, gradient descent cannot be directly used. While many sub-gradient descent methods have been developed for the hinge loss, we empirically observed that a simple sub-gradient descent with the proximal operator had poor convergence properties. Therefore, we instead use the differentiable \emph{squared hinge loss} and solve the following optimization problem that is closely related to the formulation in Eq. \ref{eq:mmcf},
\begin{eqnarray}
\label{eq:zammcf}
\min_{\mathbf{\hat{h}}} && \frac{\lambda}{2}\mathbf{\hat{h}}^{\dagger}\mathbf{\hat{T}}\mathbf{\hat{h}} + \frac{1}{2L}\sum_{l=1}^L\left[1-y_l\left(\mathbf{\hat{x}}^{\dagger}_{l}\mathbf{\hat{h}}+b\right)\right]^2_{+} \\
s.t. && \mathbf{\hat{A}}^{\dagger}\mathbf{\hat{h}} = \mathbf{0} \nonumber 
\end{eqnarray}
\noindent where $\lambda=\frac{1}{C}$. Again we adopt accelerated gradient descent along with the proximal step after each iteration to solve for the filter. The gradient of this objective function is,

\begin{eqnarray}
\nabla f(\mathbf{\hat{h}}) = \lambda\mathbf{\hat{T}\hat{h}} - \frac{1}{L}\sum_{i \in \Omega_{sv}}y_i\mathbf{\hat{x}}_i\left(1-y_i\left(\mathbf{\hat{h}}^{\dagger}\mathbf{\hat{x}}_i+b\right)\right) 
\end{eqnarray}
\noindent where the support vector set $\Omega_{sv}$ is defined as,
\begin{eqnarray}
\Omega_{sv} = \left\{i \in (1,\dots,n) \mathrel{}\middle|\mathrel{} y_i\left(\mathbf{\hat{h}}^{\dagger}\mathbf{\hat{x}}_i+b\right) < 1 \right\} \nonumber
\end{eqnarray}
We estimate the step-size $\eta_{t}$ for gradient descent by exact line search i.e.,
\begin{eqnarray}
\min_{\eta} f\left(\mathbf{\hat{h}}_{t}+\eta\nabla f\left(\mathbf{\hat{h}}_{t}\right)\right)
\end{eqnarray}
\noindent which we solve by a 1-D Newton method. The full algorithm is described in Algorithm \ref{alg:Prox_mmcf}.

\begin{algorithm}[!ht]
\protect\caption{\label{alg:Prox_unconstrained}Accelerated Proximal Gradient Descent for Unconstrained Filters (ZAMOSSE).}
\begin{algorithmic}
\STATE \textbf{Compute} $\hat{\mathbf{h}}_{conv}$ (conventional CF solution)
\STATE \textbf{Initialize} $\hat{\mathbf{v}}_{0} \leftarrow prox(\hat{\mathbf{h}}_{conv})$ and $\hat{\mathbf{w}}_{0} \leftarrow prox(\hat{\mathbf{h}}_{conv})$
\REPEAT 
\STATE $\nabla f(\hat{\mathbf{w}}_{t})=2\mathbf{\hat{T}}\mathbf{\hat{w}}_{t}-2\mathbf{\hat{p}}$
\STATE $\eta_{t} \leftarrow \mbox{exact line search}$
\STATE $\mathbf{\hat{v}}_{t+1} = prox\left(\mathbf{\hat{w}}_{t}-\eta_{t}\nabla f(\hat{\mathbf{w}}_{t})\right)$
\STATE $\mathbf{\hat{w}}_{t+1} = \mathbf{\hat{v}}_{t+1} + \left(\frac{t-1}{t+2}\right)(\mathbf{\hat{v}}_{t+1}-\mathbf{\hat{v}}_{t})$
\UNTIL $\frac{|f(\mathbf{\hat{v}}_{t+1})-f(\mathbf{\hat{v}}_{t})|}{|f(\mathbf{\hat{v}}_{t})|} < \epsilon$
\STATE \textbf{Output} $\mathbf{\hat{h}} \leftarrow \mathbf{\hat{v}}_{t+1}$
\STATE 
\STATE \textbf{Function} $prox(\mathbf{\hat{h}})$
\end{algorithmic}

\begin{algorithmic}[1]
\STATE $\mathbf{h}\leftarrow\mathcal{F}^{-1}(\mathbf{\hat{h}})$
\STATE $\mathbf{h}\leftarrow P_{\Omega}({\mathbf{h}})$
\STATE $\mathbf{\hat{h}}\leftarrow\mathcal{F}(\mathbf{h})$
\end{algorithmic}
\end{algorithm}

\begin{algorithm}[!ht]
\protect\caption{\label{alg:Prox_constrained}Accelerated Proximal Gradient Descent for Equality Constrained Filters (ZAMACE and ZAOTSDF).}
\begin{algorithmic}
\STATE \textbf{Compute} $\hat{\mathbf{h}}_{conv}$ (conventional CF solution)
\STATE \textbf{Initialize} $\hat{\mathbf{v}}_{0} \leftarrow prox(\hat{\mathbf{h}}_{conv})$ and $\hat{\mathbf{w}}_{0} \leftarrow prox(\hat{\mathbf{h}}_{conv})$
\REPEAT
\STATE $\nabla f(\hat{\mathbf{w}}_{t})=2\mathbf{\hat{T}}\mathbf{\hat{w}}_{t}-2\mathbf{\hat{p}}$
\STATE $\eta_{t} \leftarrow \mbox{exact line search}$
\STATE $\mathbf{\hat{v}}_{t+1} = prox\left(\mathbf{\hat{w}}_{t}-\eta_{t}\nabla f(\hat{\mathbf{w}}_{t})\right)$
\STATE $\mathbf{\hat{w}}_{t+1} = \mathbf{\hat{v}}_{t+1} + \left(\frac{t-1}{t+2}\right)(\mathbf{\hat{v}}_{t+1}-\mathbf{\hat{v}}_{t})$
\UNTIL $\frac{|f(\mathbf{\hat{v}}_{t+1})-f(\mathbf{\hat{v}}_{t})|}{|f(\mathbf{\hat{v}}_{t})|} < \epsilon$
\STATE \textbf{Output} $\mathbf{\hat{h}} \leftarrow \mathbf{\hat{v}}_{t+1}$
\STATE
\STATE \textbf{Function} $prox(\mathbf{\hat{h}})$
\end{algorithmic}

\begin{algorithmic}[1]
\STATE $\mathbf{h} \leftarrow \mathcal{F}^{-1}(\mathbf{\hat{h}})$
\STATE $\mathbf{h}\leftarrow P_{\Omega}({\mathbf{h}})$
\STATE $\mathbf{h}^{\#} \leftarrow P_{C}({\mathbf{h}})$
\STATE $\mathbf{h}_{\Delta}=\mathbf{X}(\mathbf{X}^{T}\mathbf{X})^{-1}(\mathbf{u}-\mathbf{X}^{T}\mathbf{h}^{\#})$
\STATE $\mathbf{h}^{\#}\leftarrow\mathbf{h}^{\#}+\mathbf{h}_{\Delta}$
\STATE $\mathbf{h}^{\#} \leftarrow P_{Z}({\mathbf{h}})$ 
\STATE $\mathbf{\hat{h}}\leftarrow\mathcal{F}(\mathbf{h})$
\end{algorithmic}
\end{algorithm}

\begin{algorithm}[!ht]
\protect\caption{\label{alg:Prox_mmcf}Accelerated Proximal Gradient Descent for Inequality Constrained Filters (ZAMMCF).}
\begin{algorithmic}
\STATE \textbf{Compute} $\hat{\mathbf{h}}_{conv}$ (any conventional CF solution)
\STATE \textbf{Initialize} $\hat{\mathbf{v}}_{0} \leftarrow prox(\hat{\mathbf{h}}_{conv})$ and $\hat{\mathbf{w}}_{0} \leftarrow prox(\hat{\mathbf{h}}_{conv})$
\REPEAT
\STATE $\Omega_{sv} \leftarrow \left\{i \in (1,\dots,n) \mathrel{}\middle|\mathrel{} \left[1-y_i(\mathbf{\hat{w}}_{k}^{\dagger}\mathbf{\hat{x}}_i+b)\right] < 0 \right\}$
\STATE $\nabla f(\hat{\mathbf{w}}_{k})=\lambda\mathbf{\hat{T}\hat{w}}_{k} - \frac{1}{L}\sum_{i \in \Omega_{sv}}y_i\mathbf{\hat{x}}_i(1-y_i(\mathbf{\hat{w}}_{k}^{\dagger}\mathbf{\hat{x}}_i+b))$
\STATE $\eta_{t} \leftarrow \mbox{exact line search}$
\STATE $\mathbf{\hat{v}}_{t+1} = prox\left(\mathbf{\hat{w}}_{t}-\eta_{t}\nabla f(\hat{\mathbf{w}}_{t})\right)$
\STATE $\mathbf{\hat{w}}_{t+1} = \mathbf{\hat{v}}_{t+1} + \left(\frac{t-1}{t+2}\right)(\mathbf{\hat{v}}_{t+1}-\mathbf{\hat{v}}_{t})$
\UNTIL $\frac{|f(\mathbf{\hat{v}}_{t+1})-f(\mathbf{\hat{v}}_{t})|}{|f(\mathbf{\hat{v}}_{t})|} < \epsilon$
\STATE \textbf{Output} $\mathbf{\hat{h}} \leftarrow \mathbf{\hat{v}}_{t+1}$
\STATE
\STATE \textbf{Function} $prox(\mathbf{\hat{h}})$
\end{algorithmic}

\begin{algorithmic}[1]
\STATE $\mathbf{h}\leftarrow\mathcal{F}^{-1}(\mathbf{\hat{h}})$
\STATE $\mathbf{h}\leftarrow P_{\Omega}({\mathbf{h}})$
\STATE $\mathbf{\hat{h}}\leftarrow\mathcal{F}(\mathbf{h})$
\end{algorithmic}
\end{algorithm}

Our proximal gradient descent approach ensures that the ZA constraints (and peak constraints, for constrained filters) are satisfied. This is done in the spatial domain, rather than forming matrix \textbf{$\mathbf{\hat{A}}$ }(or matrix \textbf{$\mathbf{\hat{B}}$}, for the constrained case) as in the closed form solution. This is advantageous because we save on the memory needed to compute and store $\mathbf{\hat{A}}$ (or $\mathbf{\hat{B}}$) and the subsequent computational resources needed to solve large systems of equations. The result is a memory-stable, low complexity solution that allows for fast, efficient and scalable computation of ZACFs.

\subsection{\label{sub:Computational-Experiments}Computational Experiments}
\begin{figure}[h]
\begin{center}
\includegraphics[width=0.95\columnwidth]{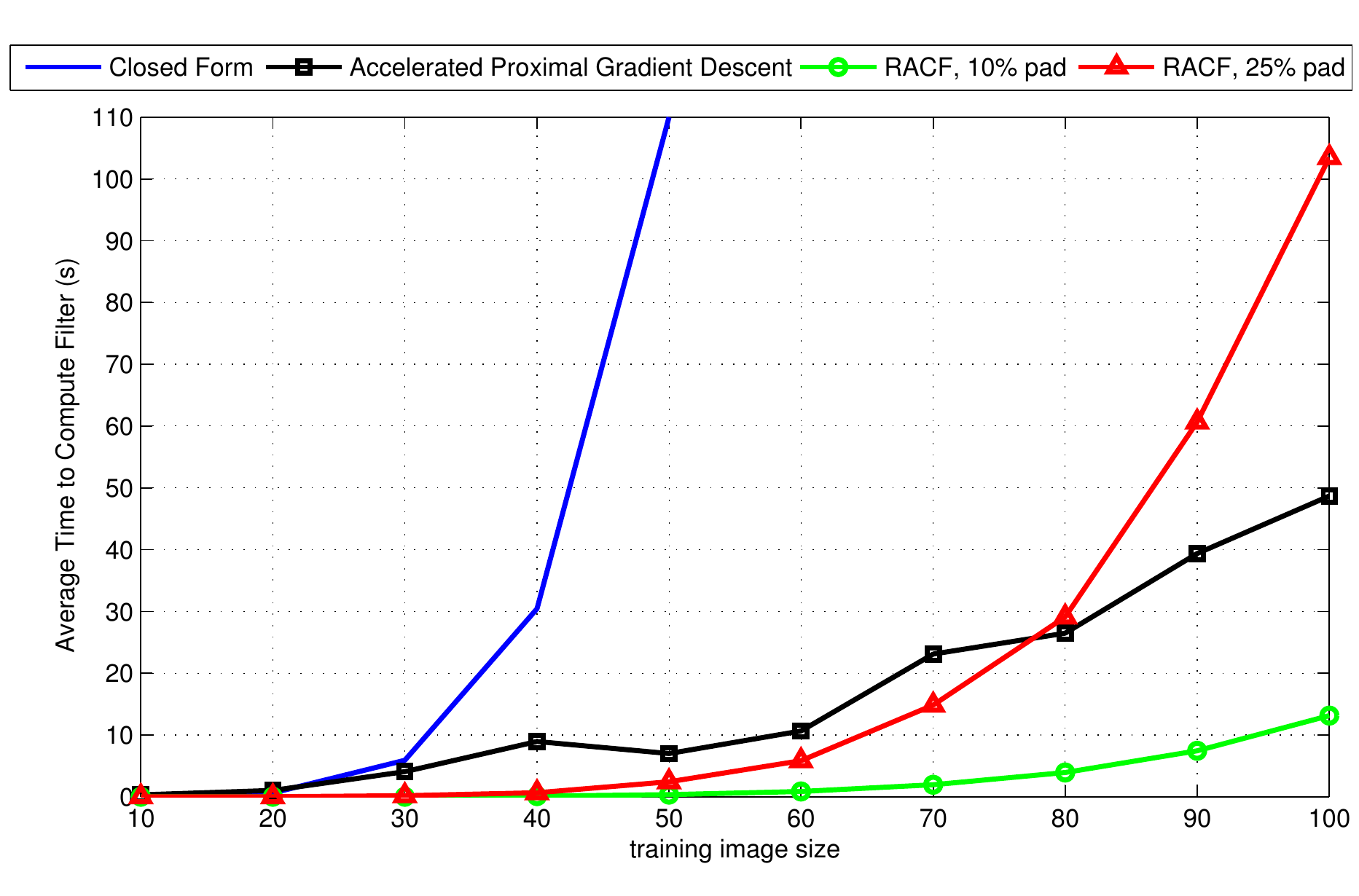}
\protect\caption{\label{fig:prox_grad_timing}Computational times for filter design
comparing RACF and the proximal gradient method compared to a closed
form solution. The ``training image size" on the horizontal axis refers to $N$ where the training image size is $N\times N$.}
\end{center}
\end{figure}

To demonstrate the benefits of RACF and the proximal gradient descent methods, we perform a computational analysis in this section (see Section \ref{sec:Experimental-Results} for accuracy comparisons) to compare both methods to the closed-form ZACF implementation. For RACF, we pad with a number of zeros equal to $10\%$ or $25\%$ of the training image size, and refer to these results as RACF-10\% and RACF-25\%, respectively. For the proximal gradient method, we compute the step size $\eta_t$ via exact line search. We initialize the filter ($\hat{\mathbf{h}}_{0}$) as the conventional filter design subjected to the $prox()$ operator. We use a stopping condition $\frac{\left|f(\mathbf{\hat{v}}_{k})-f(\mathbf{\hat{v}}_{k-1})\right|}{\left|f(\mathbf{\hat{v}}_{k-1})\right|}<10^{-10}$ to terminate the optimization. In this experiment, we used a training size of $9$ images from the AT\&T (ORL) Face Dataset \cite{ORL-ATT-dataset}; we vary the resolution and crop the training images so that they are square. The platform for this experiment was a desktop running Matlab
2011a with Windows 7, an Intel Core i7-2600 CPU (3.4 GHz), and 16 GB of RAM. We show the results of our experiment in Fig. \ref{fig:prox_grad_timing}.
We only compute the closed form solution for a resolution up to $50\times50$, as this is near the memory limit of our machine. Note that, while the fastest method is RACF-10\%, we will show in Section \ref{sec:Experimental-Results} that RACF-25\% will perform better from a recognition performance perspective. However, recognition performance is dependent on the image size and the training set, so experimentation is needed before choosing the amount of padding used for RACF. A downside of the RACF method is that it can still require a large amount of memory for larger pad sizes. Therefore, we prefer the proximal gradient method because it is more efficient from a memory perspective. In the next section, we show recognition performance on the AT\&T/ORL dataset using both methods to illustrate these points further.

\section{\label{sec:Experimental-Results}Recognition Performance}
In this section, we report recognition performance of ZACFs. We emphasize that the ZA constraints introduced in this paper are a \textit{fundamental} and \textit{significant} improvement to the formulation of CFs. We have observed that ZACFs consistently perform better than their original formulation (aliased) counterparts. This is because the original CF designs did not account for circular correlation and are fundamentally flawed. This paper's focus is to illustrate the performance improvement that is realized by adding ZA constraints to CF designs, and \textit{not} to compare different types of CFs to each other. Further, we are also \emph{not} claiming that CFs are the best approach for all object recognition problems. To investigate the performance of ZACFs, we apply them to four types of pattern recognition tasks, namely face recognition, automatic target recognition (ATR), eye localization and object detection.

\subsection{Face Recognition}
\begin{table*}[!ht]
\protect\caption{\label{tab:Performance-Comparison-ORL}Performance Comparison of Baseline CFs and ZACFs on the ORL Dataset}
\begin{center}
\resizebox{\textwidth}{!}{%
\renewcommand{\tabcolsep}{1pt}
\begin{tabular}{cc>{\centering}p{2.54cm}>{\centering}p{2.54cm}>{\centering}p{2.54cm}>{\centering}p{2.54cm}>{\centering}p{2.54cm}>{\centering}p{2.54cm}>{\centering}p{2.54cm}}
&  & Baseline CF & ZACF\linebreak(closed form) & ZACF\linebreak(proximal gradient) & RACF\linebreak($p=15$) & RACF\linebreak ($p=20$) & RACF\linebreak($p=25$) & RACF\linebreak($p=30$)\tabularnewline
\hline 
\hline 
\multirow{2}{*}{\textbf{OTSDF}} & EER & 10.89\% & 7.54\%  & 7.65\%  & 11.75\%  & 8.45\%  & 7.21\%  & 7.5\% \tabularnewline
\cline{2-9} 
& Rank-1 ID & 83.75\% & 89.5\%  & 89.5\%  & 70.75\%  & 81\%  & 85.75\%  & 87.75\% \tabularnewline
\hline 
\multirow{2}{*}{\textbf{MOSSE}} & EER & 11.75\%  & 7.94\% & 7.94\%  & 12.25\%  & 9.25\%  & 8.75\%  & 8.5\% \tabularnewline
\cline{2-9} 
& Rank-1 ID & 84.25\%  & 88.25\% & 88\%  & 72.75\%  & 83\%  & 85.75\%  & 87.5\% \tabularnewline
\hline 
\multirow{2}{*}{\textbf{MMCF}} & EER & 11.73\% & 8\% & 8.21\% & 12\% & 8.930\% & 8.75\% & 8.5\%\tabularnewline
\cline{2-9} 
& Rank-1 ID & 84.25\% & 88.5\% & 88.3\% & 73\% & 83\% & 86\% & 87.5\%\tabularnewline
\cline{2-9} 
\end{tabular}}
\end{center}
\end{table*}
We apply CFs to two different face recognition datasets: the AT\&T
(ORL) Database of Faces \cite{ORL-ATT-dataset} and the face recognition
grand challenge (FRGC) dataset \cite{FRGC}.
\subsubsection{AT\&T Database of Faces}
The AT\&T Database of Faces (formerly the ORL Database of Faces) \cite{ORL-ATT-dataset}
contains face images of size $112\times92$. There are $40$ subjects
with $10$ images each. We test using a leave-one-out cross validation
approach. For each experiment, $9$ training images were used to train
one CF for each subject. These filters were then tested on the remaining
image for all subjects. This resulted in a total of $40$ test images
(one from each class) that were tested with $40$ filters each, for
a total of $1600$ correlations per cross validation step. This was
repeated $10$ times. For each correlation output, we calculate peak
to correlation energy (PCE), which is a measure of peak sharpness
\cite{Kumars_Book}.  
We perform classification on each of the test images and compute the rank-1 ID rate. We compute the ZACFs in several ways: 1) closed form solution, 2) accelerated proximal gradient descent, and 3) various versions of the RACF (using different amounts of zero-padding). The purpose of this comparison is to demonstrate the effectiveness of the computational solutions in Section \ref{sec:Computational-Considerations}. Note that computing the closed form solution is extremely difficult and requires many hours on high-end computers. Therefore we have not done an exhaustive comparison on every possible set of parameters available to each filter formulation. The results of our validation for the MACE, OTSDF, MOSSE, and MMCF filters are shown in Table \ref{tab:Performance-Comparison-ORL}. We have used a delta function for the desired correlation output for MOSSE and MMCF.

We observe that the ZACF outperforms the baseline (conventional) CFs. Note in Table \ref{tab:Performance-Comparison-ORL} that the proximal gradient gives recognition performance close to that from the closed form solution. This is also true for RACF provided that the amount of zero-padding, $q$, is sufficiently large.

\subsubsection{FRGC}

The FRGC dataset \cite{FRGC} contains face images of resolution $128\times128$. We use $410$ subjects from the test portion of the data, removing
subjects with less than $8$ images per subject. For the data we use, the number of images from each subject varies from $8$ (minimum) to $88$ (maximum) with a mean of $~39$ images per subject. We form three training and test sets by randomly selecting $25\%$ of each class for training with the remaining $75\%$ used for testing. We then build one CF per subject and apply every CF to every test image. Like the ORL dataset, we present our results in terms of equal error rate (EER) and Rank-1 ID rate in Table \ref{tab:Performance-Comparison-FRGC}. Due to the training image size, we compute the ZACF with the accelerated proximal gradient approach for OTSDF, MACE, MOSSE, and MMCF. In the table, we refer to these results simply as ``ZACF'' for simplicity. As in the ORL experiments, we use a delta function for the desired correlation output, if required. Note that each ZACF achieves both a higher Rank-1 ID and a lower EER than the corresponding baseline filter.
\begin{table}[!h]
\begin{center}
\protect\caption{\label{tab:Performance-Comparison-FRGC}Performance Comparison of
Baseline CFs and ZACFs on the FRGC Dataset}
\begin{tabular}{cc>{\centering}p{2.54cm}>{\centering}p{2.54cm}}
&  & Baseline CF & ZACF\tabularnewline
\hline 
\hline 
\multirow{2}{*}{\textbf{OTSDF}} & EER & 2.43\% & 1.79\%\tabularnewline
\cline{2-4} 
& Rank-1 ID & 93.23\% & 94.95\%\tabularnewline
\hline 
\multirow{2}{*}{\textbf{MACE}} & EER & 15.23\% & 9.46\%\tabularnewline
\cline{2-4} 
& Rank-1 ID & 52.73\% & 78.98\%\tabularnewline
\hline 
\multirow{2}{*}{\textbf{MOSSE}} & EER & 7.35\% & 5.02\%\tabularnewline
\cline{2-4} 
& Rank-1 ID & 86.87\% & 93.78\%\tabularnewline
\hline 
\multirow{2}{*}{\textbf{MMCF}} & EER & 2.52\% & 1.80\%\tabularnewline
\cline{2-4} 
& Rank-1 ID & 91.96\% & 94.52\%\tabularnewline
\hline 
\end{tabular}
\end{center}
\end{table}

\subsection{ATR Algorithm Development Image Database}

\begin{figure}[!h]
\begin{centering}
\subfigure[Pickup]{\includegraphics[scale=0.135]{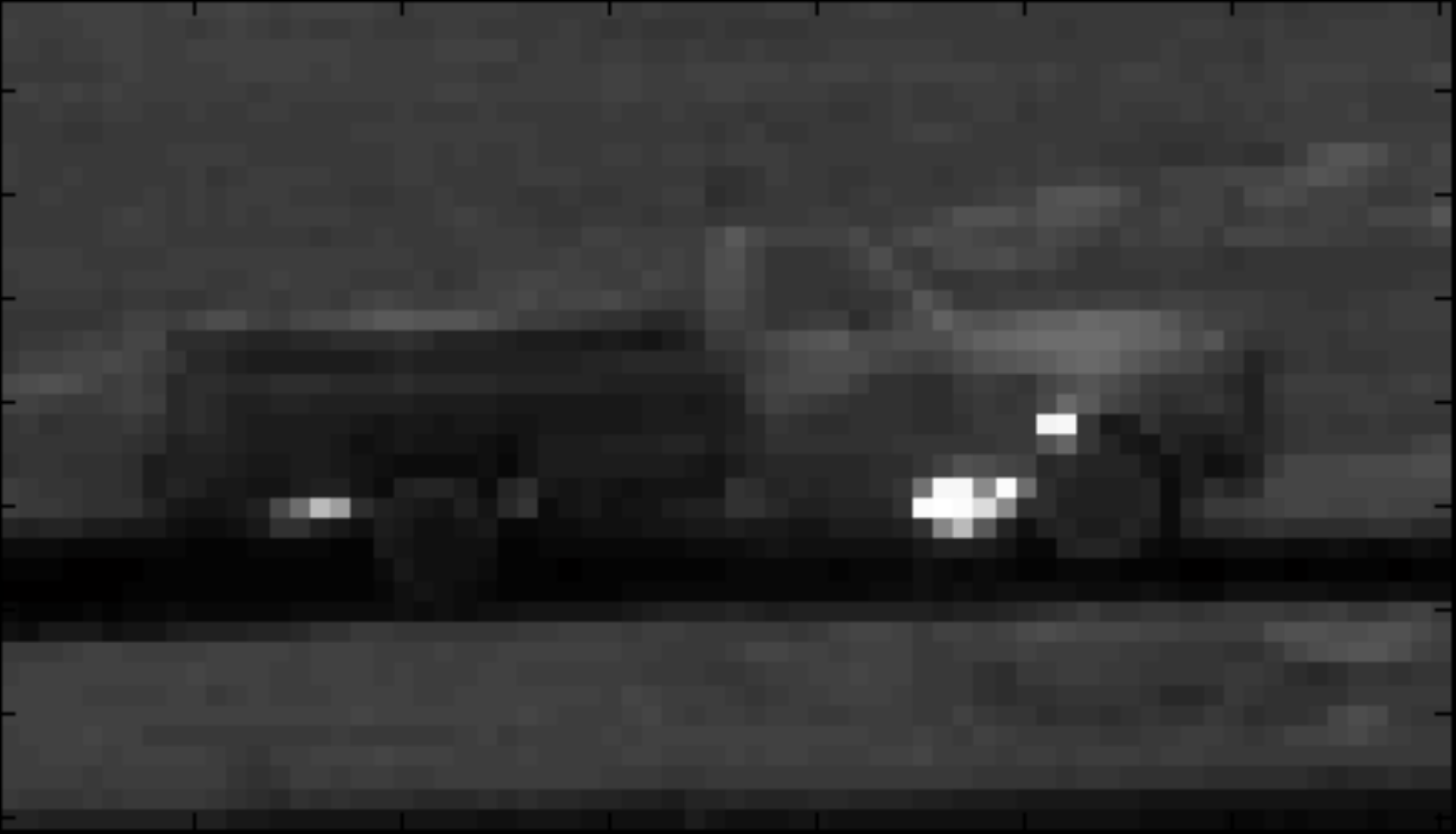}}
\subfigure[SUV]{\includegraphics[scale=0.135]{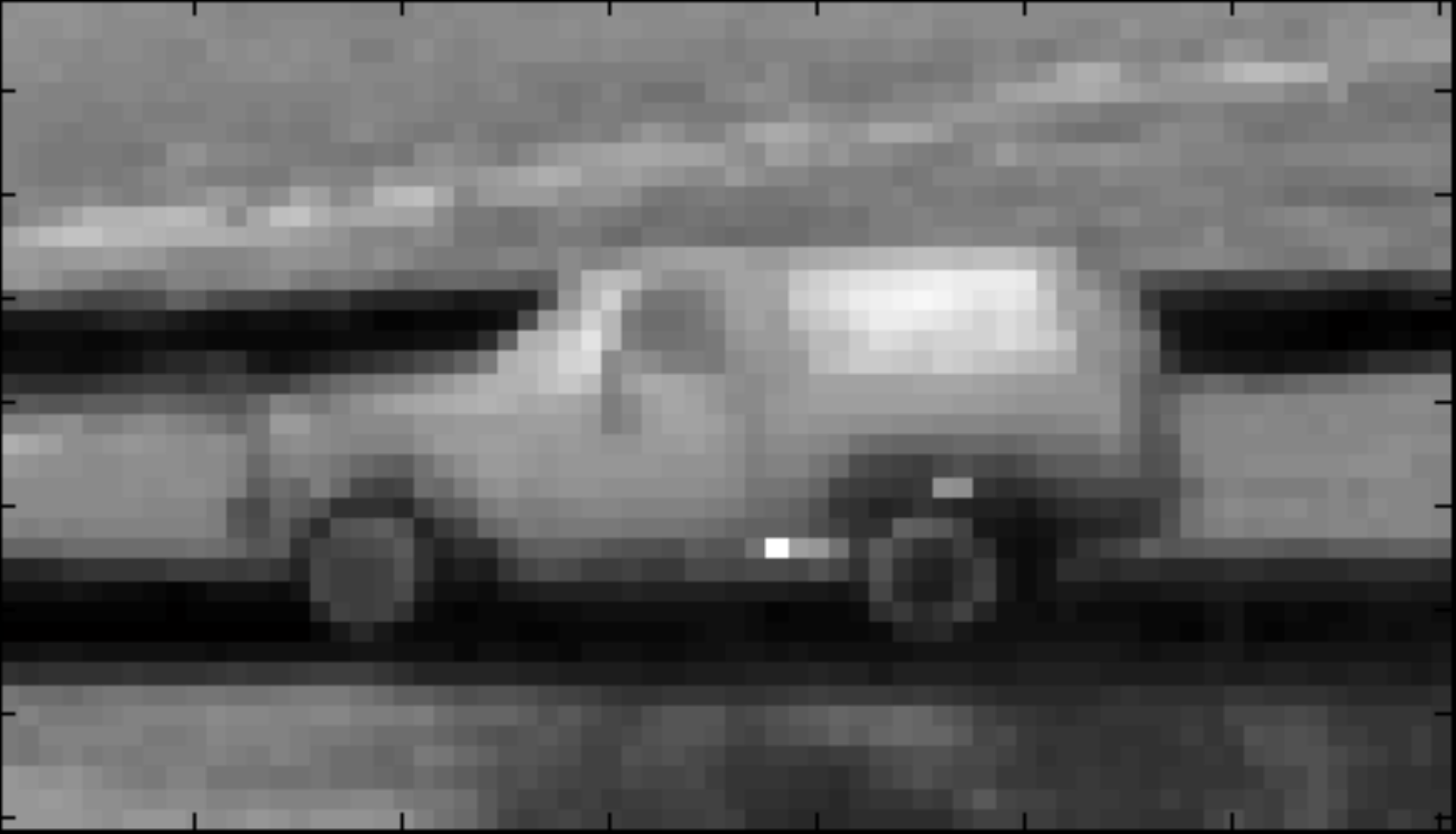}}
\subfigure[BTR70]{\includegraphics[scale=0.135]{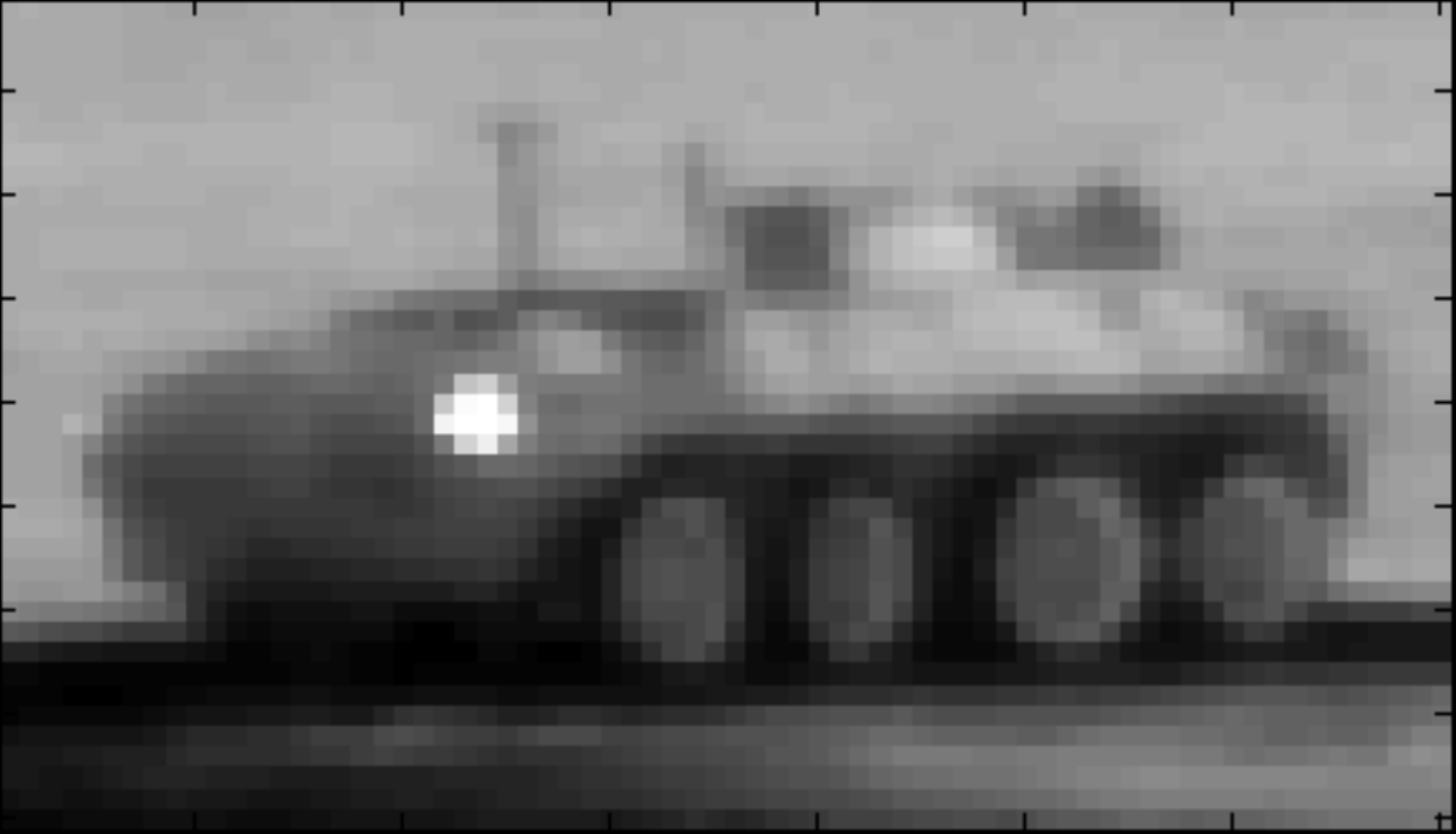}}
\subfigure[BRDM2]{\includegraphics[scale=0.135]{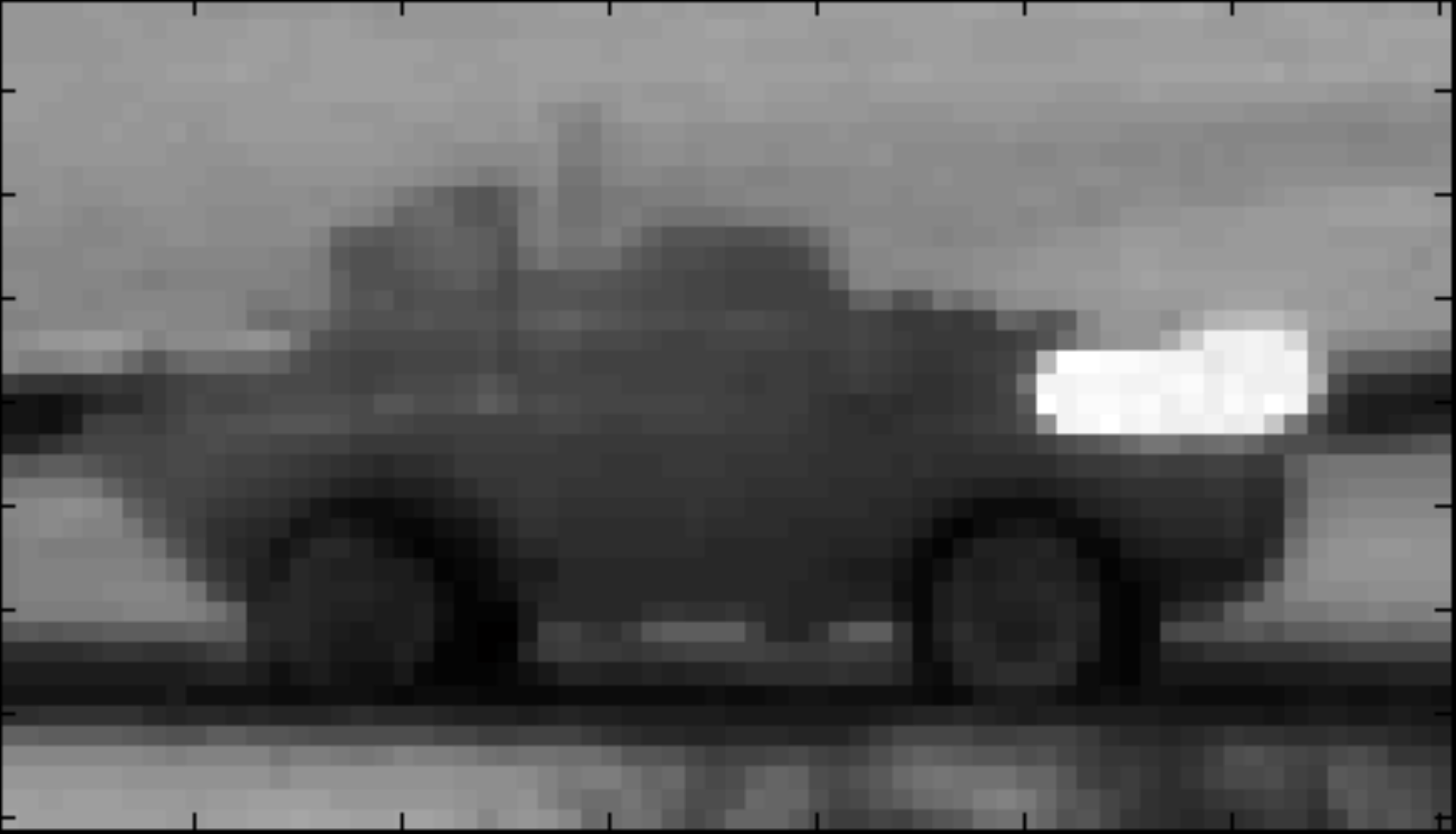}}
\par\end{centering}
\centering{}\subfigure[BMP2]{\includegraphics[scale=0.135]{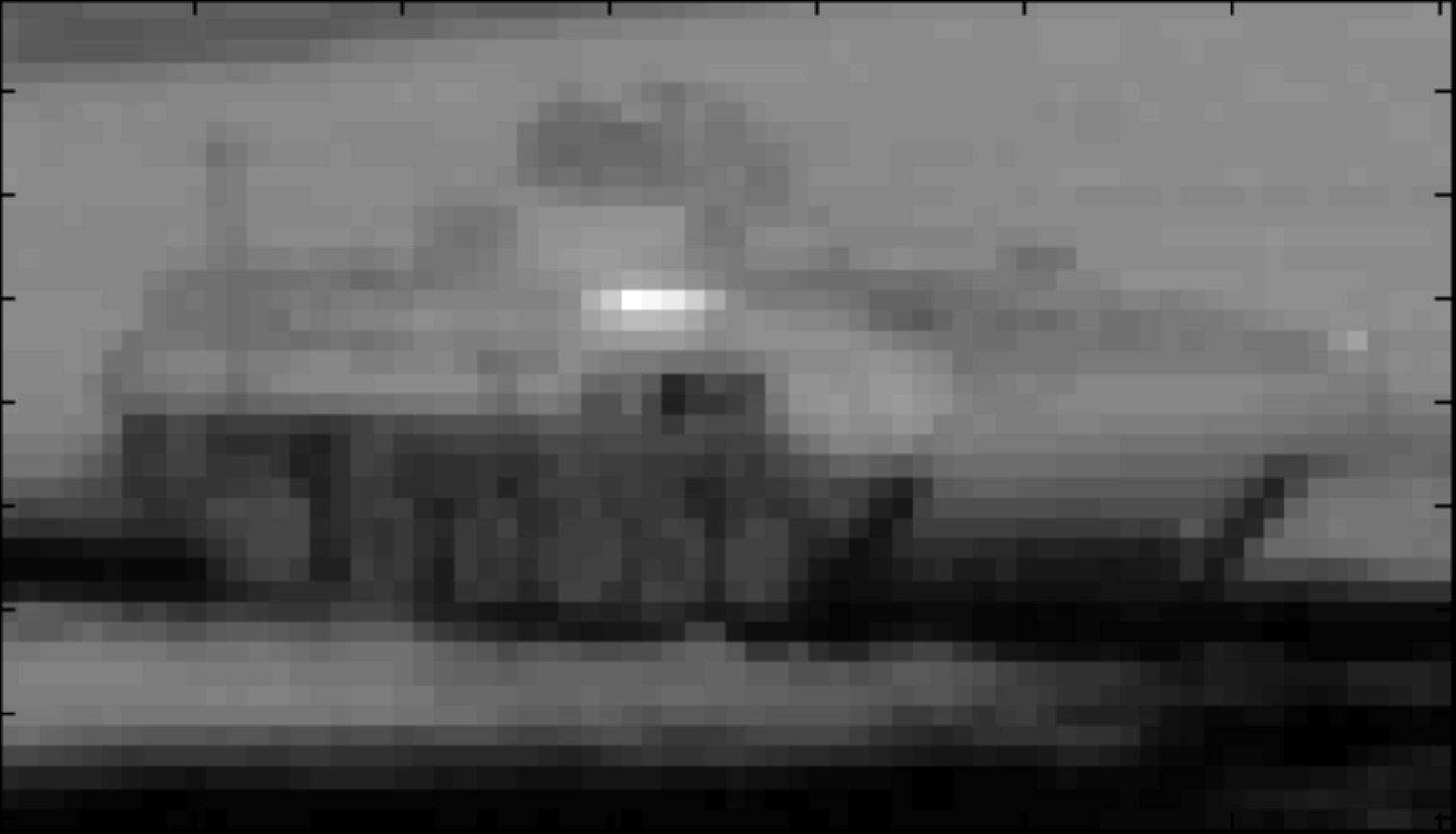}}
\subfigure[T72]{\includegraphics[scale=0.135]{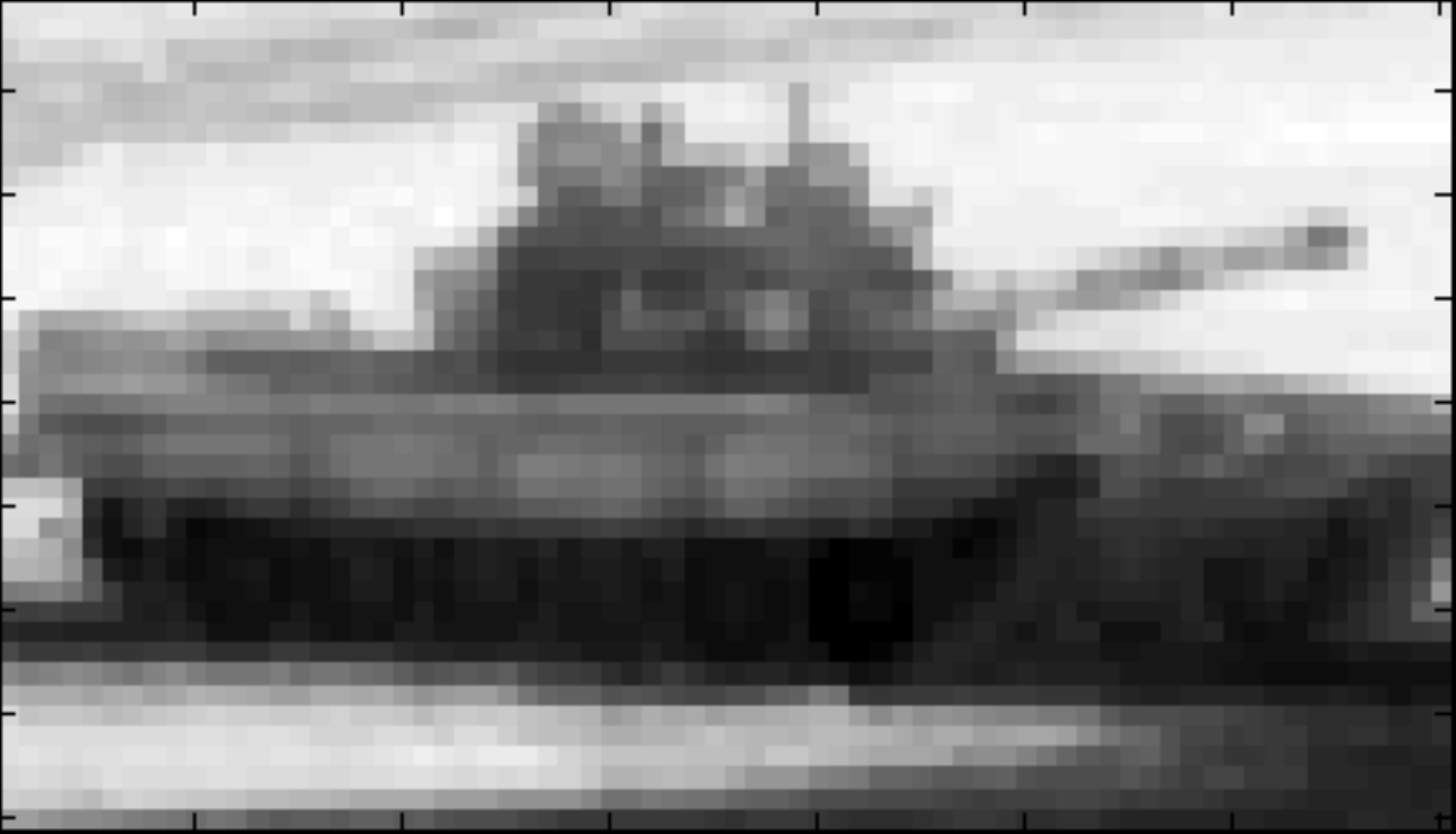}}
\subfigure[ZSU23-4]{\includegraphics[scale=0.135]{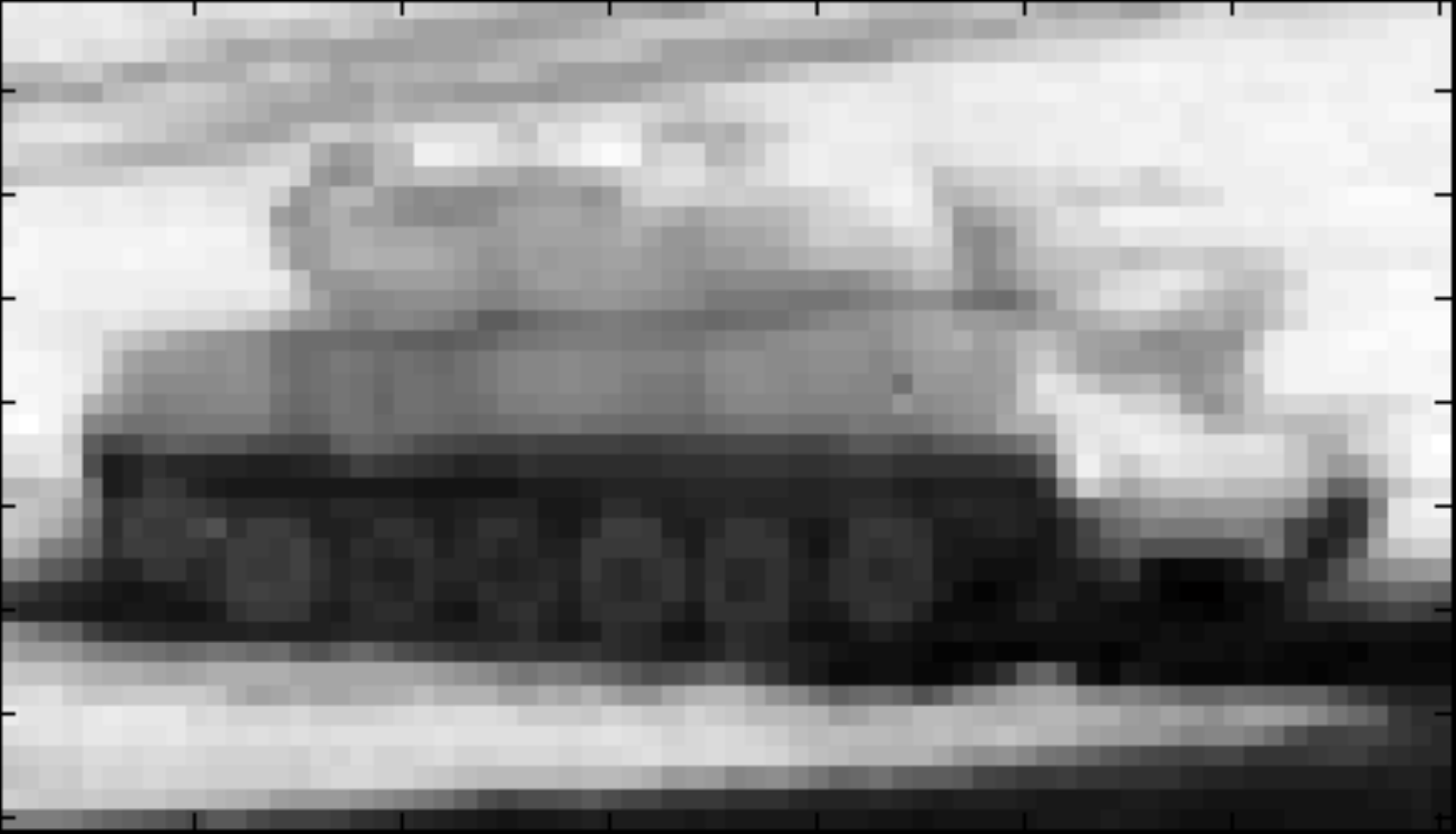}}
\subfigure[2S3]{\includegraphics[scale=0.135]{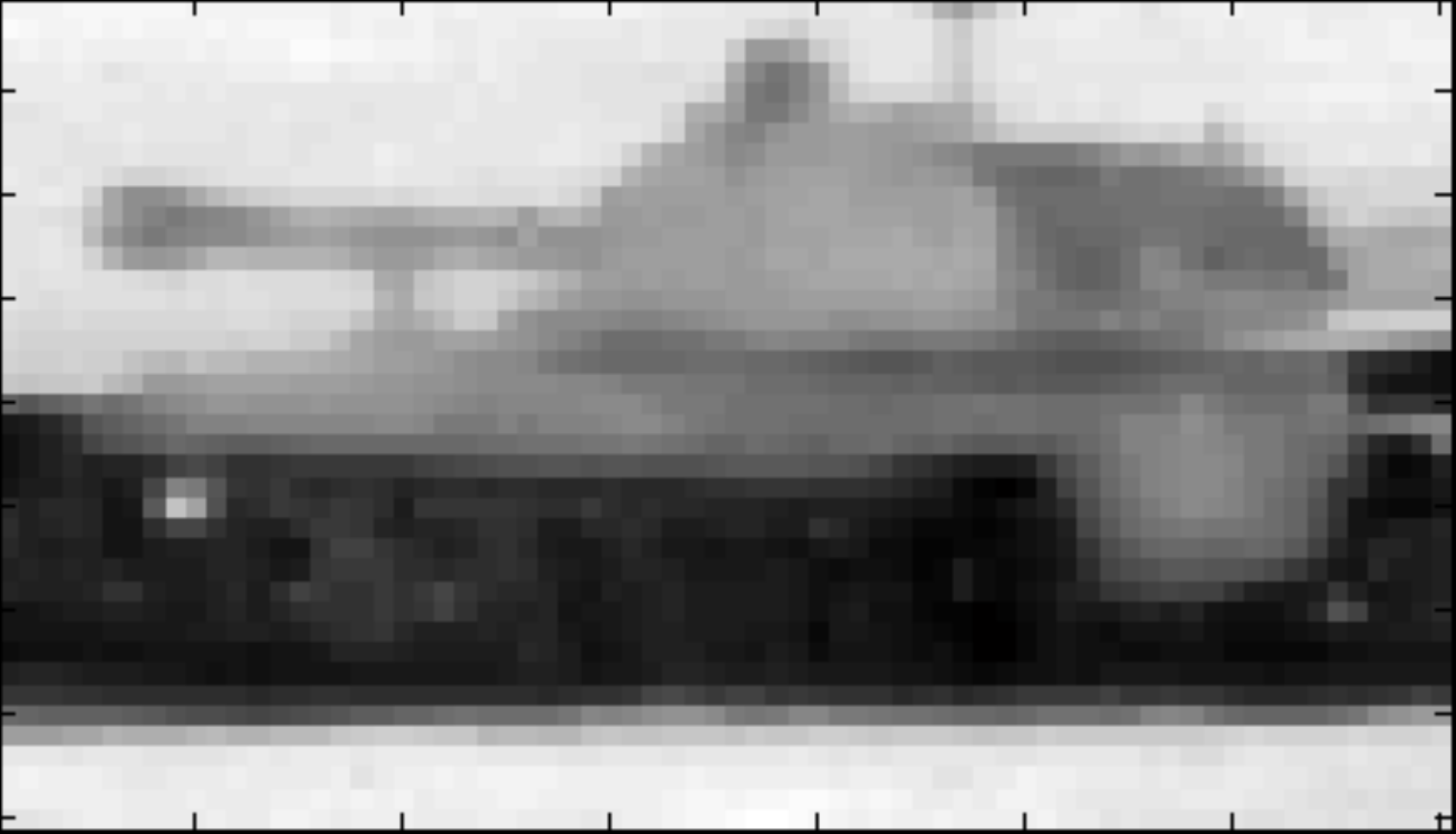}}
\protect\caption{\label{fig:types_of_classes} Example images from different classes of vehicles.}
\end{figure}

\begin{figure}[!h]
\centering{}\includegraphics[width=0.75\columnwidth]{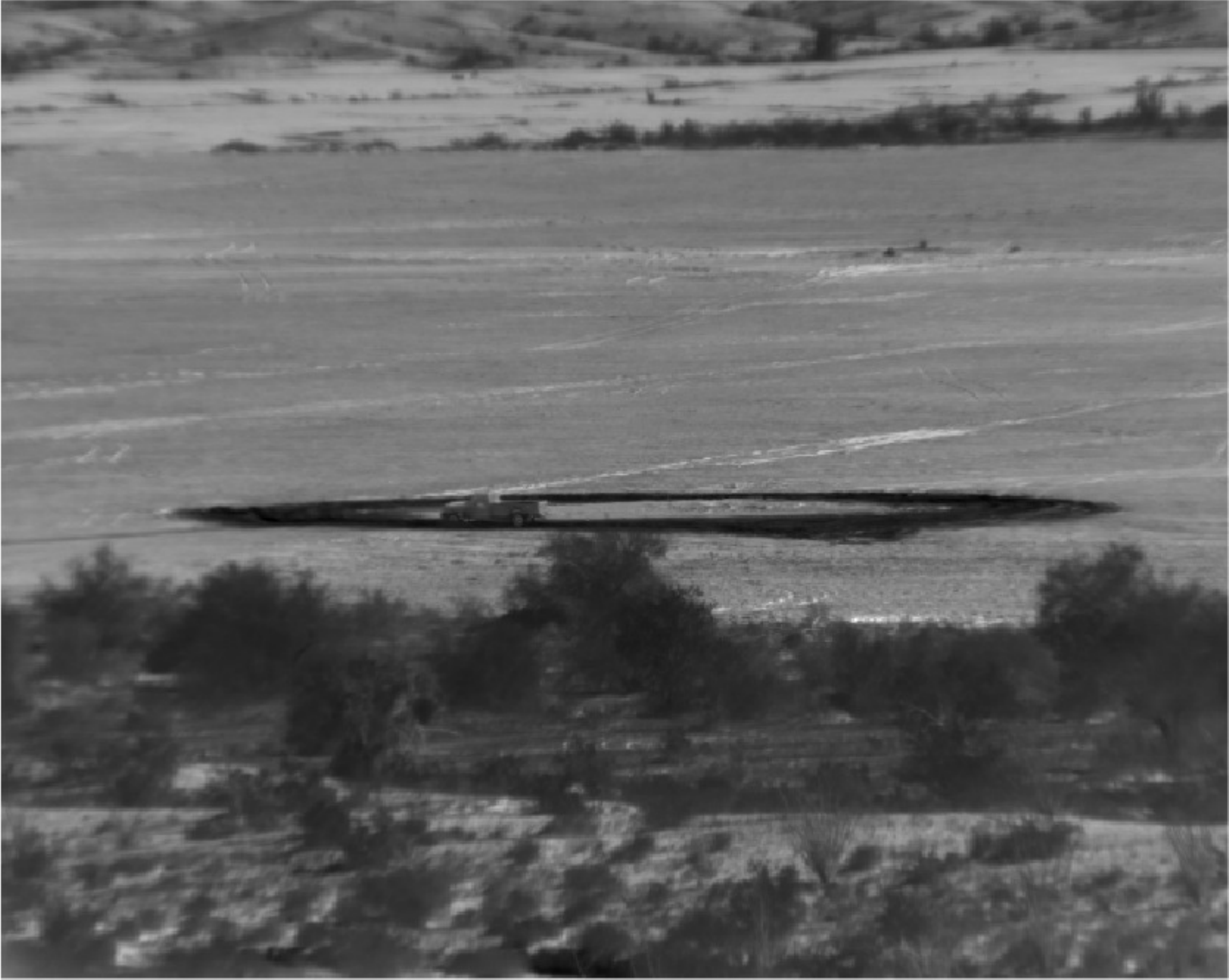}\protect\caption{\label{fig:targetwbg} Vehicle ``Pickup'' and background}
\end{figure}

We investigate vehicle recognition (i.e., simultaneous classification and localization) using a set of infrared images (frames from videos) from the \emph{ATR Algorithm Development Image Database} \cite{sensiac10}. This database contains infrared videos of $512\times640$ pixels/frame of eight vehicles (one per video), shown in Fig. \ref{fig:types_of_classes}, taken at multiple ranges during day and night at 30 fps. Note in Fig. \ref{fig:types_of_classes} that some of the vehicles have very similar appearance, making the classification task challenging. In the dataset, these vehicles are driven at $\sim5$ m/s, making a full circle of diameter of about $100$ m, therefore exhibiting $360^{\circ}$ of azimuth rotation. Each video is $1$ minute long, allowing the vehicle to complete at least one full circle. An example frame is shown in Fig. \ref{fig:targetwbg}. Note that the low quality frame and the general background makes the recognition
task challenging.
We design one template per vehicle (the template size was small enough to easily calculate the closed form ZACFs) to classify the vehicle in the presence
of $360^{\circ}$ azimuth variation. We select $20$ positive-class images per filter (manually cropped from the corresponding frames, $N\times M=40\times70$) and $80$ non-overlapping background images as negative class images for training. We select $200$ full frames for testing, verifying that none of the testing frames were used in training. For testing, we correlate the $8$ templates (one per vehicle) with each test image. For each correlation plane, we select the highest value, compute PCE, and assign the test image to the class that gives the highest PCE. If it is the correct class, we declare it as correctly classified. We declare correct localization when the location of the peak from the true class filter is within $20$ and $35$ pixels (i.e., half the size of the template) of the ground truth location in the vertical and horizontal directions, respectively. Finally, we declare a correct recognition when there is \textit{both} correct classification \textit{and} correct localization.

Table \ref{tab:CFall} shows the average classification, localization, and recognition percentages of both the conventional CF and ZACF. Dalal and Triggs \cite{dalal2005histograms} proposed cross-correlating the 2-D template with full training frames, adding the false positives as negative class training images, and retraining the template. We show our results after retraining in Table \ref{tab:CFretraining}. We observe that retraining helps all filters,
but especially the MMCF filter, whose inequality constraints allow unlimited number of training images from two classes. The most important observation is that ZACFs always perform better than or about the same as traditional CFs in classification, localization, and recognition. Note that we treat each frame independently of other frames, i.e., although using a tracker would improve results, the purpose of these experiments is to compare the performance of conventional CFs and ZACFs without additional help from a tracker.

\begin{table}[!ht]
\begin{center}
\protect\caption{\label{tab:CFall}CF Recognition Performance without Retraining, ATR Algorithm
Development Image Database}
\begin{tabular}{ccccccc}
& \multicolumn{2}{c}{Classification} & \multicolumn{2}{c}{Localization} & \multicolumn{2}{c}{Recognition}\tabularnewline
\hline 
\hline 
& Base & ZA & Base & ZA & Base & ZA\tabularnewline
\hline 
MACE & 40.0\% & 51.9\% & 84.1\% & 88.9\% & 36.2\% & 47.5\%\tabularnewline
\hline 
OTSDF & 53.6\% & 62.1\% & 90.3\% & 90.3\% & 52.4\% & 57.4\%\tabularnewline
\hline 
MOSSE & 29.7\% & 32.3\% & 64.5\% & 78.1\% & 26.1\% & 30.6\%\tabularnewline
\hline 
MMCF & 51.1\% & 57.0\% & 87.7\% & 90.0\% & 49.6\% & 52.6\%\tabularnewline
\hline 
\end{tabular}
\end{center}
\end{table}

\begin{table}[!ht]
\begin{center}
\protect\caption{\label{tab:CFretraining}CF Recognition Performance with Retraining, ATR Algorithm
Development Image Database}
\begin{tabular}{ccccccc}
& \multicolumn{2}{c}{Classification} & \multicolumn{2}{c}{Localization} & \multicolumn{2}{c}{Recognition}\tabularnewline
\hline 
\hline 
& Base & ZA & Base & ZA & Base & ZA\tabularnewline
\hline 
MACE & 46.2\% & 57.7\% & 85.1\% & 89.1\% & 42.7\% & 51.9\%\tabularnewline
\hline 
OTSDF & 59.9\% & 62.3\% & 90.7\% & 90.0\% & 58.1\% & 58.1\%\tabularnewline
\hline 
MOSSE & 31.9\% & 33.4\% & 76.5\% & 84.4\% & 31.1\% & 32.0\%\tabularnewline
\hline 
MMCF & 63.9\% & 74.3\% & 95.3\% & 96.1\% & 63.2\% & 73.5\%\tabularnewline
\hline 
\end{tabular}
\end{center}
\end{table}

\subsection{Eye Localization}
\begin{figure}[!ht]
\begin{center}
\includegraphics[height=0.62\columnwidth]{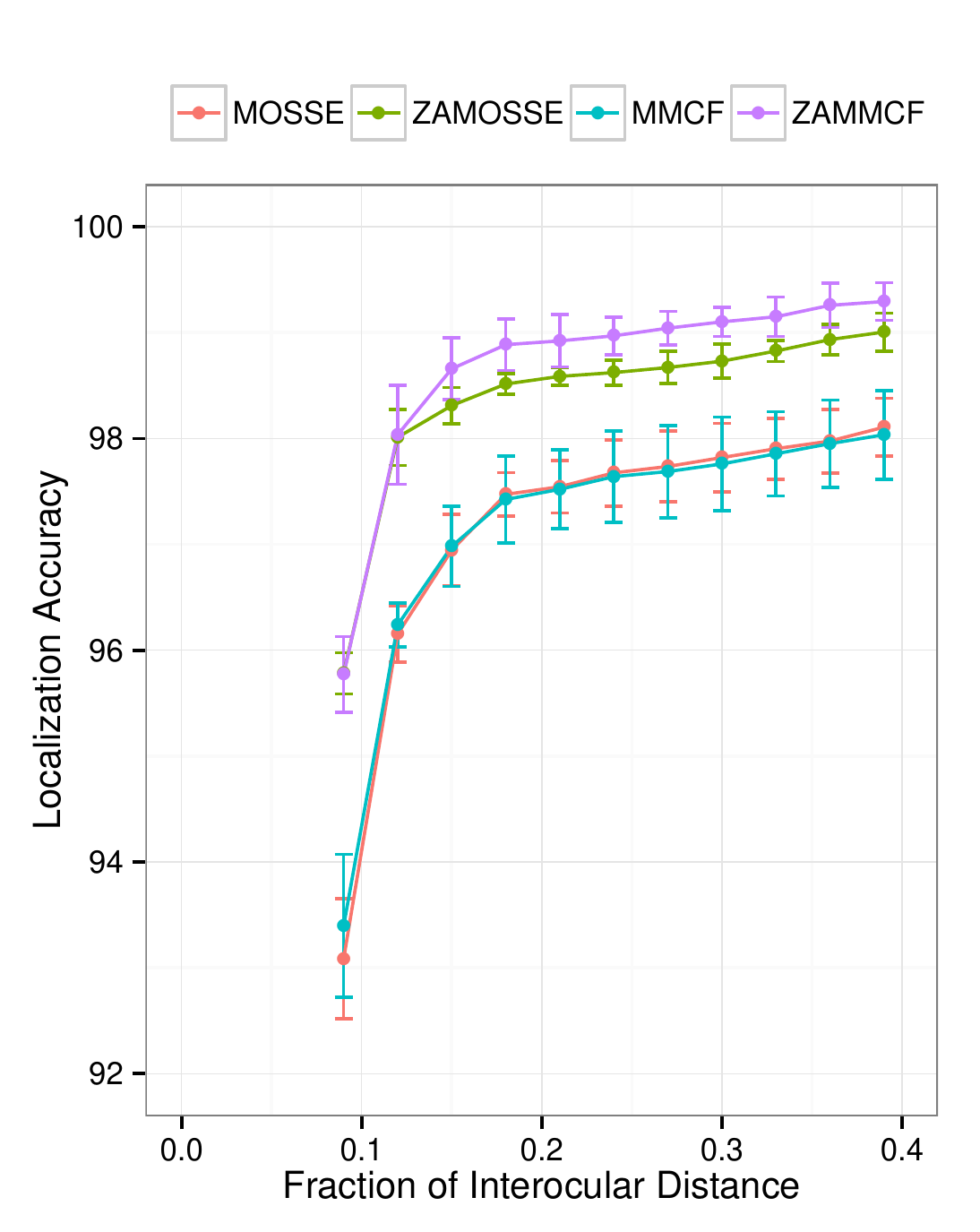}
\includegraphics[height=0.62\columnwidth]{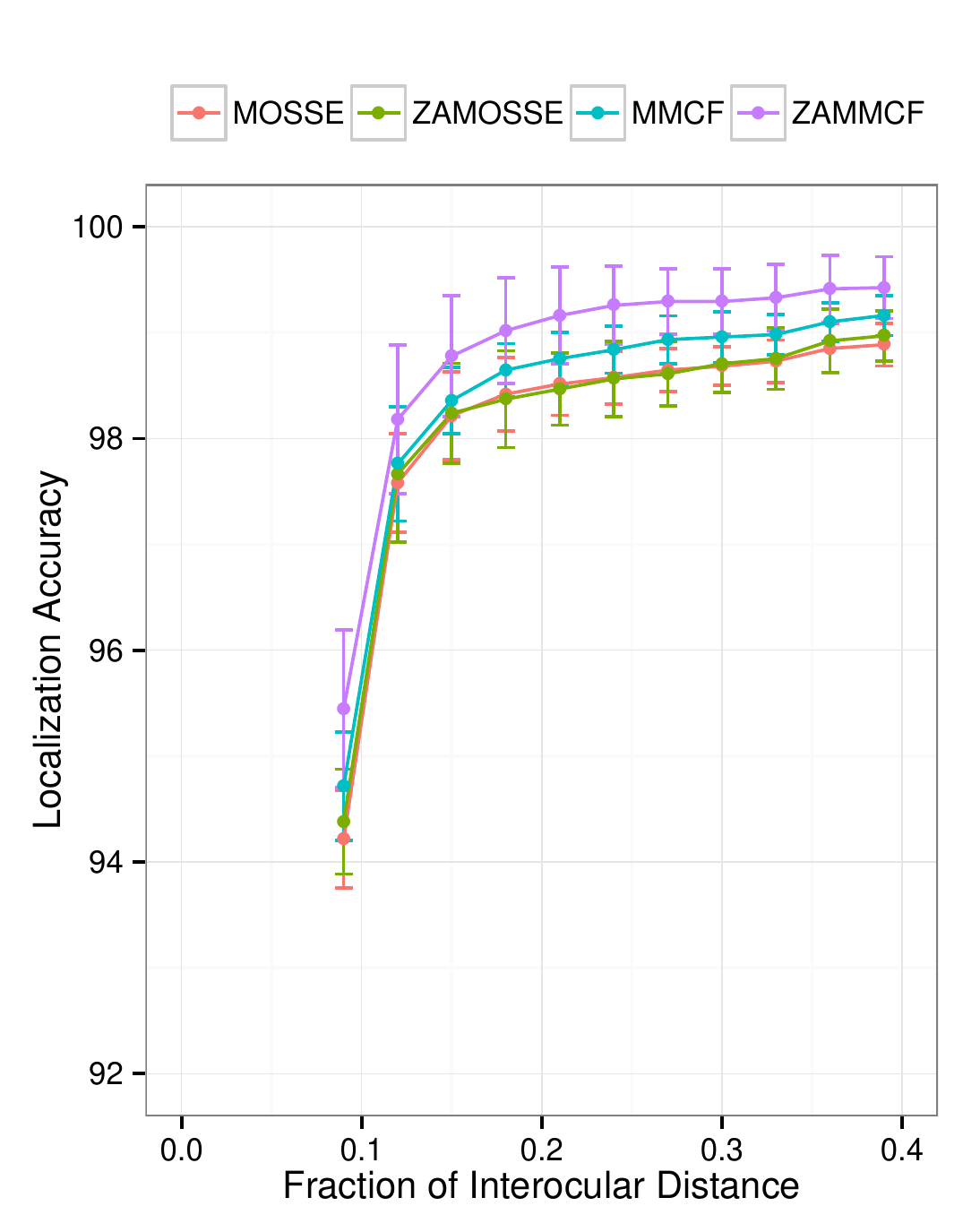}
\protect\caption{\label{fig:eye_results_exp2} Eye localization performance of the left (left) and right (right) eye, respectively, as a function of the inter-ocular distance for the different CF designs. We show the mean performance across five random runs along with the error bars. The ZA filters outperform those designed without explicitly accounting for aliasing.}
\end{center}
\end{figure}

Accurate localization of the eyes in face images is an important component of face, ocular, and iris recognition. In this experiment we consider the task of accurately determining the location of the left and the right eye given a bounding box around a face obtained from a face detector. Since a good face detector makes eye localization overly simple, following the experimental setup outlined in \cite{ASEF}, we make the problem more challenging by introducing errors in face localization. We first center the faces obtained using the \textit{OpenCV} face detector to produce $128\times128$ images with the eyes centered at ($32.0$, $40.0$) and ($96.0$, $40.0$). We then apply a random similarity transform with translation of up
to $\pm4$ pixels, scale factor of up to $1.0\pm0.1$, and rotations of up to $\frac{\pi}{16}$ radians. We used the FERET \cite{phillips2000feret} database for this task, which has about $3400$ images of $1204$ people. We randomly partitioned the database with $512$ images used for training, $675$ for parameter selection by cross-validation, and the rest for testing. The CFs are compared by evaluating the normalized distance defined as, 
\begin{equation}
D=\frac{\left\Vert P-\hat{P}\right\Vert }{\left\Vert P_{l}-P_{r}\right\Vert }
\end{equation}
\noindent where $P$ is the ground truth location, $\hat{P}$ is the predicted location, and $P_{l}$ and $P_{r}$ are the ground truth locations of the left and the right eye, respectively. The point $D=0.1$ corresponds to detecting an object that is approximately the size of a human iris. We train MOSSE, MMCF, ZAMOSSE and ZAMMCF filters using $64\times64$ image patches centered at the left and right eye regions. We note that the MACE/OTSDF design is not suitable for this task since the number of training images is greater than the degrees of freedom in the filter. In such cases MACE/OTSDF designs result in overdetermined linear system and cannot be solved for. Therefore for this experiment we do not show results with MACE and OTSDF designs. We use the proximal gradient descent based algorithm to learn the ZA versions of MOSSE and MMCF. We evaluate the eye localization performance of the CF templates by searching over the entire face image i.e., the scenario where the approximate eye location is not known a-priori. We average the results over $5$ different runs with random partitions for training and testing and random similarity transforms. We compare the eye localization performance as a function of $D$ in Fig. \ref{fig:eye_results_exp2} and Table \ref{table:eye_results} shows the performance of the MOSSE and MMCF filters along with their ZA versions at the operating point $D=0.1$. The filters, while producing a strong response for the correct eye, are sometimes distracted by the wrong eye or other parts of the face. Note that the ZA filters provide a higher localization accuracy than the filters with aliasing. The absence of aliasing results in lower noise levels in the correlation outputs thereby resulting in better localization performance.

\begin{table}
\centering
\caption{Eye Localization Performance (\%)}
\label{table:eye_results}
\begin{tabular}{ccccc}
Eye & MOSSE & ZAMOSSE & MMCF & ZAMMCF \\
\hline
\hline
Right & 94.66 & 97.24 & 95.01 & 97.14 \\
\hline
Left & 96.23 & 96.30 & 96.44 & 97.16\\
\hline
\end{tabular}
\end{table}

\subsection{Object Detection}
\begin{figure}[!h]
\centering
\subfigure[PR Curve]{\includegraphics[width=0.55\linewidth]{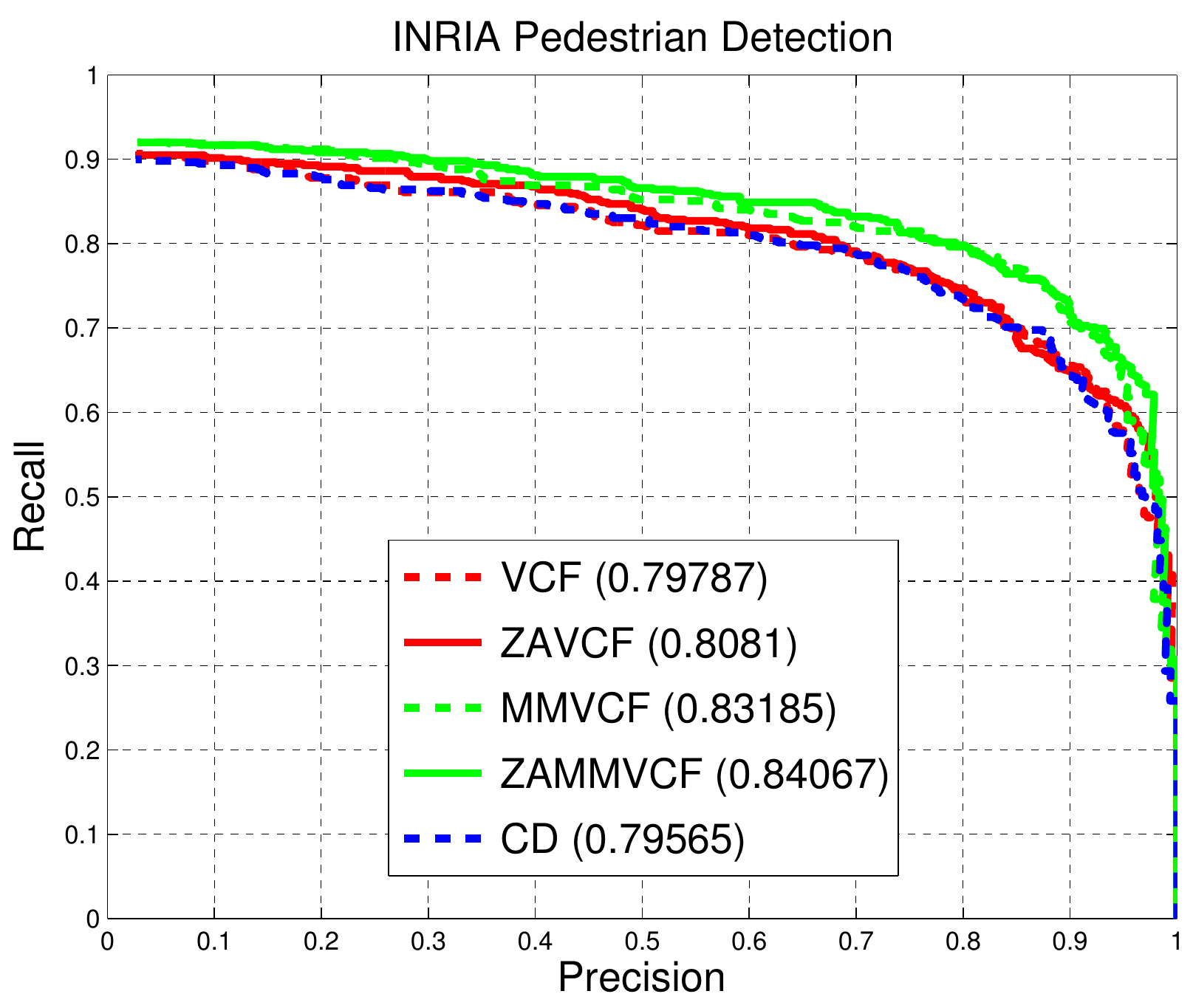}}
\subfigure[]{
\begin{tikzpicture}
	\node[] (a) {\includegraphics[width=0.19\linewidth]{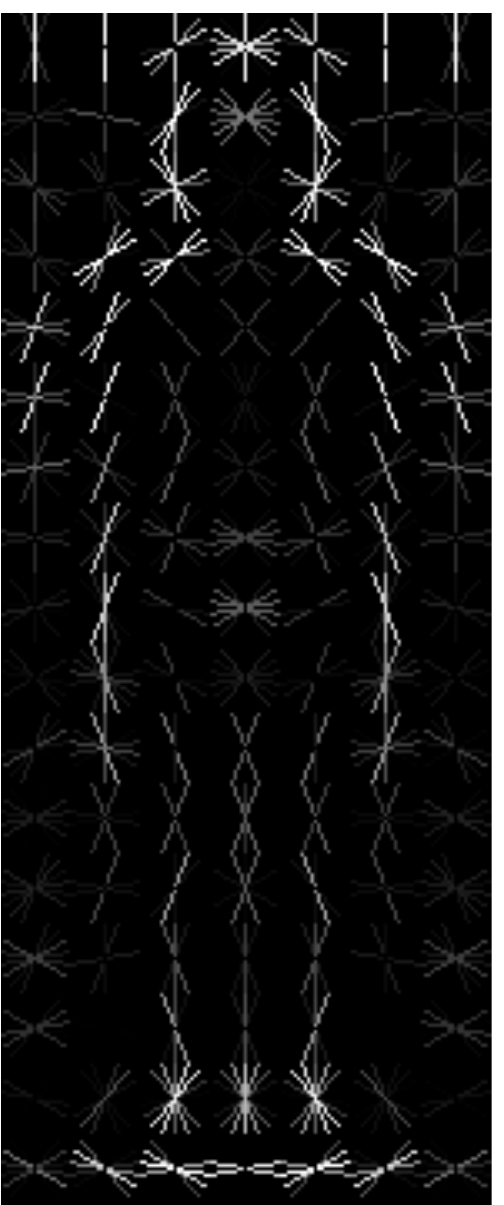}};
	\node[right of=a, node distance=1.7cm] (b) {\includegraphics[width=0.19\linewidth]{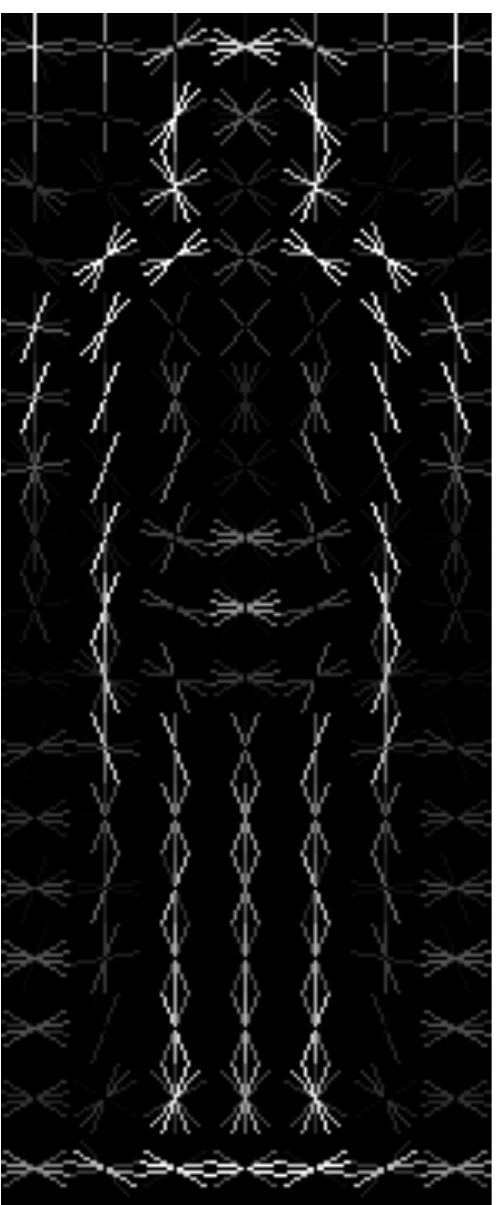}};
	\node[above of=a, node distance=2.2cm] (c) {VCF};
	\node[above of=b, node distance=2.2cm] (c) {ZAVCF};
\end{tikzpicture}
}
\caption{INRIA Pedestrian Detection: (a) performance comparison of various correlation filter designs and (b) visualization of the learned pedestrian templates for the VCF formulation and its Zero-Aliasing version. Notice that the ZAVCF template has a more pronounced pedestrian, especially the legs.}
\label{fig:inria}
\end{figure}

\begin{figure*}[!ht]
\centering
\includegraphics[scale=0.21]{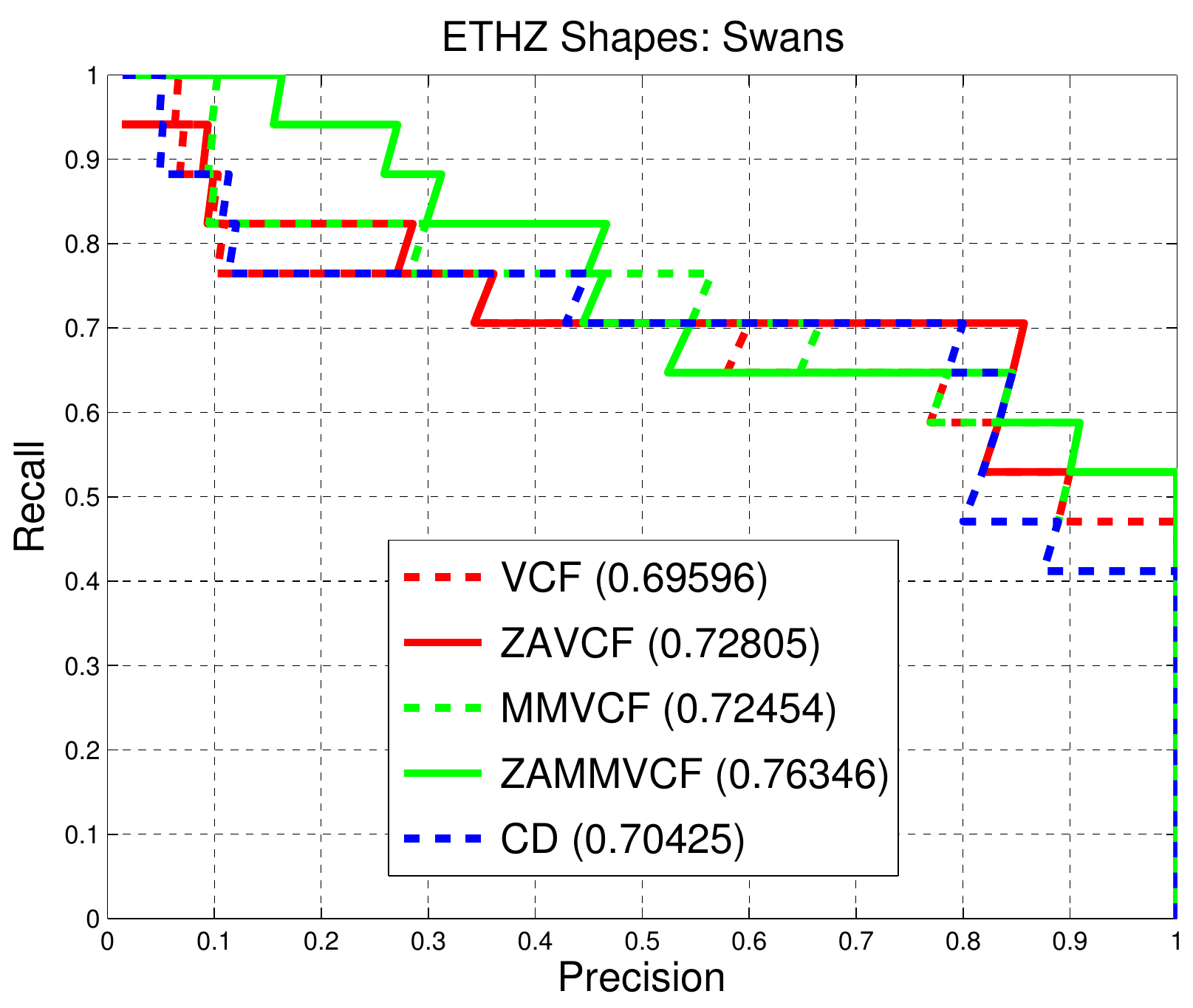}
\includegraphics[scale=0.21]{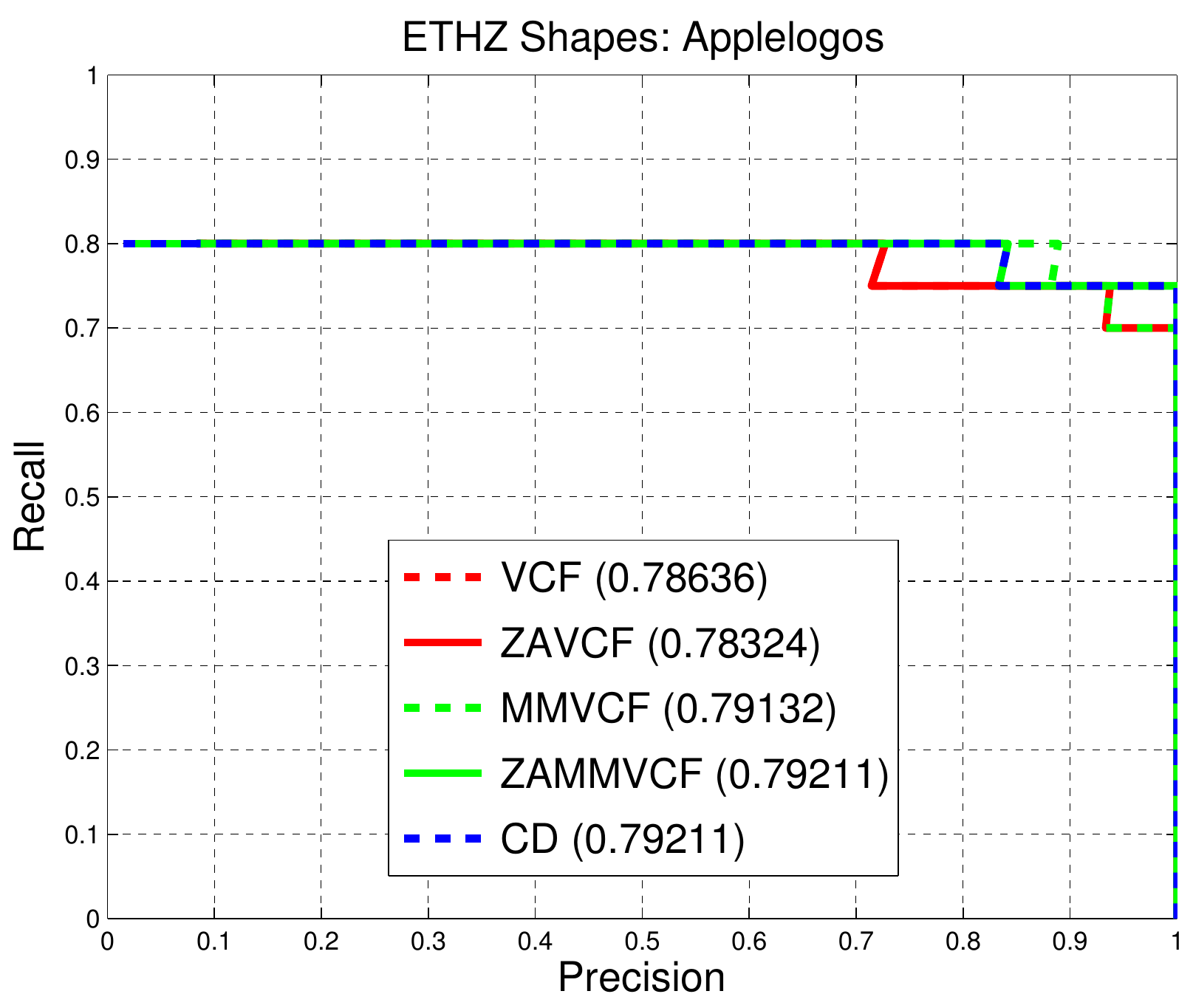}
\includegraphics[scale=0.21]{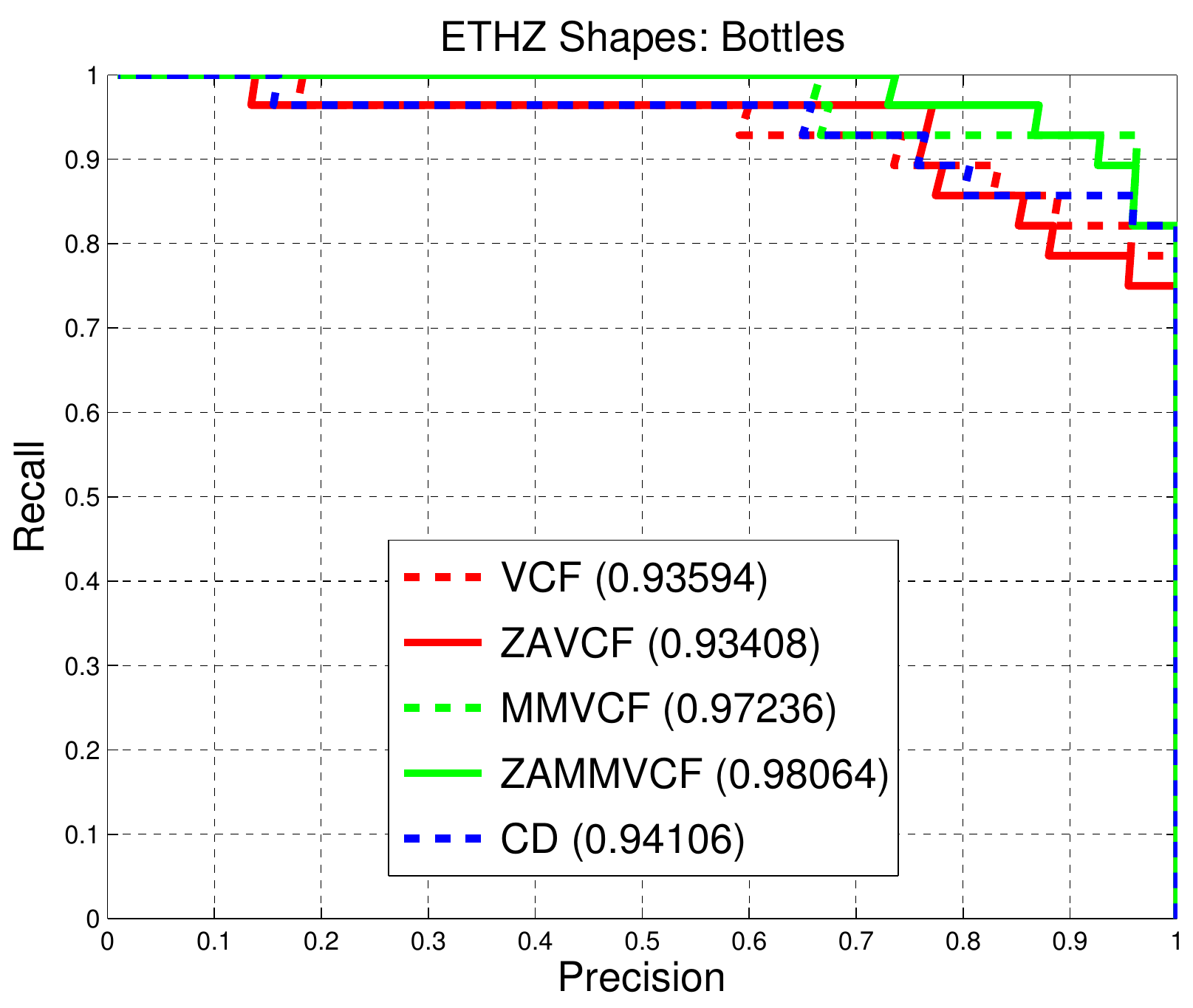}
\includegraphics[scale=0.21]{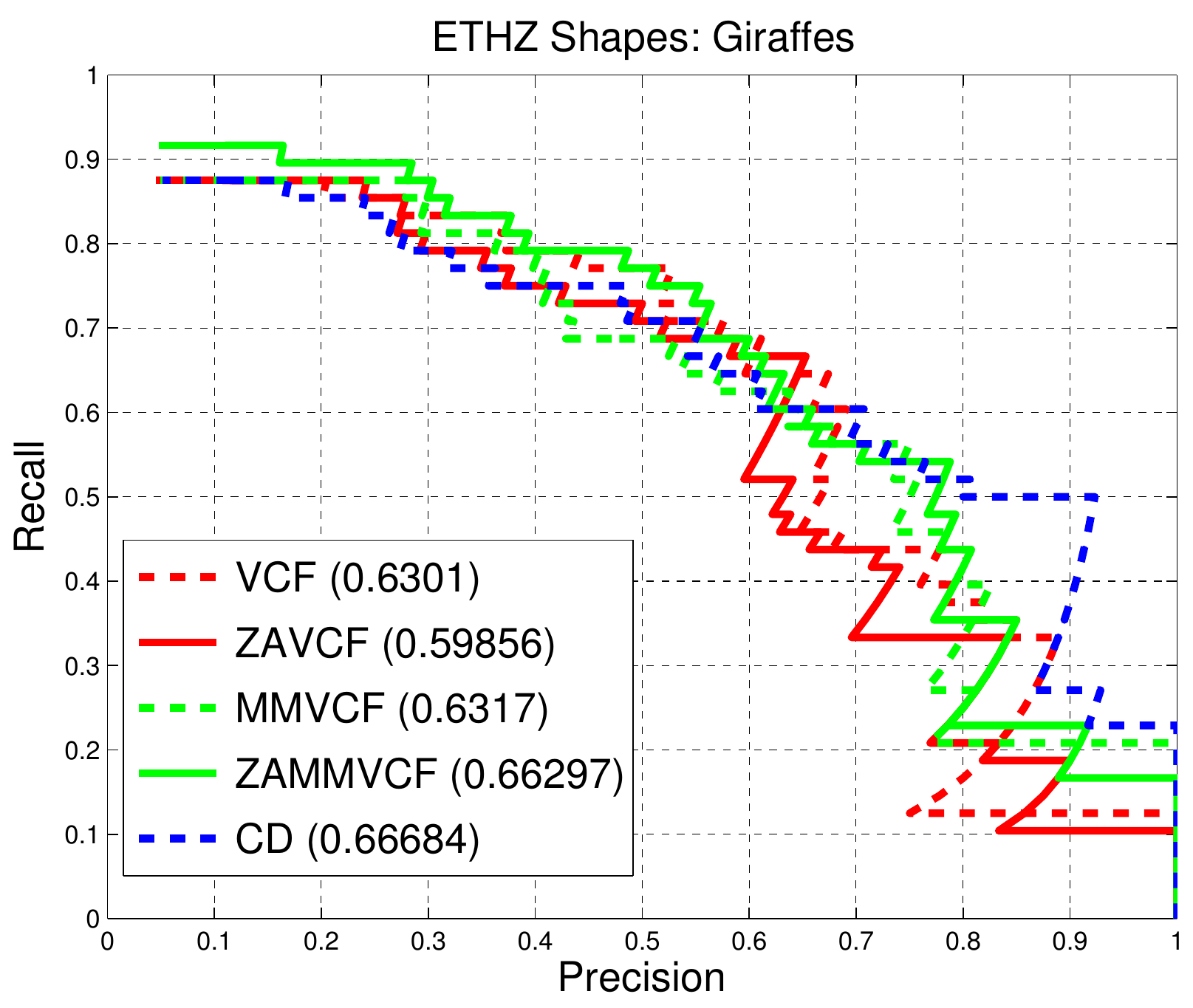}
\includegraphics[scale=0.21]{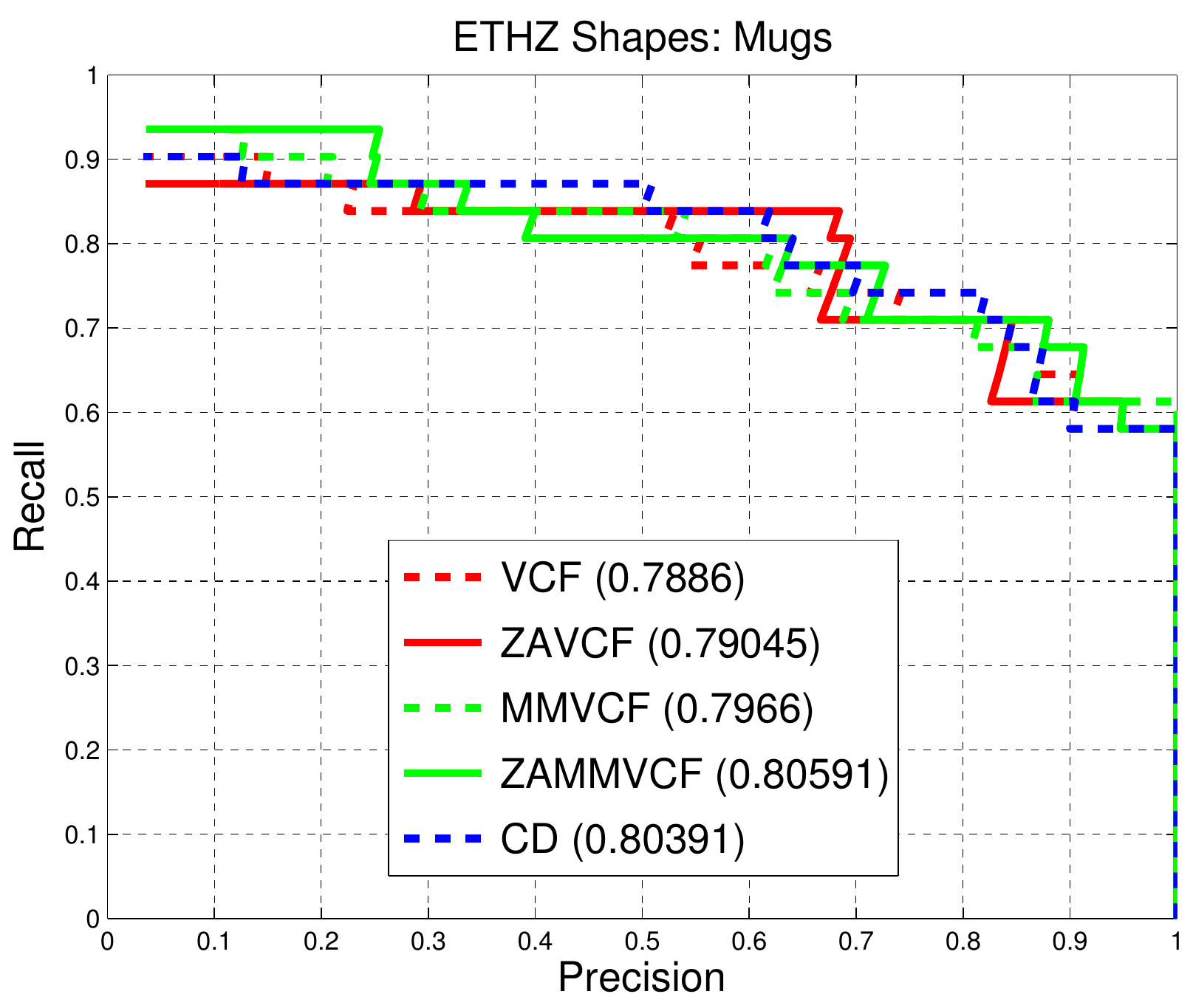}
\caption{Comparison of various correlation filter designs for object shape detection on the ETHZ Shapes dataset.}
\label{fig:ethz}
\end{figure*}

We evaluated various CF and ZACF designs for pedestrian detection on the INRIA Pedestrians \cite{dalal2005histograms} dataset and the ETHZ Shapes dataset \cite{schmid2008groups}. Instead of mining for hard negatives, as is commonly done, we simply consider a large set of negative windows following the protocol in \cite{henriques2013beyond}. For this set of experiments we use vector-valued HOG feature representation \cite{dalal2005histograms}, i.e., we have $K=36$ different feature channels. The ETHZ Shapes dataset consists of 5 categories (Mug, Bottle, Swan, Giraffe and Apple Logo) each consisting of between 22 and 45 positive samples. We compare the following different CFs: Vector Correlation Filter (VCF) \cite{Boddeti_VCF}, Maximum-Margin Vector Correlation Filter (MMVCF) \cite{boddeti2014maximum}, Circulant Decomposition (CD) \cite{henriques2013beyond} based CF, and the ZA filters ZAVCF (Eq. \ref{eq:ZAMOSSE_optim_fxn}), ZAMMVCF (Eq. \ref{eq:zammcf}). In Table \ref{table:object_detection} we present the average precision for the INRIA dataset and the mean average precision over all the ETHZ Shapes and Fig. \ref{fig:inria}(a) and Fig.\ref{fig:ethz} show the precision-recall curves for the INRIA pedestrian dataset and each object in the ETHZ Shapes dataset respectively. We note that among single linear template based methods, ZAMMVCF results in the best performance on both these datasets. The VCF and ZAVCF templates learned from the INRIA pedestrian dataset are shown in Fig. \ref{fig:inria}(b).
\begin{table}[!h]
\centering
\caption{Object Detection: Average Precision (\%)}
\label{table:object_detection}
\begin{tabular}{cccccc}
Dataset & VCF \cite{Boddeti_VCF} & ZAVCF & MMVCF \cite{boddeti2014maximum} & ZAMMVCF & CD \cite{henriques2013beyond} \\
\hline
\hline
INRIA & 79.79 & 80.81 & 83.19 & 84.07 & 79.56 \\
\hline
ETHZ & 76.74 & 76.70 & 78.33 & 80.10 & 78.16 \\
\hline
\end{tabular}
\end{table}
\section{\label{sec:Conclusions}Conclusions}
In this paper, we proposed and investigated a fundamental and significant advancement to the design of CFs. Existing CF designs that are formulated in the frequency domain do not explicitly account for the fact that multiplication of two DFTs in the frequency domain corresponds to a circular correlation in the spatial domain. As a result, existing CF designs do not actually optimize their intended cost functions. In this paper, we present a solution to this problem (ZACFs) that completely removes circular correlation effects from CF designs. While we have explicitly shown new filter derivations for several popular filter designs, this approach can be used with many more CFs and with other approaches. To address the computational challenges caused by the ZA constraints we introduced the RACF designs as an approximate solution as well as proximal gradient descent based algorithms for exactly solving for the various ZACFs. We have shown that our methods eliminate aliasing and lead to significantly better results across the board for many different CFs and for different datasets.

\section*{Acknowledgment}
The authors would like to acknowledge the U.S. Air Force Research Laboratory (AFRL) DoD Supercomputing Resource Center (DSRC) for providing the computational resources to quickly run many of the experiments presented.

\bibliographystyle{ieeetr}
\bibliography{mybib}

\begin{IEEEbiography}[{\includegraphics[width=1in,height=1.25in]{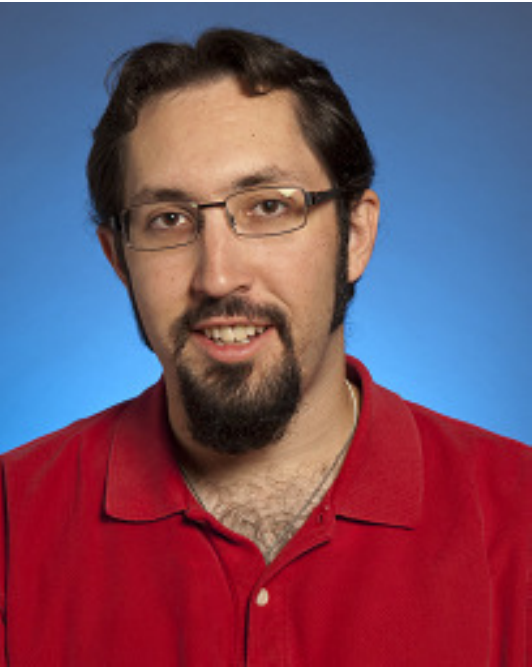}}]{Joseph A. Fernandez}
Joseph A. Fernandez received his BS in Electrical Engineering from New Mexico Tech in 2008 and his MS in Electrical and Computer Engineering from Carnegie Mellon University in 2010. He received his Ph.D. degree from Carnegie Mellon University in 2014 and is currently with Northrop Grumman Corporation in the Future Technical Leaders program. Joseph was awarded the best paper gold award for a student/young researcher at ISPA 2013 for his paper ``Zero-Aliasing Correlation Filters". He is a 2010 National Defense Science and Engineering Graduate (NDSEG) Fellow. His research interests include correlation filters and improving the efficiency of signal processing algorithms. 
\end{IEEEbiography}
\begin{IEEEbiography}[{\includegraphics[width=1.1in,height=1.25in,clip,keepaspectratio]{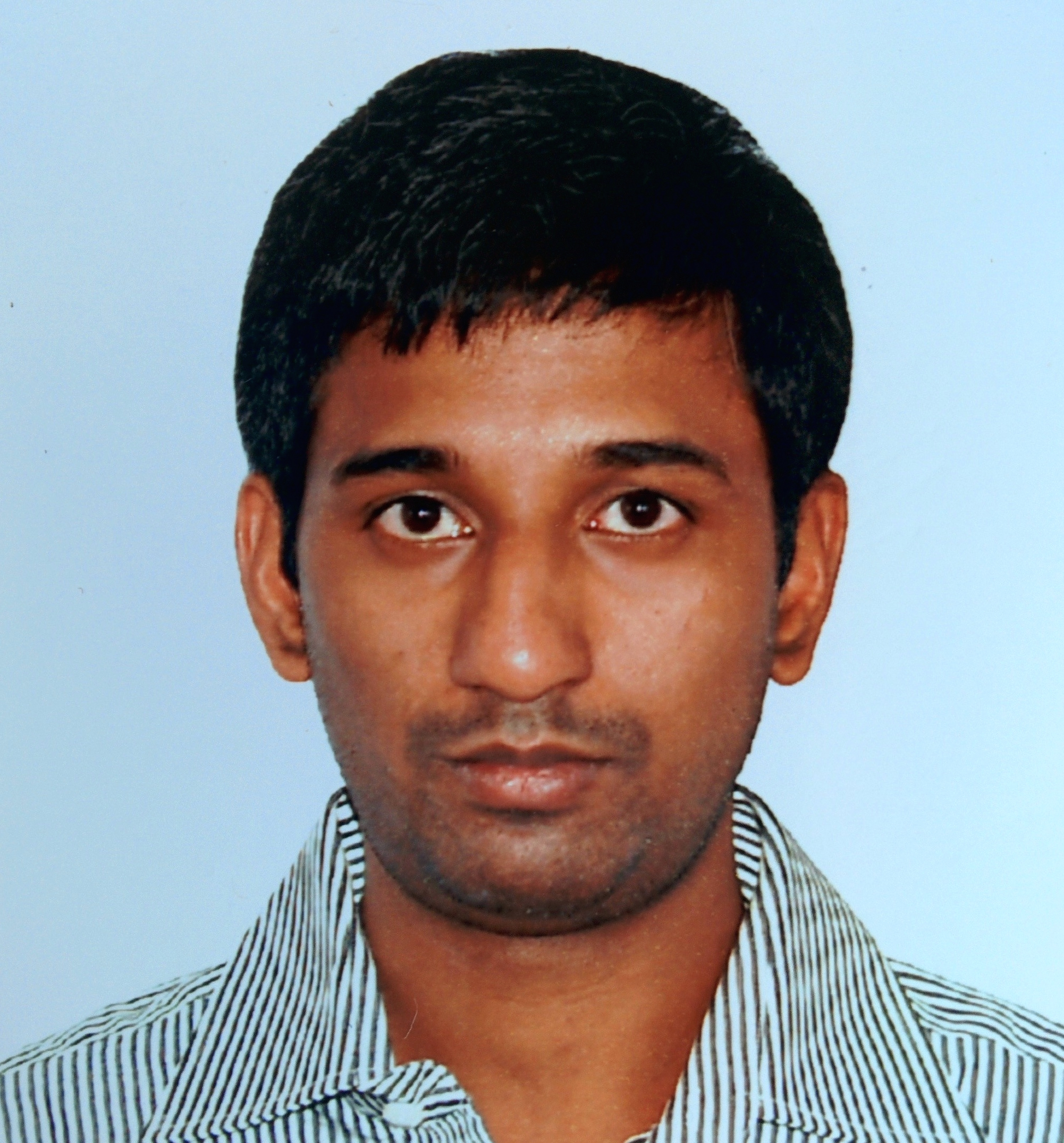}}]{Vishnu Naresh Boddeti} received a BTech degree in Electrical Engineering from the Indian Institute of Technology, Madras in 2007, and his MS and Ph.D. degree in Electrical and Computer Engineering program at Carnegie Mellon University. He is currently at postdoc at the Robotics Institute at Carnegie Mellon University. His research interests are in Computer Vision, Pattern Recognition and Machine Learning. He was awarded the Carnegie Institute of Technology Dean's Tuition Fellowship in 2007 and received the best paper award at the BTAS conference in 2013.
\end{IEEEbiography}
\begin{IEEEbiography}[{\includegraphics[width=1in,height=1.25in]{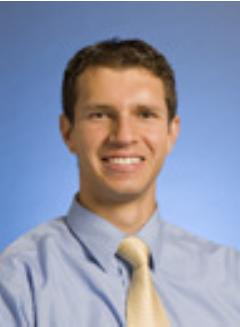}}]{Andres Rodriguez}
is a research scientist at the Air Force Research Laboratory Sensors Directorate. His research interests are in pattern recognition, machine learning, and computer vision. He received a BS and MS degree in Electrical Engineering from Brigham Young University in 2006 and 2008, respectively, and his PhD in Electrical and Computer Engineering from Carnegie Mellon University in 2012. He was awarded the Carnegie Institute of Technology Dean's Tuition Fellowship in 2008 and the Frank J. Marshall Graduate Fellowship in 2009. He has published over a dozen technical papers and received the best student paper award at the SPIE ATR conference in 2010.
\end{IEEEbiography}
\begin{IEEEbiography}[{\includegraphics[width=1.1in,clip,keepaspectratio]{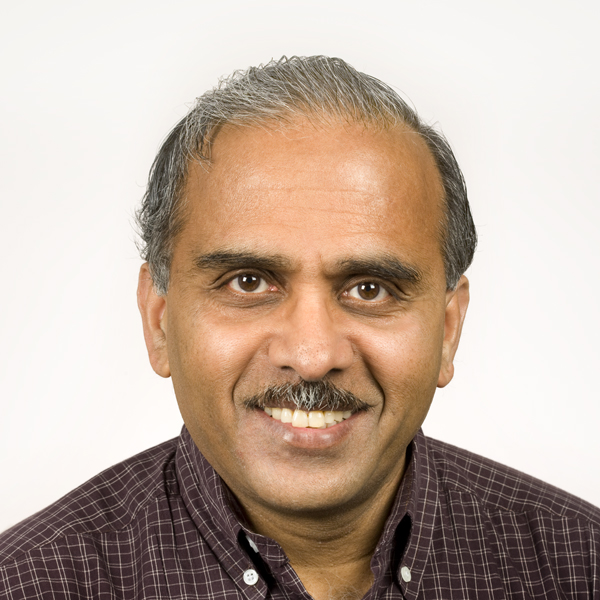}}]{B. V. K. Vijaya Kumar}
is the U.A. \& Helen Whitaker Professor of Electrical and Computer Engineering (ECE) and the Associate Dean for Graduate and Faculty Affairs of College of Engineering at Carnegie Mellon University (CMU). Professor Kumar served as the Interim Dean of College and Engineering at CMU from 2012 to 2013. Professor Kumar's research interests include Pattern Recognition, Biometrics and Coding and Signal Processing for Data Storage Systems. Professor Kumar is a Fellow of IEEE, SPIE, OSA and IAPR. His publications include the book entitled Correlation Pattern Recognition, twenty book chapters and more than 550 technical papers. He served as a Pattern Recognition Topical Editor for Applied Optics and as an Associate Editor for IEEE Transactions on Information Forensics and Security. Prof. Kumar is a co-recipient of the 2009 Carnegie Institute of Technology\textquoteright s Outstanding Faculty Research Award.
\end{IEEEbiography}
\end{document}